\documentclass[conference]{IEEEtran}
\IEEEoverridecommandlockouts
\usepackage{cite}
\usepackage{amsmath,amssymb,amsfonts}
\usepackage{graphicx}
\usepackage{url} 
\usepackage{multirow}
\usepackage{textcomp}
\usepackage{xcolor}
\usepackage{siunitx}
\usepackage[ruled,vlined]{algorithm2e}
\usepackage{algpseudocode}
\usepackage[labelfont=bf,labelsep=space]{caption}
\usepackage{subcaption}
\usepackage{upgreek}
\usepackage{hyperref}
\usepackage{soul}

\def\BibTeX{{\rm B\kern-.05em{\sc i\kern-.025em b}\kern-.08em
    T\kern-.1667em\lower.7ex\hbox{E}\kern-.125emX}}
\begin{document}

\title{Event-based Spiking Neural Networks for Object Detection: A Review of Datasets, Architectures, Learning Rules, and Implementation}

\author{\IEEEauthorblockN{Craig Iaboni}
\IEEEauthorblockA{\textit{Ying Wu College of Computing (YWCC)} \\
\textit{New Jersey Institute of Technology}\\
Newark, NJ, USA \\
csi3@njit.edu}
\and
\IEEEauthorblockN{Pramod~Abichandani}
\IEEEauthorblockA{\textit{Associate Professor at the Newark College of Engineering (NCE)} \\
\textit{Director of the Robotics and Data Laboratory (RADLab)} \\
\textit{New Jersey Institute of Technology}\\
Newark, NJ, USA \\
pva23@njit.edu}
}

\maketitle

\begin{abstract}
Spiking Neural Networks (SNNs) represent a biologically inspired paradigm offering an energy-efficient alternative to conventional artificial neural networks (ANNs) for Computer Vision (CV) applications. This paper presents a systematic review of datasets, architectures, learning methods,  implementation techniques, and evaluation methodologies used in CV-based object detection tasks using SNNs. Based on an analysis of 151 journal and conference articles, the review codifies: 1) the effectiveness of fully connected, convolutional, and recurrent architectures; 2) the performance of direct unsupervised, direct supervised, and indirect learning methods; and 3) the trade-offs in energy consumption, latency, and memory in neuromorphic hardware implementations. An open-source repository along with detailed examples of Python code and resources for building SNN models, event-based data processing, and SNN simulations are provided. Key challenges in SNN training, hardware integration, and future directions for CV applications are also identified.
\end{abstract}


\section*{Code Availability}
The code for this work is accessible at: \newline \url{https://github.com/radlab-sketch/Event-SNN-Resources}.

\section{Introduction}
Spiking Neural Networks (SNNs) represent a promising paradigm in artificial intelligence and computational neuroscience, closely mimicking the neuronal dynamics observed in biological brains. Unlike traditional artificial neural networks (ANNs) that rely on continuous activation functions, SNNs use discrete, event-driven spikes to encode and process information. This characteristic enables SNNs to exploit temporal information and exhibit energy-efficient computation, making them suitable for applications in real-time processing and low-power environments \cite{maass2000computational, pfister2006triplets}. 

The evolution of neural networks can be broadly categorized into three generations, each advancing computational capabilities. The first generation introduced the Perceptron and linear models, capable of solving linearly separable problems but limited by simple threshold activations \cite{rosenblatt1958perceptron, minsky2017perceptrons}. The second generation overcame these limitations with Multilayer Perceptrons (MLPs), using non-linear activations (e.g., sigmoid \cite{verhulst1838notice}, ReLU \cite{nair2010rectified}) and the backpropagation algorithm to train deep networks for complex tasks, including CNNs and RNNs for image and sequence processing \cite{rumelhart1986learning, lecun2015deep}. The third generation, SNNs, adopts a biologically inspired approach, using discrete spikes and temporal dynamics to encoding information for energy-efficent, event-driven computation suitable for time-sensitive applications.

The fundamental units of SNNs, known as spiking neurons, communicate via synaptic connections that transmit spikes, or action potentials, in response to reaching specific membrane potential thresholds. This mechanism allows SNNs to model complex temporal patterns and dependencies, offering significant advantages in tasks that involve sequential data, such as speech recognition, video analysis, and event-based sensing \cite{ghosh2009new, tavanaei2019deep}. The design of efficient and scalable SNN architectures remains an ongoing research focus, particularly when addressing large-scale datasets and complex cognitive tasks \cite{ponulak2010supervised, tavanaei2019bp}. 

Given the rapid advancements in SNN research, it is crucial to synthesize the existing literature to build a comprehensive understanding of the current state of SNNs. While prior surveys and reviews focus on learning rules \cite{yi2023learning, agebure2021survey}. hardware implementations and optimizations \cite{pham2021review, bouvier2019spiking}, and practical applications \cite{wang2023brain, niu2023research, schliebs2013evolving, ibad2020evolving}, they address these areas in isolation. This has led to what can be described as a collection of approaches that, while individually promising, lack integration, particularly in scaling SNNs for real-world applications \cite{eshraghian2023training}. As noted in recent reviews, SNN architectures and their learning algorithms are still in development, with significant efforts required to overcome challenges in unifying these components. In contrast, the goal of this paper is to provide a cohesive review of the state-of-the-art in applied SNNs, with a focus on bridging gaps between hardware, learning algorithms, and implementation, thereby advancing their adoption in real-world settings.

The remainder of the paper is structured as follows: Section \ref{sec:method} outlines the literature review process, following PRISMA guidelines to identify, screen, and analyze relevant research on SNNs, focusing on their application in CV. Section \ref{sec:framework} presents a structured framework for developing and deploying SNNs in CV tasks. Section \ref{sec:neural_dynamics} delves into the biological inspirations behind SNNs, detailing the behavior of biological neurons, various types of neurons modeled in SNNs, and the mechanisms of neural coding that enable these models to process temporal and spatial information effectively. In Section \ref{sec:prominent_datasets}, prominent datasets used in SNN research are discussed. Section \ref{sec:implementation_medium} examines the simulation environments and neuromorphic hardware
platforms used for implementing SNNs. In Section \ref{sec:architecture}, SNN architectures including fully connected networks, hierarchical structures, convolutional networks, deep belief networks (DBNs), and recurrent neural networks (RNNs) are explored. Section \ref{sec:learning_rules} categorizes the learning methodologies used in SNNs into unsupervised, supervised, and indirect learning approaches. Section \ref{sec:evaluation} focuses on the metrics used to assess SNN performance. Finally, Section \ref{sec:future_directions} identifies areas for further research, highlighting the ongoing advancements and potential future developments in SNNs.

Python code and examples for the event-based SNN learning techniques and methods for event data processing, spike generation, and SNN model implementation discussed within this paper can be found at: \url{https://github.com/radlab-sketch/Event-SNN-Resources}.

\begin{figure}[t!]
    \centering
    \includegraphics[scale = 0.45]{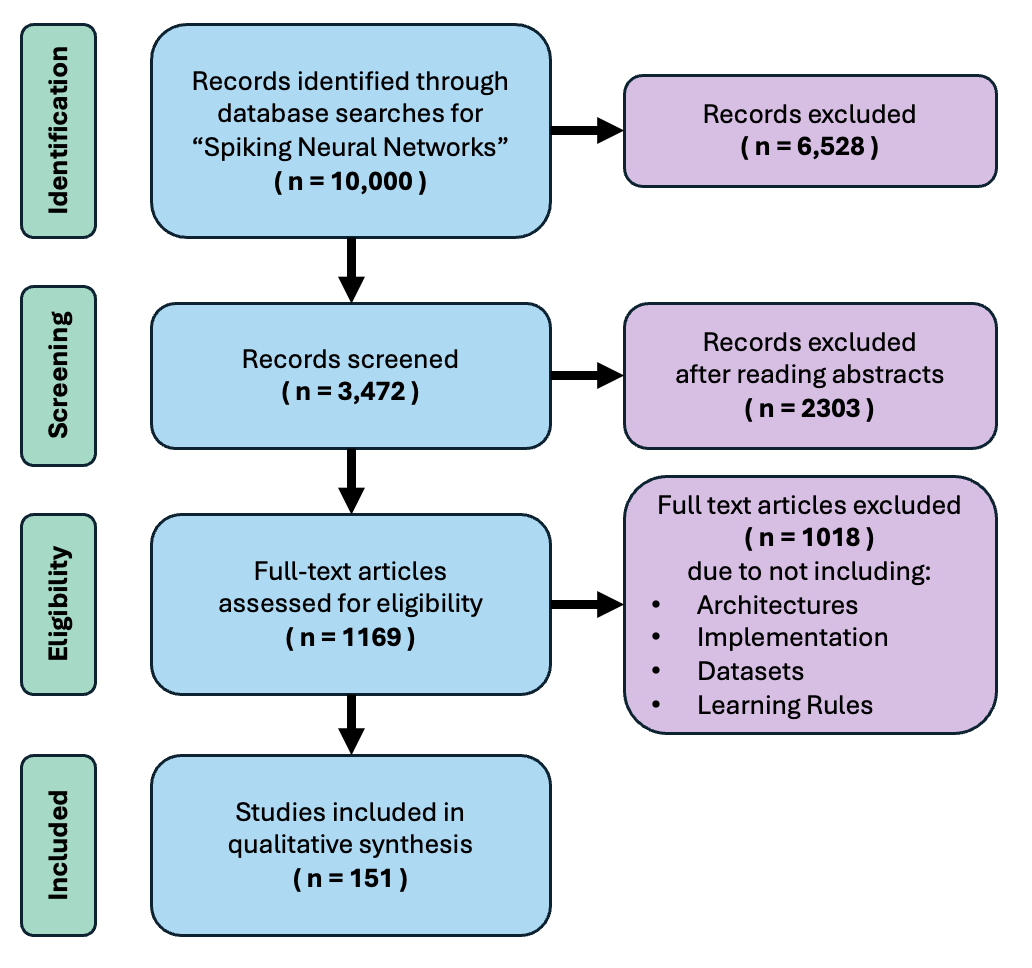}
    \caption{Overview of the PRISMA approach used for conducting the literature review for this study.}
    \label{fig:prisma}
\end{figure}

\section{Methodology}
\label{sec:method}

A comprehensive literature search was conducted to gather publications on SNNs and their application in various domains, focusing on computer vision. The guidelines for this review adhered to the Preferred Reporting Items for Systematic Reviews and Meta-Analyses (PRISMA) approach \cite{moher2009preferred}. The PRISMA flowchart, depicted in Figure \ref{fig:prisma}, illustrates the stages of identification, screening, eligibility assessment, and inclusion of articles relevant to this study.

\subsection{Identification}
The initial search used terms such as "Spiking Neural Networks," "SNN applications," "SNN learning rules," "SNN architecture," "neuromorphic hardware," and "SNN evaluation." This search yielded over 10,000 results from various databases, including IEEE Xplore, PubMed, Scopus, and Google Scholar. Abstracts of these results were scanned to refine search terms, leading to combinations such as "spiking neural networks in computer vision," "SNNs in neuromorphic computing," and "SNN training methods." This refinement narrowed down the number of relevant articles to 3,472.

\subsection{Screening}
Further screening was performed on literature from a number of journals (IEEE, Springer, Wiley, Elsevier among) and conference proceedings. Select examples of journals include IEEE Transactions on Neural Networks and Learning Systems, Frontiers in Neuroscience, Neural Computation, IEEE Transactions on Pattern Analysis and Machine Intelligence, International Journal of Computer Vision, IEEE Transactions on Robotics, Pattern Recognition Letters, and Neurocomputing. Select examples of conference proceedings include Conference on Neural Information Processing Systems, Conference on Computer Vision and Pattern Recognition, International Conference on Computer Vision, European Conference on Computer Vision, and the International Conference on Neuromorphic Systems. This screening narrowed the selection to approximately 1,169 articles (i.e. 2303 articles were excluded).

\begin{figure*}[t!]
    \centering
    \includegraphics[scale = 0.5]{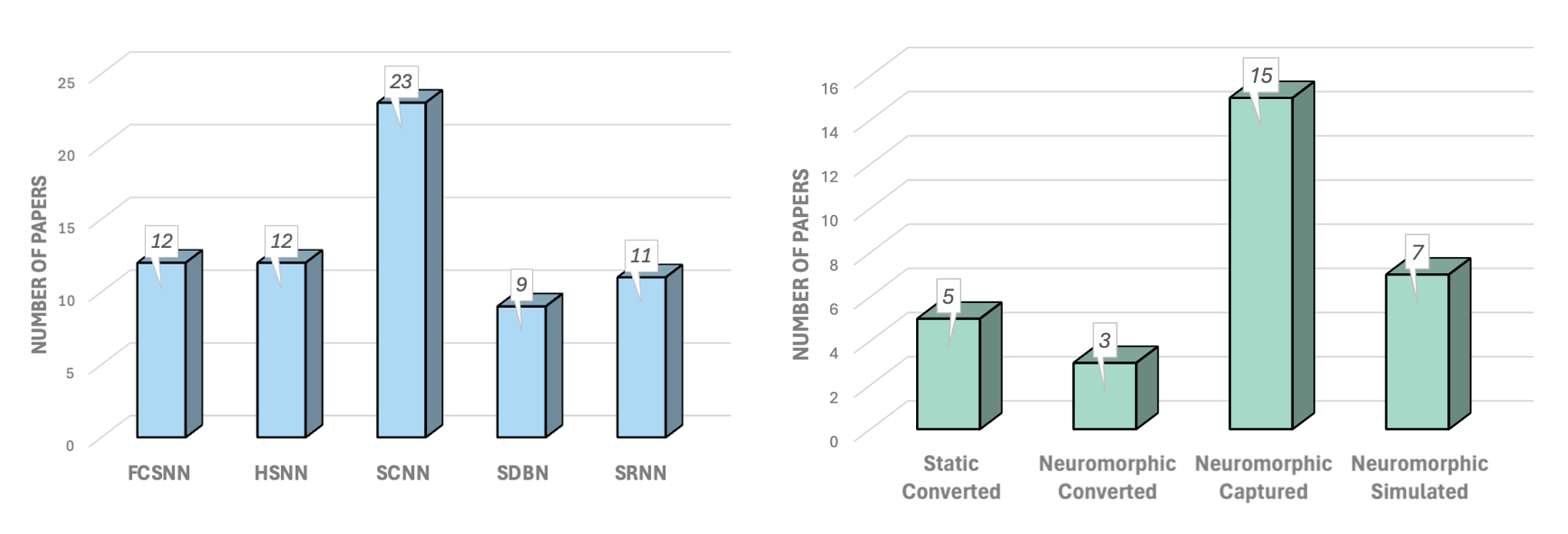}
    \caption{(Left) Different architectures discussed in the reviewed papers. The most discussed architecture is SCNNs. (Right) Distribution of dataset types reported in the reviewed articles.}
    \label{fig:bars}
\end{figure*}

\begin{figure*}[t!]
    \centering
    \includegraphics[scale = 0.5]{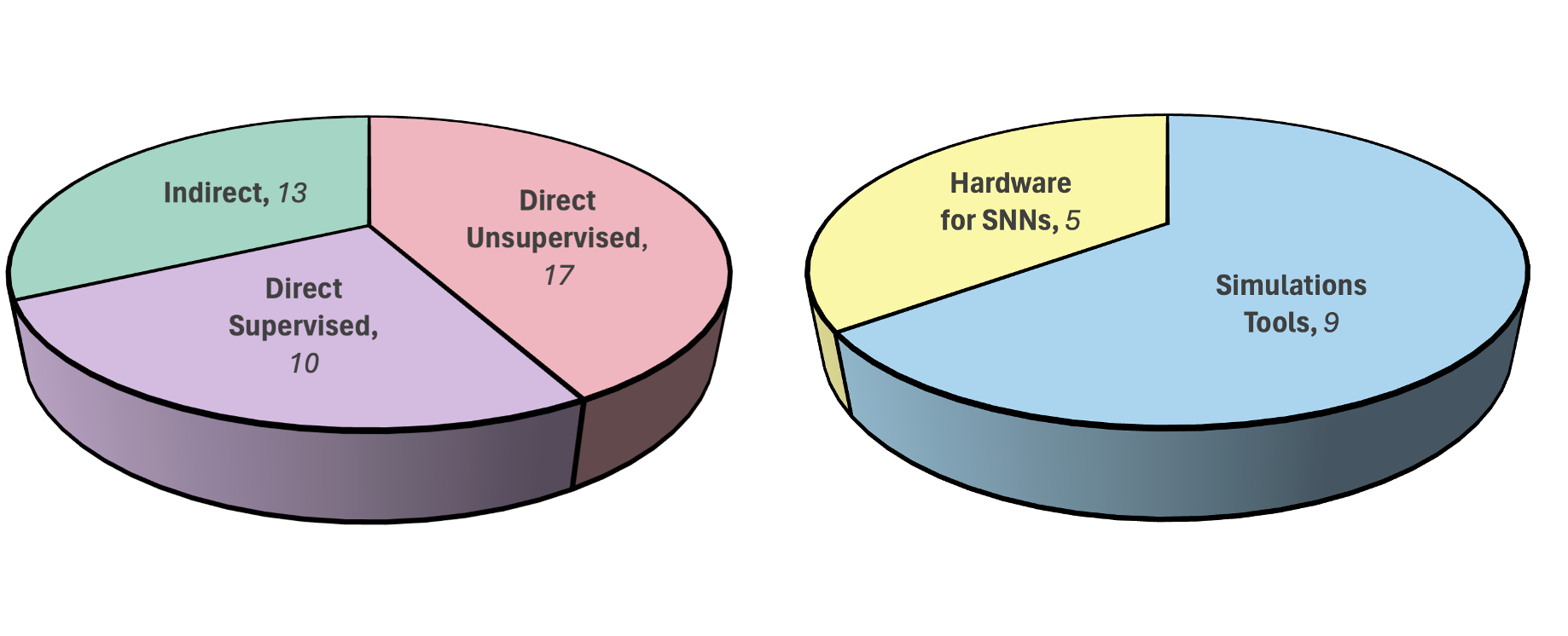}
    \caption{ (Left) The distribution of learning methods employed in the reviewed papers. (Right) The distribution of implementation mediums for SNNs across the reviewed papers, distinguishing between software tools and hardware-based implementations.}
    \label{fig:pies}
\end{figure*}

\subsection{Eligibility}
The eligibility criteria for inclusion in this review were:
\begin{itemize}
    \item Research articles presenting novel SNN architectures or modifications to existing models.
    \item Studies detailing learning rules specific to SNNs, including unsupervised, supervised, and indirect learning methods.
    \item Papers discussing the implementation of SNNs on neuromorphic hardware or through simulation.
    \item Articles on datasets for SNNs, including static converted datasets, neuromorphic converted datasets, native captured datasets, and simulation data.
    \item Publications from 2000 to 2023 to capture the evolution and recent advancements in the field.
    \item Peer-reviewed articles published in English.
\end{itemize}
After applying these criteria, 151 articles were deemed suitable for detailed review.

\subsection{Analysis}
The selected articles were systematically analyzed. The article title, authors, publication year, SNN application domain, architecture type, learning rules, implementation medium, and evaluation metrics were extracted and tabulated.

\begin{itemize}
    \item \textbf{Neuronal Dynamics}: Articles were classified based on their contributions to understanding the neuronal dynamics of SNNs, including biological neuron models, types of neurons, and neural coding strategies.
    
    \item \textbf{Prominent SNN Datasets}: As shown in Figure \ref{fig:bars} (Right), studies were categorized according to the datasets used, emphasizing relevant dataset categories in the field. 
    
    \item \textbf{Implementation Mediums}: The papers were classified based on whether the SNN implementations were simulated or deployed on neuromorphic hardware. Figure \ref{fig:pies} (Right) shows the distribution of works in each category.
   
    \item \textbf{Architecture}: As illustrated in Figure \ref{fig:bars} (Left), different SNN architectures, such as fully connected (FCSNN), hierarchical (HSNN), convolutional (SCNN), deep belief networks (SDBN), and recurrent networks (SRNN), were identified and categorized.
   
    \item \textbf{Learning Rules}: The articles were classified based on direct unsupervised, indirect, and direct supervised learning methodologies. Figure \ref{fig:pies} (Left) depicts the distribution of learning rules in the reported papers.
  
    \item \textbf{Evaluation}: Performance evaluation metrics used in the studies were cataloged, including accuracy, precision, recall, IoU, mAP, energy consumption, latency, and memory footprint.
\end{itemize}

\begin{figure*}[t!]
    \centering
    \includegraphics[scale = 0.5]{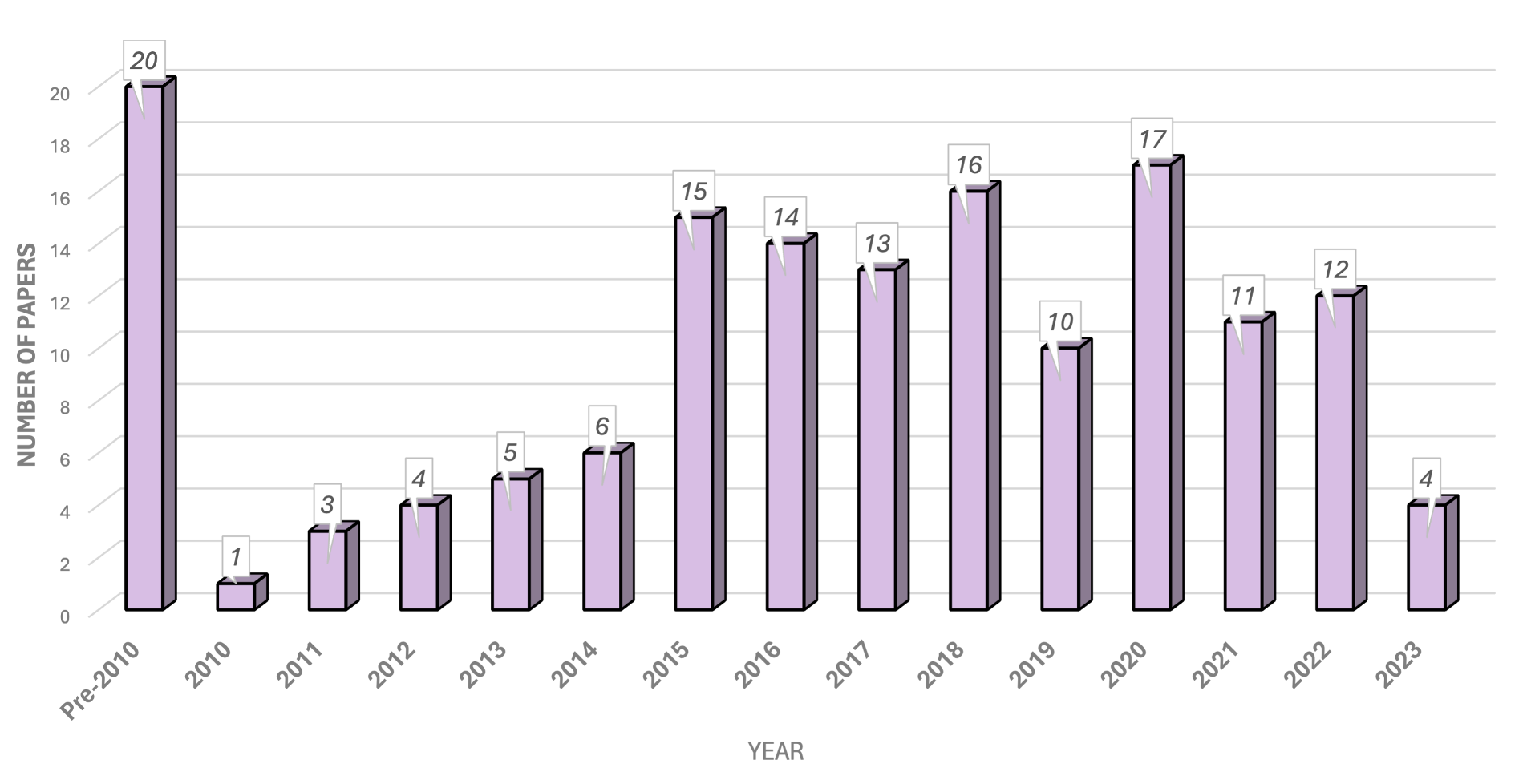}
    \caption{The systematic review presented in this paper encompasses publications from 2000 to 2023. Starting around 2015, there was a noticeable uptick in activity. This increase coincides with the growing traction of SNNs within the research community.}
    \label{fig:years}
\end{figure*}

\subsection{Results}
After a meticulous review based on the outlined criteria, 151 papers were selected for in-depth analysis. The distribution of these papers is as follows:
\begin{itemize}
    \item \textbf{Publication Years}: The selected articles span from 1997 to 2023, with a noticeable increase in publications starting 2015 as depicted in Figure \ref{fig:years}.
    \item \textbf{Prominent SNN Datasets}: 30 unique datasets were identified among the selected studies.
    \item \textbf{Implementation Mediums}: 9 papers reported on the use of simulated environments and 5 papers reported on neuromorphic hardware implementations.
    \item \textbf{Architecture}: Significant advancements in convolutional architectures were observed with 23 papers reporting on SCNNs. 12 papers each reported on FCSNNs and HSNNs respectively. 11 papers reported on SRNNs and 9 papers reported on SDBNs. 
    \item \textbf{Learning Rules}: 17 papers reported on Direct Unsupervised rules in contrast to 10 papers reporting on Direct Supervised rules. 13 papers reported on Indirect rules. 
    \item \textbf{Evaluation}: A comprehensive set of evaluation metrics provided a detailed performance assessment of the SNN models.
\end{itemize}

\section{SNN Framework for CV Practitioners}
\label{sec:framework}
This section outlines a structured approach to developing and deploying SNNs for CV tasks. It covers key stages including data acquisition, encoding, architecture selection, learning methods, implementation, and evaluation, with a focus on event-based vision sensors.

While SNNs are used in areas like speech recognition \cite{wu2020deep, dominguez2018deep}, robotics \cite{bing2018survey, hagras2004evolving, clawson2016spiking}, and bio-signal processing \cite{chu2022neuromorphic, yan2021energy, donati2019discrimination}, the SNN Framework for CV Practitioners, as depicted in Figure \ref{fig:framework}, outlines a comprehensive pipeline for developing and deploying SNNs for CV tasks. CV applications have been a prominent area of success for SNNs due to their alignment with event-based visual sensors.

The following pipeline outlines the key stages involved in developing and deploying SNNs for computer vision tasks, covering everything from data acquisition to evaluation:

\begin{enumerate}
    \item {\bf Data Type:} At the outset, the data type is categorized into event and RGB data. Event data is further divided into event-captured and event-converted data, which is crucial for dynamic vision sensors and event-based sensing applications. RGB data, representing standard image data, is processed through visual encoding methods.

    \item {\bf Encoding:} Once acquired, the data undergoes encoding to prepare it for SNN processing. Rate Coding and Temporal Coding are the primary strategies employed \cite{guo2021neural, kim2022rate, rueckauer2021temporal}. Rate Coding translates input intensity into spike rates, while Temporal Coding encodes information in the precise timing of spikes. These methods bridge the gap between conventional data formats (e.g., images, frames) and the spike-based nature of SNNs.

    \begin{figure*}[t!]
    \centering
    \includegraphics[scale = .7]{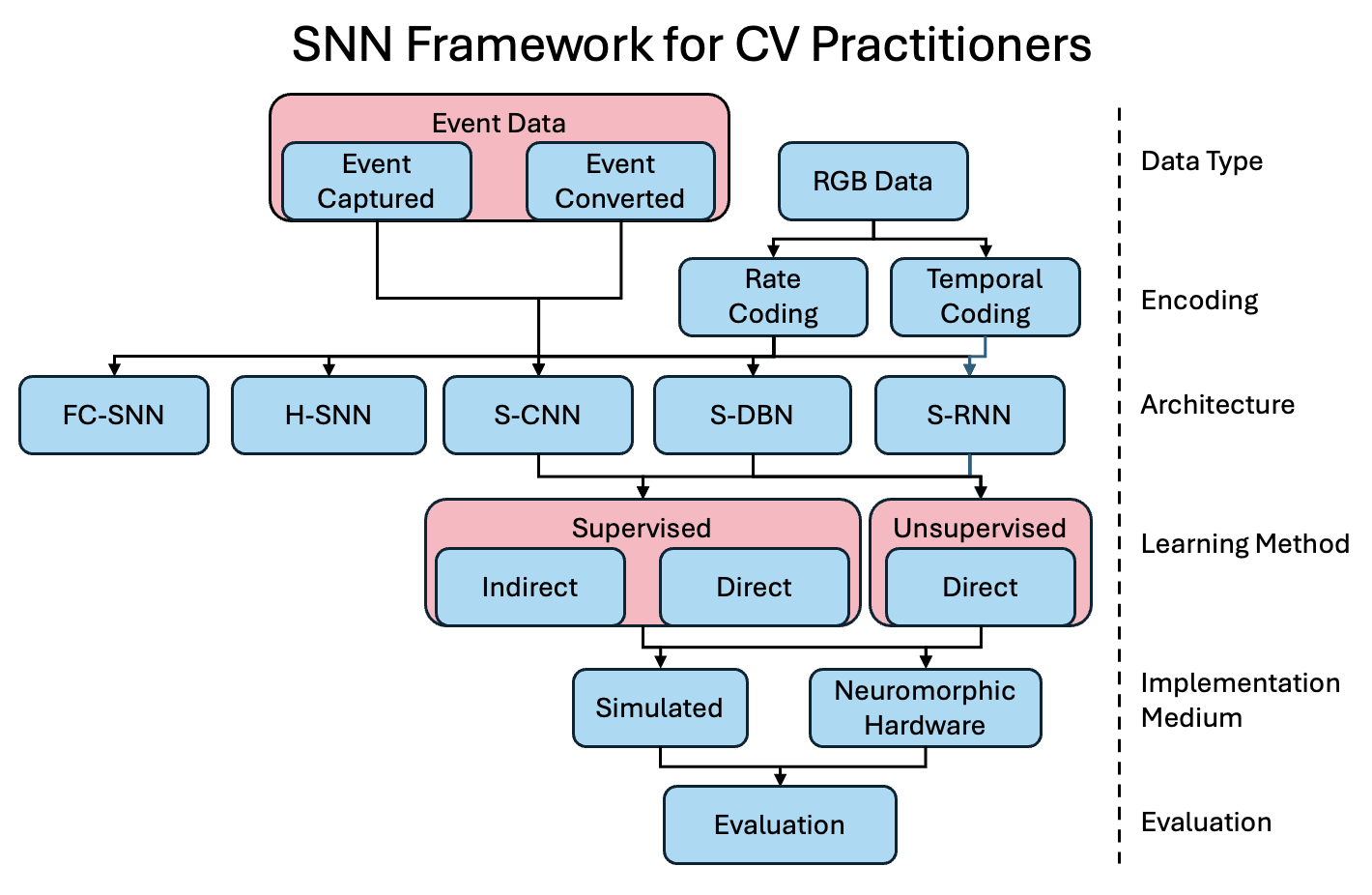}
    \caption{The SNN Framework for CV Practitioners provides a comprehensive approach from data acquisition to evaluation, specifically tailored for CV applications using SNNs.}
    \label{fig:framework}
\end{figure*}

    \item {\bf Architecture:} The architecture used depends on the specific task. FCSNNs connect every neuron in one layer to every neuron in the next, making them simple and flexible but computationally expensive, often used for basic pattern recognition tasks \cite{cohen2016skimming, diehl2015unsupervised, iyer2017unsupervised}. HSNNs organize neurons in a layered structure, enabling the network to process increasingly abstract features, which is useful for multi-stage tasks like scene understanding, and semantic or instance segmentation \cite{xiao2019event, negri2018scene, orchard2015hfirst}. SCNNs are well-suited for image classification and object detection, as they efficiently extract spatial features from images using convolutional layers \cite{lee2018deep, rueckauer2016theory, cao2015spiking}. SRNNs excel at capturing temporal dependencies, making them ideal for tasks involving sequential data \cite{kim2019simple, lotfi2020long, yin2020effective}. SDBNs build hierarchical representations of data, making them useful for feature extraction and dimensionality reduction, especially with unlabeled data \cite{lee2007sparse, kaiser2017spiking, stromatias2015robustness}.

    \item {\bf Learning Method:} SNN learning methods are typically categorized into supervised, unsupervised, and indirect learning approaches. Supervised learning in SNNs involves methods like temporal backpropagation and surrogate gradients, which adapt traditional gradient-based techniques to the temporal nature of spikes, enabling precise control over learning in tasks like classification \cite{bohte2000spikeprop, mckennoch2006fast, lee2016training}. Spike-timing-dependent plasticity (STDP) drives unsupervised learning in SNNs, allowing the network to self-organize by adjusting synaptic weights based on the timing of input spikes, making it ideal for feature extraction and pattern discovery \cite{masquelier2007unsupervised, diehl2015unsupervised}. In indirect learning, an ANN is pre-trained and then converted into an SNN, leveraging the computational efficiency of SNNs while benefiting from the pre-training process of traditional ANNs \cite{rueckauer2017conversion, hunsberger2015spiking}.

    \item {\bf Implementation Medium:} SNNs can be implemented in two primary environments. Simulated environments allow for detailed experimentation and testing during the development phase \cite{bekolay2014nengo, rueckauer2017conversion, hazan2018bindsnet, eshraghian2023training}. Neuromorphic hardware, on the other hand, enables efficient, low-power execution of SNNs, mimicking the brain's architecture for real-time, energy-efficient processing \cite{akopyan2015truenorth, davies2018loihi}.

    \item {\bf Evaluation:} The evaluation of SNNs focuses on both accuracy and efficiency, assessing how well the network performs on tasks like classification and detection, while also considering factors such as energy consumption and processing speed \cite{sorbaro2020optimizing, lemaire2022synaptic}.
\end{enumerate}

In addition to practitioners, this framework provides also provides a structured approach for educators focused on bringing SNNs into their classrooms \cite{abichandani2023artificial}.

\begin{figure}[b!]
    \centering
    \includegraphics[scale = 0.29]{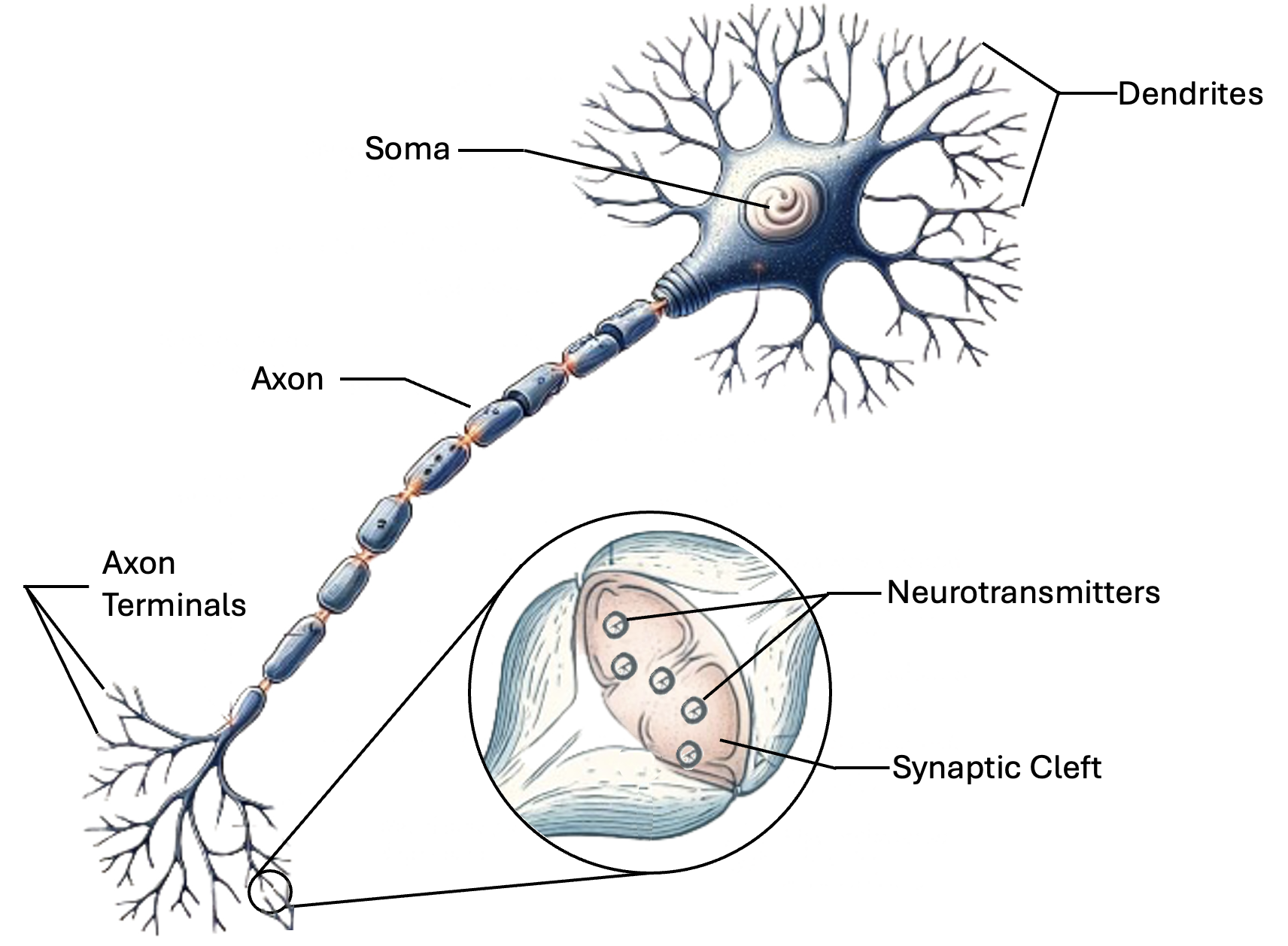}
    \caption{A biological neuron with its main components: soma, dendrites, axon, and axon terminals. Dendrites receive signals, which are integrated in the soma. The inset depicts neurotransmitter release at the synaptic cleft, enabling communication with other neurons.}
    \label{fig:neuron}
\end{figure}

\begin{figure*}[t!]
    \centering
    \includegraphics[scale = .4]{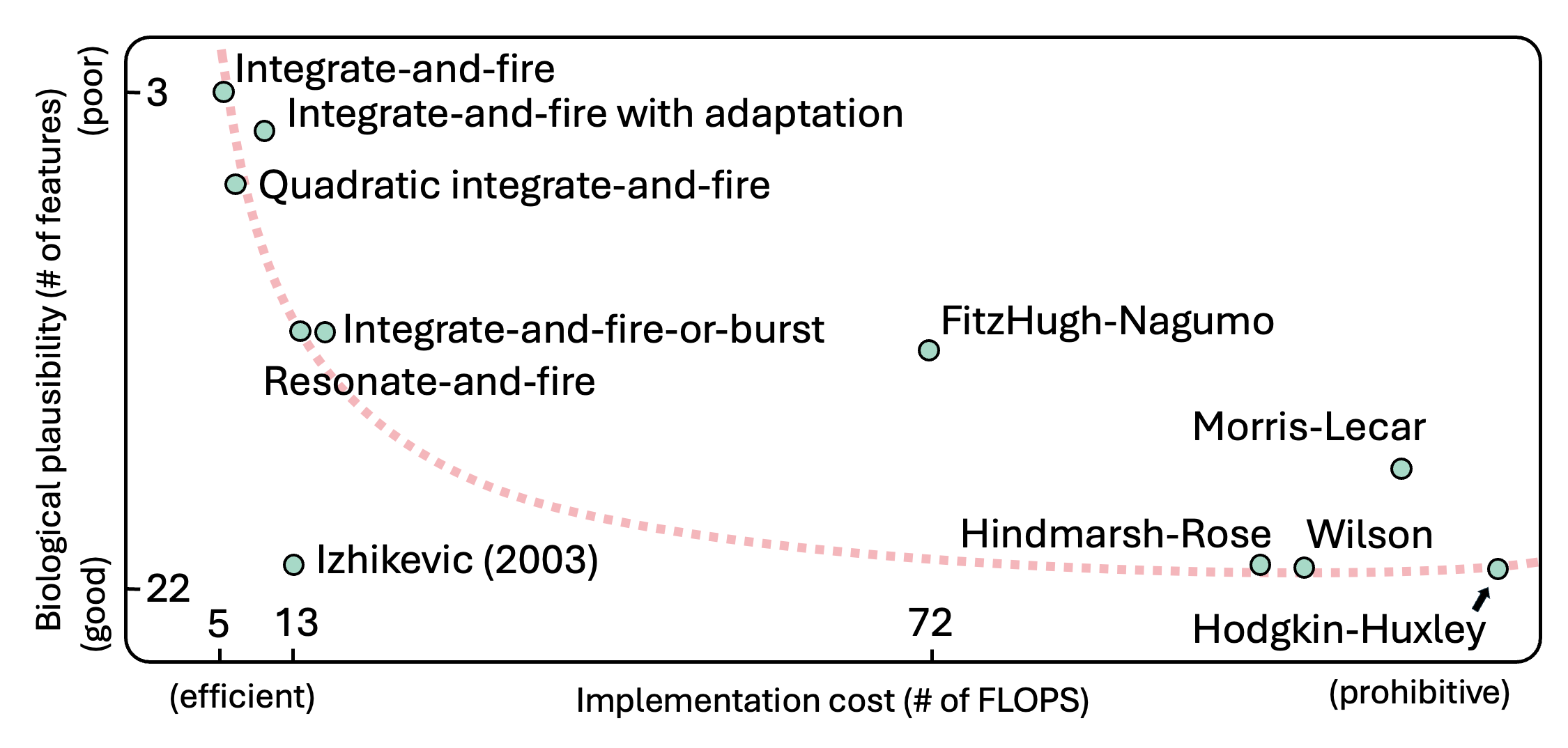}
    \caption{A comparison of spiking neuron models evaluated on implementation cost, in terms of number of FLOPS, and biological plausibility, in terms of number of features (adapted from \cite{izhikevich2004model})}.
    \label{fig:neuronmodels}
\end{figure*}

\section{Neuronal Dynamics}
\label{sec:neural_dynamics}
This section provides an overview of neuron models used in SNNs, including biological neuron behavior, different computational models, and neuronal reset behaviors. It also addresses neural coding methods for converting input data into spikes and decoding spike outputs for tasks like classification. Understanding the types of neurons used in SNNs is fundamental for appreciating their computational properties and dynamics.

\subsection{Biological Neurons}
Biological neurons inspire SNNs, providing a blueprint for their design and understanding. Neurons communicate through electrical impulses known as action potentials or spikes, which typically last about ten milliseconds \cite{schuetze1983discovery}. Figure \ref{fig:neuron} illustrates a biological neuron's main components. Each neuron has three main parts: dendrites, an axon, and a soma (cell body) \cite{gerstner2002spiking}. Dendrites receive signals from the axons of other neurons. When these signals reach the dendrites, they alter the membrane potential of the soma. If this potential surpasses a threshold, the soma generates an action potential that travels along the axon, conducting it away from the soma to other neurons.

At the axon terminal, the arrival of an action potential prompts synaptic vesicles to release neurotransmitters into the synaptic cleft, the gap between the presynaptic and postsynaptic neurons \cite{jahn1994synaptic}. The synapse is the junction where the axon of one neuron communicates with the dendrite of another via neurotransmitters. These chemicals bind to receptors on the postsynaptic neuron's dendrites, influencing the likelihood of generating a new spike by opening or closing ion channels \cite{brown2008presynaptic}.

During an action potential, the neuron's membrane potential goes through several phases:
\begin{enumerate}
    \item Resting State: The membrane potential is about -70 mV.
    \item Depolarization: Sodium channels open, sodium ions flow in, and the membrane potential rises.
    \item Repolarization: Potassium channels open, potassium ions flow out, and the membrane potential drops back.
    \item Hyperpolarization: The membrane potential briefly drops below resting level, preventing immediate new spikes.
\end{enumerate}

After the neuron begins to fire, it enters a refractory period that can be divided into two phases: the absolute refractory period and the relative refractory period \cite{kandel2000principles}. The absolute refractory period corresponds to the depolarization and repolarization phases, during which the neuron is completely unable to fire another action potential, regardless of the strength of the stimulus. This period prevents neurons from firing too rapidly in succession. Following this, the neuron enters the relative refractory period, which coincides with the hyperpolarization phase. During this time, the neuron can fire again, but only in response to a much stronger stimulus than usual. The membrane potential is lower than its resting state, making it more difficult (but not impossible) for the neuron to reach the firing threshold.

\begin{table*}
\centering
\caption{Summary of Neuron Models}
\begin{tabular}{|p{2cm}|p{4cm}|p{2.5cm}|p{2.5cm}|}
\hline
Neuron Model                   & Description                                                                                                                                                                                                                                                    & Biological Plausibility                                                          & Computational Efficiency                                \\ \hline
Integrate-and-Fire (IF) \cite{lapicque1907louis}        & Simplest spiking neuron. Integrates incoming spikes until membrane potential reaches a threshold. Upon reaching the threshold, it fires and resets.                                                                                                            & Low                                                                              & High                                                    \\ \hline
Leaky Integrate-and-Fire (LIF) \cite{lapicque1907louis} & Extends IF model by including a leakage term that causes the membrane potential to decay towards a resting potential over time.                                                                                                                                & Moderate (Introduces decay towards resting potential)                            & Moderate (Slightly more complex than IF)                \\ \hline
Spike-Response Model (SRM) \cite{gerstner2002spiking}     & Generalizes the LIF model using filters instead of differential equations. Uses kernel functions for action potential and response to input. Integrates past spikes and external inputs for membrane potential evolution.                                      & High (Captures complex neural behaviors including refractoriness and adaptation) & Moderate (More complex integral formulation)            \\ \hline

Izhikevich Model  \cite{izhikevich2004model}             & Balances biological realism and computational efficiency. Uses two differential equations and reset conditions to model various neuron types. Can simulate different firing patterns by adjusting parameters.                                                  & High (Simulates diverse neural firing patterns)                                  & Moderate to High (Efficient yet biologically realistic) \\ \hline
Hodgkin-Huxley Model  \cite{hodgkin1952quantitative}         & Provides a highly detailed and biologically accurate representation of neuron dynamics by modeling ionic mechanisms. Describes conductance and voltage dynamics of ion channels through differential equations. Derived from experiments on squid giant axons. & Very High (Detailed biophysical properties)                                      & Low (Computationally intensive due to complexity)       \\ \hline
\end{tabular}
\label{tab:neuron_models}

\end{table*}

Synaptic strength, or weight, is influenced by neurotransmitter release, receptor availability, and signal resistance. Excitatory synapses increase membrane potential, while inhibitory synapses decrease it. The membrane potential returns to its resting state over time through exponential decay.

\subsection{Types of Artificial Neurons}
Understanding the types of neurons used in SNNs is essential for appreciating these models' versatility. Different neuron models offer varying degrees of biological plausibility and computational efficiency, which are crucial for designing and implementing effective NNs. Table \ref{tab:neuron_models} categorizes these neuron models based on criteria such as computational complexity, biological plausibility, and the types of neuronal behaviors they can simulate. Figure \ref{fig:neuronmodels} provides a comparative overview of select neuron models.

\subsubsection{Integrate-and-Fire (IF) Neurons}
The Integrate-and-Fire (IF) neuron model is among the most computationally efficient spiking neuron models, often used for its simplicity and ease of implementation. It functions by integrating incoming spikes, resulting in an accumulation of the membrane potential over time. Once the membrane potential reaches a predefined threshold, the neuron generates a spike (action potential) and subsequently resets the membrane potential to its resting value, typically zero. While the model captures the core dynamics of spiking behavior, it omits key aspects of biological realism, such as the refractory period and more complex synaptic dynamics \cite{lapicque1907louis}.

The dynamics of the IF neuron are governed by the following equation:

\begin{equation} \tau_m \frac{dV(t)}{dt} = RI(t) \end{equation}

where $\tau_m$ represents the membrane time constant, $R$ is the membrane resistance, and $I(t)$ is the input current. The model is defined by its simplicity, where membrane potential $V(t)$ evolves in response to synaptic input, yet its limitations must be considered when aiming for high-fidelity biological simulation or in tasks requiring more intricate neural dynamics.

\subsubsection{Leaky Integrate-and-Fire (LIF) Neurons}
\label{lifneuron}
The Leaky Integrate-and-Fire (LIF) neuron model extends the basic Integrate-and-Fire (IF) model by introducing a leakage term, which causes the membrane potential to decay over time toward its resting value. This leakage term more accurately represents the gradual dissipation of charge across a neuron's membrane, enhancing the biological realism of the model. The LIF model builds upon Louis Lapicque's foundational 1907 work on neural excitability \cite{lapicque1907louis}, in which Lapicque modeled nerve excitability based on a capacitor circuit, inspired by experiments on the sciatic nerve of frogs.

The LIF model, along with its variants, has gained significant popularity for studying the dynamics of spiking neural networks (SNNs) due to its balance between biological plausibility and computational efficiency \cite{dayan2005theoretical}. The membrane potential $V(t)$ of a LIF neuron is governed by the following differential equation:

\begin{equation} \tau_m \frac{dV(t)}{dt} = -V(t) + RI(t) \end{equation}

where $\tau_m$ represents the membrane time constant, $R$ is the membrane resistance, and $I(t)$ is the input current. When the membrane potential $V(t)$ reaches a predefined threshold, the neuron emits a spike and resets its potential, similar to the IF model.

The "leaky" term $-V(t)$ models the passive decay of membrane potential in the absence of input, reflecting the natural tendency of the neuron to lose charge over time. This decay prevents indefinite accumulation of potential, ensuring that the neuron's response to stimuli remains dynamic and transient, which is essential for realistic neural processing \cite{lapicque1907louis}.

For readers interested in hands-on implementation of neuron models like the LIF discussed here, the repository provides practical simulation examples that can enhance understanding and facilitate experimentation \cite{Iaboni2024EventSNN}.

\subsubsection{Spike-Response Model (SRM)}
The Spike-Response Model (SRM) generalizes the Leaky Integrate-and-Fire (LIF) model by using kernel functions to model neural dynamics, offering more flexibility than traditional differential equations. First introduced in \cite{gerstner2002spiking}, the SRM describes neural behavior with two primary kernel functions: $\eta$ and $\kappa$. The $\eta$ kernel accounts for the effects of past spikes, including refractoriness (the temporary period after a spike when a neuron is less likely to fire again) and after-potentials (the brief changes in membrane potential following a spike), while the $\kappa$ kernel models the neuron's response to external input currents.

In the SRM, the membrane potential $u(t)$ evolves based on both the history of past spikes and the external input current. At rest, the membrane potential equals the resting potential $u_{\text{rest}}$. When an external current $I_{\text{ext}}(t')$ is applied, the membrane potential deviates from $u_{\text{rest}}$ and gradually decays over time, governed by the kernel $\kappa(s)$, which captures the neuron's response to this input. This dynamic process is mathematically described through the convolution of the kernel $\kappa(s)$ and the input current.

The membrane potential $u(t)$ is influenced by two main components:

\begin{itemize}
    \item Past spikes: Represented by the sum $\sum_f \eta(t - t^{(f)})$, where $t^{(f)}$ are the times of previous spikes. The kernel $\eta(t)$ captures the effect of past spikes on the current membrane potential, including the refractoriness (less likelihood of firing) and after-potentials (temporary voltage shifts).

    \item External current: Represented by the convolution $\int_{0}^{\infty} \kappa(s) I_{\text{ext}}(t-s) ds$, which models the neuron's response to incoming stimuli. The kernel $\kappa(s)$ governs how the neuron integrates this external input over time.
\end{itemize}
The full equation governing the membrane potential in the SRM is:

\begin{equation} u(t) = \sum_f \eta(t - t^{(f)}) + \int_{0}^{\infty} \kappa(s) I_{\text{ext}}(t-s)  ds + u_{\text{rest}} \end{equation}

A spike is generated when the membrane potential $u(t)$ exceeds a dynamic threshold $\vartheta(t)$. This threshold $\vartheta(t)$ can vary depending on the recent spiking history of the neuron, allowing the SRM to capture phenomena such as refractoriness and spike-frequency adaptation. A spike is emitted when the following condition is met:

\begin{equation} t = t^{(f)} \iff u(t) = \vartheta(t) \quad \text{and} \quad \frac{d[u(t) - \vartheta(t)]}{dt} > 0 \end{equation}

This condition ensures that a spike is fired when the membrane potential surpasses the threshold while the potential is increasing.

Despite being more general than the LIF model, the SRM can reduce to the LIF model under specific kernel choices, making it a versatile framework for modeling a wide range of neural behaviors, including refractoriness, after-potentials, and adaptation to stimuli \cite{gerstner2002spiking}.

\begin{figure*}[t!]
\centering
\includegraphics[scale = 0.55]{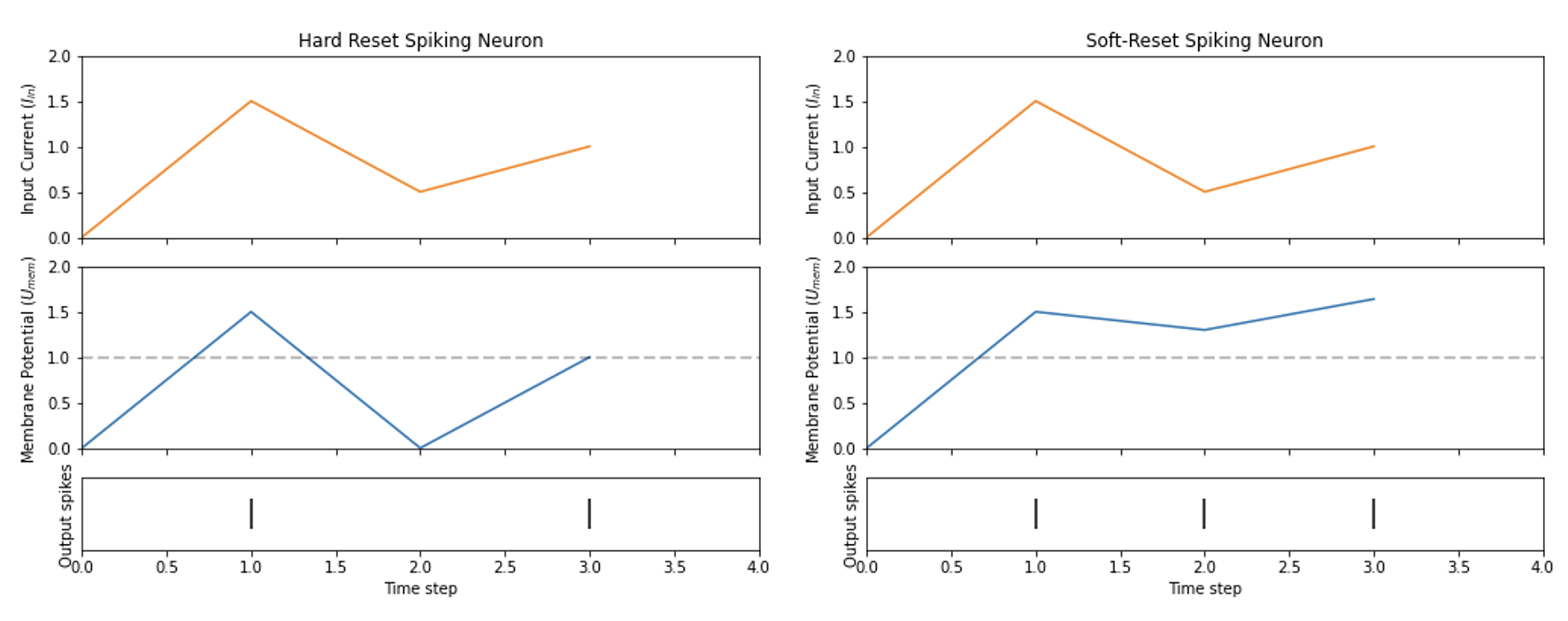}
\caption{At t=1, t=2, and t=3, input current values of 1.5, 0.5, and 1.0 units, respectively, are provided to both neurons. Hard Reset Neuron (left): This neuron resets its membrane potential to zero after each spike. Due to the hard reset mechanism, it produces only two output spikes, discarding the remaining potential after each spike reset. Soft Reset Neuron (right): This neuron reduces its membrane potential by the threshold amount after each spike, allowing the remaining potential to accumulate. Consequently, the soft reset neuron produces three output spikes.}
\label{fig:reset-types}
\end{figure*}

\subsubsection{Izhikevich Model}
The Izhikevich Model represents a significant advancement in computational neuroscience, balancing biological realism with computational efficiency. Proposed in \cite{izhikevich2004model}, the model captures the essential dynamics of neuronal behavior using two coupled differential equations:

\begin{equation} \frac{dv}{dt} = 0.04v^2 + 5v + 140 - u + I \end{equation}

\begin{equation} \frac{du}{dt} = a(bv - u) \end{equation}

with reset conditions when the membrane potential $v$ reaches or exceeds 30 mV:

\begin{equation} v \leftarrow c \end{equation}

\begin{equation} u \leftarrow u + d \end{equation}

In this model: \begin{itemize} \item $v$ represents the membrane potential of the neuron, \item $u$ is a recovery variable that governs the inactivation of sodium ion channels and activation of potassium ion channels, thereby regulating the neuron's excitability and firing dynamics. \end{itemize}

The parameters $a$, $b$, $c$, and $d$ play critical roles in shaping the neuron's behavior: \begin{itemize} \item $a$ controls the recovery time scale, determining how quickly the neuron recovers after firing. \item $b$ sets the sensitivity of the recovery variable $u$ to changes in membrane potential $v$. \item $c$ specifies the reset value of $v$ following a spike. \item $d$ defines the post-spike reset increment for the recovery variable $u$. \end{itemize}

These parameters allow the model to simulate a wide variety of neuron types, such as: \begin{itemize} \item Regular spiking (RS) neurons, which exhibit consistent low-frequency firing, \item Intrinsically bursting (IB) neurons, characterized by bursts of spikes followed by periods of quiescence (temporary inactivity), \item Fast-spiking (FS) neurons, which maintain high-frequency firing without adapting (sustaining a stable firing rate over prolonged periods). \end{itemize}

The Izhikevich model's ability to replicate complex neuronal firing patterns with minimal computational overhead makes it useful for simulating large-scale NNs, particularly in sensory systems such as visual processing, where diverse response patterns are critical to function \cite{sapounaki2019high}.

\subsubsection{Hodgkin-Huxley Model}
The Hodgkin-Huxley model provides a detailed representation of neuron dynamics by describing the ionic mechanisms underlying action potentials \cite{hodgkin1952quantitative}. Developed by Hodgkin and Huxley in 1952 through experiments on the giant squid’s axon, this model is based on the flow of sodium (Na\textsuperscript{+}) and potassium (K\textsuperscript{+}) ions through voltage-gated channels.

The model is governed by the following equation for the membrane potential $V$:

\begin{equation} C_m \frac{dV}{dt} = I_{\text{ext}} - \sum I_{\text{ion}} \end{equation}

where:

$C_m$ is the membrane capacitance,
$I_{\text{ext}}$ is the external current,
$I_{\text{ion}}$ represents the ionic currents through sodium, potassium, and leakage channels.
The ionic currents are described by the following equations:

\begin{equation} I_{\text{Na}} = g_{\text{Na}} m^3 h (V - E_{\text{Na}}) \end{equation}

\begin{equation} I_{\text{K}} = g_{\text{K}} n^4 (V - E_{\text{K}}) \end{equation}

\begin{equation} I_{\text{L}} = g_{\text{L}} (V - E_{\text{L}}) \end{equation}

where:

$g_{\text{Na}}$, $g_{\text{K}}$, $g_{\text{L}}$ are the conductances for sodium, potassium, and leakage channels,
$E_{\text{Na}}$, $E_{\text{K}}$, $E_{\text{L}}$ are the reversal potentials for these ions.
The gating variables $m$, $h$, and $n$, which govern the probability of the ion channels being open, follow voltage-dependent first-order differential equations.

The Hodgkin-Huxley model captures the essential biophysical properties of neurons but requires significant computational resources due to the complexity of the equations \cite{skocik2013capabilities}.

\subsection{Discussion about Neuron Models}
Several neuron models are pivotal to SNN research, balancing simplicity with biological plausibility. The Integrate-and-Fire (IF) and Leaky Integrate-and-Fire (LIF) models are widely used due to their efficiency and adequate realism. The LIF model’s leakage term adds biological accuracy without significant computational cost. The Spike-Response Model (SRM) extends LIF by using kernel functions to capture past spikes and input responses, providing greater flexibility. The Izhikevich model efficiently replicates diverse firing patterns, while the Hodgkin-Huxley model offers precise ion channel dynamics at a high computational cost. Despite the added sophistication of these latter models, IF and LIF remain popular due to their balance of efficiency and realism.

\subsubsection{Membrane Potential Reset Mechanisms}
Membrane potential reset mechanisms, though not types of neurons, are crucial for understanding how different neurons behave after firing. These mechanisms determine whether the neuron resets to zero (hard reset) or retains part of its residual membrane potential (soft reset), influencing information retention and response to subsequent inputs. While these reset mechanisms are most commonly applied to simpler spiking neuron models, such as LIF and IF neurons, they are not typically used in more biologically detailed models, such as Izhikevich or Hodgkin-Huxley neurons, which simulate membrane potential dynamics through continuous differential equations rather than discrete resets \cite{han2020rmp, niu2023event, izhikevich2004model, donati2019discrimination}. Figure \ref{fig:reset-types} illustrates the following two reset mechanisms. 

\begin{enumerate}
    \item {Hard-Reset Spiking Neuron \cite{cao2015spiking, diehl2015fast}}: 
    In the hard-reset mechanism, when the neuron's membrane potential exceeds the firing threshold, it is reset to zero. This approach is simple and computationally efficient, but it discards any residual potential above the firing threshold, leading to a loss of information. This can affect the accuracy of the neuron's response to continuous inputs.

    \item {Soft-Reset Spiking Neuron \cite{rueckauer2016theory, rueckauer2017conversion, han2020rmp}}: 
    In the soft-reset mechanism, the membrane potential is reset by an amount equal to the voltage threshold when it exceeds the firing threshold. This means that the neuron retains any residual potential above the threshold, preserving more information about the input signal. This mechanism, also known as "reset-by-subtraction" \cite{rueckauer2016theory, rueckauer2017conversion} or Residual Membrane Potential (RMP) \cite{han2020rmp}, allows for more accurate information encoding and better performance in inference tasks.
    
    Formally, the output firing rate, $f_{out}$ of the RMP can be described by
    \begin{equation}\label{eq:rmp}
         f_{out}=\left\{\begin{matrix}
            \left \lfloor \eta f_{in}N \right \rfloor N^{-1} & n \ge 0\\ 
            \approx \eta f_{in} & n \ge 0 \text{ and } n \gg 1
        \end{matrix}\right.
    \end{equation}
    where $f_{in} \le 1$, $f_{out} \le 1$, $\eta = \frac{V_{in}}{V_{th}}$ is the ratio between the average amplitude of the weighted input sum $V_{in}$ and firing threshold $V_{th}$, and $N$ is the inference latency.
\end{enumerate}

\begin{figure*}[t!]
    \centering
    \includegraphics[scale = 0.35]{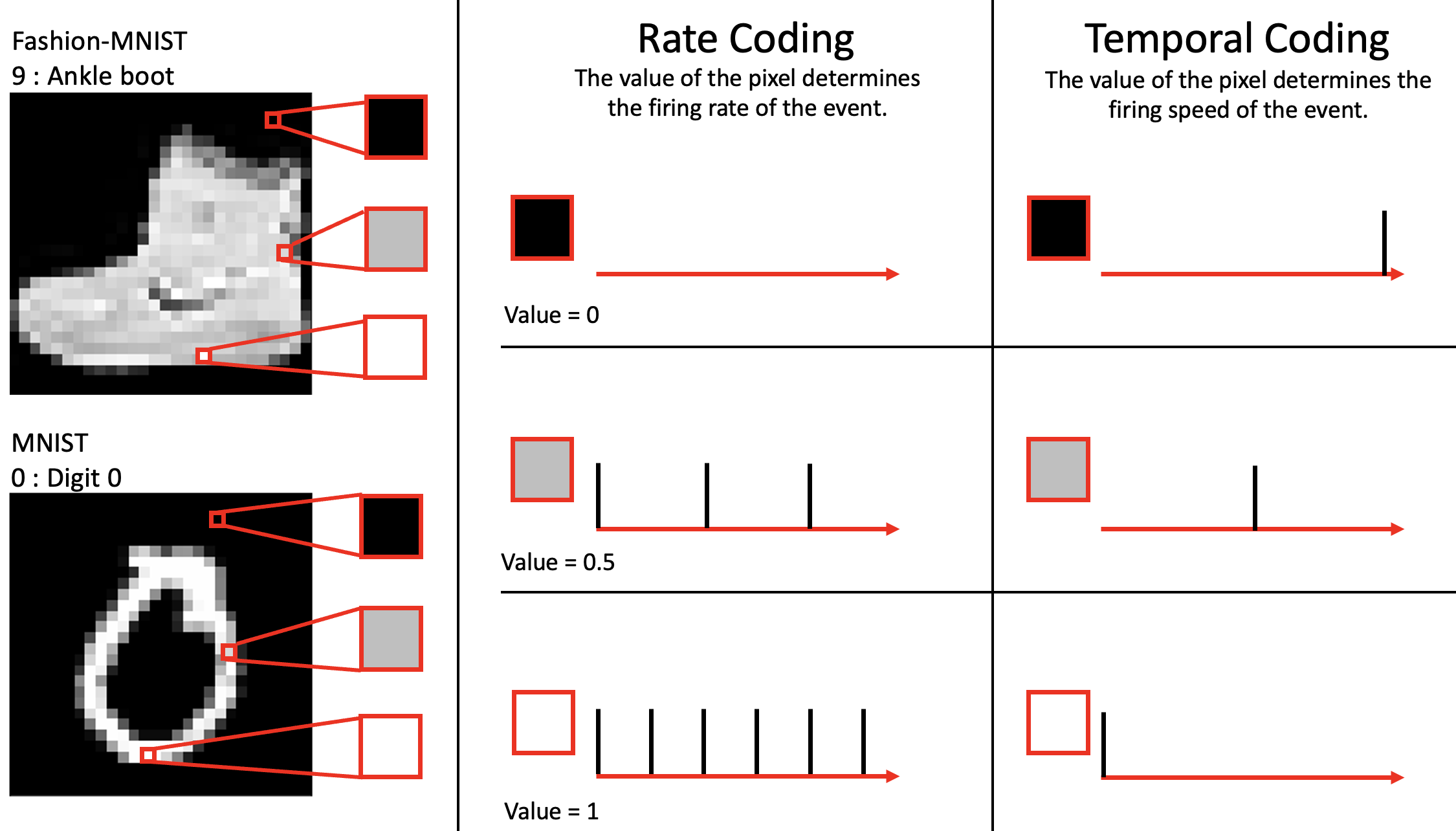}
    \caption{Rate Encoding (left): Pixel values are translated into spike rates, where higher pixel values result in higher firing rates. The number of spikes in a given time window corresponds to the pixel intensity. Temporal Encoding (Right): Pixel values are encoded as precise spike timings. Higher pixel values correspond to earlier spike times, whereas lower pixel values result in later spikes.}
    \label{fig:encoding}
\end{figure*}

\subsection{Neural Coding}
\label{sec:coding}

Neural coding is crucial for translating static input data into spikes and interpreting the spiking output. This section covers both input encoding and output decoding mechanisms.

\subsubsection{Input Encoding}
\label{poisson}
Input data is encoded into spike trains using various methods to represent different aspects of the stimulus.

\begin{enumerate}
    \item {Poisson Encoding \cite{diehl2015fast, diehl2015unsupervised, rueckauer2017conversion, stromatias2017event}:} This static encoding method converts each pixel of the input frame into a Poisson spike train, where the firing rate is proportional to the pixel intensity \cite{heeger2000poisson}. A bright pixel results in a higher firing rate, while a dim pixel results in a lower one. Poisson encoding introduces variability in spike timing, which can help capture the stochastic nature of neural responses.
    
    \item {Intensity-to-Latency Encoding \cite{masquelier2007unsupervised, kheradpisheh2018stdp}:} 
    Each pixel generates a single spike based on its intensity in this static encoding method. The intensity is transformed into a spike delay relative to a common reference time, with brighter pixels firing earlier. This reduces the number of events and speeds up processing, as only one spike per pixel is generated \cite{stromatias2017event}.
    
    \item {Delta Modulation:} This dynamic encoding method translates changes in input intensity into spikes, effectively capturing dynamic changes in the stimulus. It converts temporal variations in pixel intensity into a series of spikes, providing a sparse and efficient representation of the input. Event cameras and Dynamic Vision Sensors (DVS) work using this principle, detecting changes in the visual scene and generating spikes corresponding to these changes rather than capturing frame-based images \cite{yang2015dynamic}.
    
\end{enumerate}

Figure \ref{fig:encoding} illustrates the two static input encoding strategies.

\subsubsection{Output Decoding}
Output decoding mechanisms translate the spike train output back into a form that can be used for classification or other tasks.

\begin{enumerate}
    \item Rate Decoding: This method determines the predicted class based on the neuron with the highest firing rate or spike count. It is robust to noise and errors, aggregating information over multiple spikes \cite{gerstner2002spiking}.
    
    \item Temporal Decoding: This method predicts the output class based on the first neuron to spike. It is energy-efficient and fast, as it relies on the timing of a single spike rather than the total number of spikes. This method is advantageous for hardware implementations because it reduces the number of memory accesses and the overall computational load \cite{zhou2021temporal}.

    \item Population Coding: This approach uses multiple neurons representing each class and applies rate or temporal decoding methods to aggregate their activity. It enhances robustness and accuracy by leveraging the collective activity of a group of neurons \cite{pan2019neural}.
\end{enumerate}

\subsubsection{Discussion}
Neural coding in SNNs is crucial for converting input data into spikes and decoding these spikes for interpretation.
Input encoding methods, such as Poisson, intensity-to-latency, and delta modulation, each offer unique advantages. While Poisson encoding introduces variability by converting pixel intensities into Poisson-distributed spikes, intensity-to-latency encoding simplifies processing by generating a single spike per pixel based on intensity. Delta modulation stands out as the method used by neuromorphic sensors, such as event cameras, which are valuable in machine vision \cite{lichtensteiner2008128x128, posch2010qvga, brandli2014240, son20174, gallego2020event, he2024microsaccade}. This technique focuses on capturing dynamic changes in stimuli, providing a sparse and efficient input representation. Event cameras using delta modulation detect changes in the visual scene and generate spikes corresponding to these changes rather than capturing static frames.
Output decoding methods translate spike trains into usable information for tasks like classification. Rate decoding, which aggregates information over multiple spikes, is robust against noise. Temporal decoding predicts outputs based on the first neuron to spike, offering speed and energy efficiency. Population coding combines the activities of multiple neurons using either decoding method.

\begin{table*}[t!]
\centering
\caption{Prominent SNN Datasets}
\label{tab:snn_datasets}
\begin{tabular}{|p{2cm}|p{3cm}|p{9cm}|}
\hline
\multirow{5}{2cm}{Static Datasets 
} & \textbf{MNIST} \cite{mnist} & Collection of handwritten digits (0-9) with 60,000 training images and 10,000 testing images, normalized and centered on a 28 $\times$ 28 pixel greyscale image. \\ \cline{2-3}
 & \textbf{Fashion MNIST} \cite{fashion_mnist} & Drop-in replacement for MNIST with real-world image content, same size, and data amount, containing 10 classes of clothing articles. \\ \cline{2-3}
 & \textbf{Caltech101} \cite{Caltech101} & Contains 9,146 images of objects in 101 categories and one background category, with most categories having around 50 images. Images are approximately 300 $\times$ 200 pixels. \\ \cline{2-3}
 & \textbf{CIFAR10} \cite{cifar} & Consists of 60,000 32 $\times$ 32 pixel color images in 10 classes of real-world objects, with 50,000 training and 10,000 testing images. \\ \cline{2-3}
 & \textbf{ImageNet} \cite{imagenet} & Database of over 14 million annotated images, with a subset of around 1 million images provided with bounding boxes. \\ \hline
\multirow{2}{2cm}{Neuromorphic Converted Datasets 
} & \textbf{N-MNIST and N-Caltech101} \cite{nmnist} & Event camera on a motorized platform to record saccade-like motions viewing MNIST/Caltech101 images on an LCD monitor. \\ \cline{2-3}
 & \textbf{MNIST-DVS, MNIST-FLASH-DVS, CIFAR10-DVS} \cite{mnist-dvs} & Fixed event camera with digitally moving frame-based imagery or flashing pixels. \\ \hline
\multirow{16}{2cm}{Neuromorphic Captured Datasets 
} 
& \textbf{Poker-DVS} \cite{poker-dvs} & 131 poker pip symbols from 3 separate DVS recordings of rapidly browsing poker cards. \\ \cline{2-3}
 & \textbf{Postures-DVS} \cite{postures-dvs} & Recordings of two individuals adopting 6 different body postures. \\ \cline{2-3}
 & \textbf{PAFBenchmark} \cite{miao2019neuromorphic} & Pedestrian detection dataset with 4670 frame images and bounding box annotations for traffic surveillance scenarios. \\ \cline{2-3}
 & \textbf{N-Cars} \cite{sironi2018hats}  & Large event-based dataset for car classification with 12,336 car samples and 11,693 non-car samples. \\ \cline{2-3}
 & \textbf{Gen1 Automotive} \cite{de2020large} & Contains 39 hours of open road and various driving scenarios with manual bounding box annotations for pedestrians and cars. \\ \cline{2-3}
 & \textbf{1 Megapixel Automotive} \cite{perot2020learning} & 14 hours of annotated vehicles and pedestrians in roadway scenes, with the highest resolution for neuromorphic object detection. \\ \cline{2-3}
 & \textbf{EV-IMO} \cite{mitrokhin2019ev} & 32 minutes of multiple objects moving independently in an indoor setting, with annotations for pixel-wise masks, depth maps, and camera trajectories. \\ \cline{2-3}
 & \textbf{PKU-DAVIS-SOD} \cite{li2023sodformer} & 220 event and frame-based driving sequences with labels, focusing on spatiotemporal Transformer architecture evaluation. \\ \cline{2-3}
 & \textbf{DDD17} \cite{binas2017ddd17} & 12 hours of frame and event data for driving scenarios with manual annotations for synchronized streams. \\ \cline{2-3}
 & \textbf{Scenes-DVS} \cite{scenes-dvs}  & Recordings of long hikes in urban environments, split into small sequences. \\ \cline{2-3}
 & \textbf{DVS-Gesture} \cite{Amir2017} & 11 hand gestures from 29 subjects under different illumination conditions. \\ \cline{2-3}
 & \textbf{SL-ANIMALS-DVS} \cite{sl-animals-dvs} & Recordings of humans performing sign language gestures of various animals. \\ \cline{2-3}
 & \textbf{Pred18} \cite{pred18} & Recordings from a DAVIS240 camera on a robot chasing another robot. \\ \cline{2-3}
& \textbf{NU-AIR} \cite{iaboni2023nu} & Recordings of pedestrians and vehicles in urban environments from an aerial perspective. \\ \cline{2-3}
 & \textbf{Indoor Ground Robots} \cite{iaboni2021event} & Recordings of indoor ground robots tracing shapes from a god's eye view. \\ \cline{2-3}
 & \textbf{Indoor Quadrotors} \cite{iaboni2022event} & Approximately 10,000 samples of quadrotors flying variable paths in various conditions. \\ \hline
\multirow{7}{2cm}{Simulated Data 
} & \textbf{ESIM} \cite{Rebecq2018} & Accurate simulation of camera motion in 3D scenes with events, images, inertial measurements, and full ground truth information. \\ \cline{2-3}
 & \textbf{InteriorNet} \cite{li2018} & End-to-end pipeline for large-scale interior scene understanding with events, IMU, and camera trajectories. \\ \cline{2-3}
 & \textbf{DVS Gazebo Plugin} \cite{Kaiser2016} & Event camera simulation within the Gazebo robot simulator for vehicle and environment representations. \\ \cline{2-3}
 & \textbf{PIX2NVS} \cite{bi2017pix2nvs} & Converts pixel frames to brightness spike events as generated by event-based sensors. \\ \cline{2-3}
 & \textbf{pyDVS} \cite{garcia2016pydvs} & Extensible behavioral emulator of an event camera using a conventional digital camera. \\ \cline{2-3}
 & \textbf{Video-to-Events} \cite{Gehrig2020}& Converts any existing video dataset to synthetic event data. \\ \cline{2-3}
 & \textbf{V2E} \cite{Hu2021} & Synthesizes realistic event camera data from conventional frame-based video. \\ \hline
\end{tabular}
\end{table*}

\section{Prominent SNN Datasets}
\label{sec:prominent_datasets}

This section categorizes datasets used in event-based research into four groups: static datasets, neuromorphic converted datasets, neuromorphic captured datasets, and simulated data. It summarizes key datasets in each category, discussing their relevance to SNN applications, data generation methods, and any associated limitations or advantages.
Table \ref{tab:snn_datasets} summarizes the most widely used SNN datasets along with their salient features, presented in order of prevalence in the literature.

\subsection{Randomness in Event Camera Data}
Event cameras are designed to capture asynchronous intensity changes, yet inherent randomness in event generation can degrade performance in detection and classification tasks. Several factors contribute to this randomness:

\begin{itemize} \item Background Activity (BA): Events may be erroneously generated without any corresponding intensity change, often due to sensor noise. This spurious background activity can lead to false positives, degrading the performance of downstream algorithms and increasing bandwidth consumption. 
\item Missed Events: Despite actual variations in pixel intensity, certain changes may not trigger an event, leading to missed detections of critical scene information. This loss of data is particularly detrimental in dynamic environments where timely event capture is essential. 
\item Unpredictable Event Generation: The irregular and non-deterministic nature of event occurrences complicates the modeling of consistent spatio-temporal patterns, posing challenges for real-time processing and accurate learning in SNNs. 
\item Event Rate Variability: Although the number of events generally scales with the magnitude of intensity change. For example, high contrast generates more events than low contrast; the exact number of events for a given change magnitude remains unpredictable, introducing variability in data representation. \end{itemize}

\subsection{Static Datasets (Conversion of frame-based imagery to event-based data via encoding)}

The number of diverse, frame-based datasets with annotations available online is continually growing \cite{mnist, fashion_mnist, Caltech101, cifar, imagenet}. The SNN community has leveraged these open-source frame-based datasets by converting them to event datasets using rate encoding and temporal encoding methods \cite{shrestha2018slayer,liu2020effective, patino2020event, lee2016training, wu2018spatio, wu2019direct, xiao2019event, cannici2019attention, zhu2021efficient, stromatias2017event, li2018deep, fang2021deep}. 

Commonly reported frame-based datasets that are encoded for event-based tasks are:
\begin{itemize}
    \item {\bf MNIST \cite{mnist}:} This collection of handwritten digits from 0 to 9 has 60,000 images in the training set and 10,000 examples in the testing set. The digits are size-normalized and centered on a 28 $\times$ 28-pixel greyscale image. 
    
    \item {\bf Fashion MNIST \cite{fashion_mnist}:} This set is intended to replace the MNIST dataset with real-world image content, making it more relevant to machine vision tasks. Fashion-MNIST shares the exact image size and amount of training and testing data with MNIST and contains ten classes of standard clothing articles.
    
    \item {\bf Caltech101 \cite{Caltech101}:} This set contains 9,146 images of objects belonging to 101 distinct categories and one background (clutter) category. Most categories have approximately 50 images. The size of each image is approximately 300 $\times$ 200 pixels. Each image is provided with annotations describing the outlines of each object.
    
    \item {\bf CIFAR10 \cite{cifar}:} This set consists of 60,000 32 $\times$ 32-pixel color images representing ten classes of real-world objects, birds, and animals with 6000 images per class. There are 50,000 training images and 10,000 testing images. The classes are mutually exclusive, with no overlap between images.
    
    \item {\bf ImageNet \cite{imagenet}:} ImageNet is a database of over 14 million images annotated by hand. A subset of the dataset, approximately 1 million images, is provided with bounding boxes. ImageNet is one of the essential datasets that has enabled critical advancements in machine vision and deep learning research \cite{krizhevsky2017imagenet}. 
\end{itemize}

\subsubsection{Discussion}
Rate and temporal coding of open-source, pre-annotated frame-based datasets reduce the time required for testing SNN applications. This facilitates rapid benchmarking of SNN algorithms and accuracy comparisons \cite{lamba2019spiking}. However, the encoding process can only capture spatial information from frames, as frame-based cameras retain no temporal information between successive frames. Thus, the natural temporal dynamics of the recorded scene are unavailable for updating synaptic weights during the SNN training process \cite{orchard2015converting}. Furthermore, the inherent noise characteristics specific to event cameras — such as background activity, missed events, unpredictable event generation, and event rate variability — are absent in frame-based imagery \cite{dilmaghani2023control}. As a result, the encoded datasets fail to realistically approximate the noise and variability encountered in event-based recordings, limiting the utility of frame-based datasets for training and evaluating SNNs in real-world, event-driven applications.

\subsection{Neuromorphic Converted Datasets (Recording frame-based imagery using an event camera)}

Another approach to generating event data involves recording a screen (computer monitor or TV) using an event camera mounted on a tripod \cite{orchard2015converting, mnist-dvs, cifar10-dvs}. Displaying the image statically on the screen is insufficient since event-based sensors only respond to changes in the scene, so some amount of motion is introduced by either moving the camera or the image on the screen \cite{nmnist, mnist-dvs, orchard2015converting, mnist-dvs, cifar10-dvs}. 
\begin{itemize}
    \item The authors of N-MNIST and N-Caltech101 datasets mounted an event camera on a motorized pan-tilt platform and programmed the platform to perform saccade-like motions as the event camera viewed MNIST/Caltech101 images on an LCD monitor \cite{nmnist}.
    
    \item Alternatively, the authors of MNIST-DVS \cite{mnist-dvs}, MNIST-FLASH-DVS, and CIFAR10-DVS \cite{cifar10-dvs} datasets opted to keep the event camera fixed on a tripod and instead move the frame-based imagery digitally on the screen. Digitally moving imagery involves translating the pixels on the screen by some distance or flashing the pixels rapidly \cite{mnist-dvs}.
\end{itemize}

\subsubsection{Discussion}
Automating the display on the screen and the corresponding motion substantially reduces the burden of generating arbitrary new event datasets with an event camera. Such automated approaches have been termed as live-in-the-loop recording \cite{orchard2015converting}. The saccade-like movements of the sensor resemble the subconscious retinal movements observed in primates and humans, thus providing a biologically realistic event dataset \cite{engbert2006microsaccades}. In contrast to the encoding approach, this method presents naturally occurring event noise in the recordings. 

The main limitations of this approach are observed when the event camera remains stationary while the pixels on the screen are translated or flickered. Translating the pixels at a high screen refresh rate is sensed as discrete jumps of events by an event-based sensor, thus introducing unnatural artifacts in the recording \cite{orchard2015converting}. Flickering pixels on a screen to generate changes in brightness presents limited real-world relevance since natural changes in brightness are continuous.

\label{sec:native-event}
\subsection{Neuromorphic Captured Datasets (Native data capture using an event camera)} 

This approach involves using event cameras to capture real-world scenes. Cameras from manufacturers as Prophesee, iniVation, Samsung, Insightness and CelePixel have been used in the literature to capture footage of people, animals, objects, robots, and drones \cite{iaboni2022event, iaboni2021event, poker-dvs, postures-dvs, sironi2018hats, scenes-dvs, Amir2017, sl-animals-dvs, pred18}.

Public datasets created using this approach include:
\begin{itemize}
    \item {\bf Poker-DVS \cite{poker-dvs}:} consists of 131 poker pip symbols tracked and extracted from 3 separate DVS recordings while rapidly browsing poker cards.
    \item {\bf Postures-DVS \cite{postures-dvs}:} consists of recordings of two individuals adopting six different body postures classified as `bending’, `hand1’, `hand2’, `squat’, `swing’ and `stand’.
    
    \item {\bf PAFBenchmark \cite{miao2019neuromorphic}:} a pedestrian detection dataset contains 4670 frame images at 20ms time intervals with bounding box annotations of scenarios seen in traffic surveillance tasks such as pedestrian overlapping, occlusion, and collision.
    
    \item {\bf N-Cars \cite{sironi2018hats}:} a large real-world event-based dataset for car classification. It comprises 12,336 car samples and 11,693 non-car samples (background). The dataset is split into 7940 car and 7482 background training samples and 4396 car and 4211 background testing samples. Each example lasts 100 milliseconds.

    \item {\bf Gen1 Automotive \cite{de2020large}:} contains 39 hours of the open road and various driving scenarios ranging from urban, highway, suburbs, and countryside scenes. Manual bounding box annotations are provided for two classes of detection: pedestrians and cars. The capture resolution of this set was 304 $\times$ 240. This study has not evaluated the data for classification or detection accuracy.

    \item {\bf The 1 Megapixel Automotive \cite{perot2020learning}:} includes 14 hours of annotated vehicles and pedestrians captured within roadway scenes. This set is the highest resolution neuromorphic object detection dataset from a vehicle-mounted perspective to date, with a capture resolution of 1280 × 1024.

    \item {\bf EV-IMO \cite{mitrokhin2019ev}:} contains 32 minutes of multiple objects moving independently against various backgrounds in an indoor setting. The imagery was obtained at a capture speed of 200 frames per second. Annotations were provided as pixel-wise masks, depth maps, and camera trajectories.

    \item {\bf The PKU-DAVIS-SOD dataset \cite{li2023sodformer}:} contains 220 event and frame-based driving sequences and labels at a frequency of 25 Hz and a resolution of 346 $\times$ 260. The set has 276,000 labeled frames and 1,080,100 bounding boxes. A spatiotemporal Transformer architecture was used to evaluate the proposed dataset for three classes of detection.

    \item {\bf The DDD17 dataset \cite{binas2017ddd17}:} focuses on driving scenarios to provide over 400GB and 12 hours of frame and event data captured using a 346×260 pixel DAVIS sensor. Manual annotations are supplied for synchronized frame and event streams. The dataset was tested on an object detection task using a fine-tuned YOLOv3 model.
    
    \item {\bf Scenes-DVS \cite{scenes-dvs}:} consists of recordings of long hikes in urban environments. The sequences exceeded 15 minutes long and included indoor and outdoor environments. The records were split into small sequences of 50ms. 
    
    \item {\bf DVS-Gesture \cite{Amir2017}:} This set contains 11 hand gestures from 29 subjects under three illumination conditions.

    \item The authors of \cite{el2022high} proposed a dataset combining event- and frame-based traffic data. Using a DAVIS 240c camera operating at a resolution of 240 $\times$ 180 pixels and 24 Hz, two distinct scenes are captured with 9891 vehicle annotations. A static camera was mounted to the side of a building to observe a street scene. The fidelity of the dataset was evaluated on a multi-object tracking task.
    
    \item {\bf SL-ANIMALS-DVS \cite{sl-animals-dvs}:} consists of recordings of humans performing sign language gestures of various animals. The dataset has about 1100 samples of 58 subjects performing 19 sign language gestures.
    
    \item {\bf Pred18 \cite{pred18}:} contains recordings from a DAVIS240 camera mounted on a computer-controlled robot (the predator) that chases and attempts to capture another human-controlled robot (the prey).

    \item {\bf NU-AIR \cite{iaboni2023nu}:} A neuromorphic aerial dataset featuring over 70 minutes of event data captured by a 640 $\times$ 480 px event-based camera on a quadrotor, covering urban scenes like campuses, pathways, and intersections under varying lighting. It includes over 93,000 bounding box annotations for pedestrians and vehicles at 30 Hz.
    
    \item {\bf Indoor Ground Robots \cite{iaboni2021event}:} This set contains multiple indoor ground robots tracing circular and square shapes, observed from a God's eye view by the sensor.
    
   \item {\bf Indoor Quadrotors \cite{iaboni2022event}:} This dataset comprises approximately 10,000 samples of one to three quadrotors following variable paths in an indoor arena, captured by a moving event camera under low-light conditions, as well as outdoor flights.

\end{itemize}

\subsubsection{Discussion}
Event cameras use recording parameters such as contrast sensitivity thresholds, low- and high-pass filters, photo receptor bandwidth, and pixel refractory period settings that can be tuned to capture relevant data \cite{muglikar2021calibrate}. These parameters allow users to directly control the number of events output by changes in illumination, the maximum rate of illumination change, and the rate of noise background events detected by the camera. 

The main limitations of this approach are the costs associated with capturing native event camera datasets and the significant investments of time and resources required. Creating arbitrary footage with annotations using event cameras requires significant investments of time and resources.

The repository also provides links for accessing common neuromorphic-captured datasets used in SNN research, making it easier for researchers to quickly start experimenting with SNN models \cite{Iaboni2024EventSNN}.

\subsection{Simulated data generated using via rendering application or interpolated video frames}

Event data can be generated using simulation software to evaluate arbitrary camera trajectories in 3D scenes or input video from a conventional digital camera \cite{Rebecq2018, Kaiser2016, li2018, garcia2016pydvs, bi2017pix2nvs, Gehrig2020, Hu2021}. Event simulation techniques are grouped into the following two categories:

\begin{table*}[t!]
\centering
\caption{Summary of SNN Simulation Frameworks}
\begin{tabular}{|p{2cm}|p{3cm}|p{2cm}|p{2cm}|p{2cm}|p{1.8cm}|p{1.8cm}|}
\hline
\textbf{Framework} & \textbf{Key Contributions} & \textbf{Neuron Models} & \textbf{Learning Rules} & \textbf{Compatibility} & \textbf{Strengths} & \textbf{Limitations} \\
\hline
NEST \cite{diesmann2001nest} & Modular, object-oriented architecture; focuses on combining biological realism and computational efficiency & Leaky Integrate-and-Fire (LIF), Hodgkin-Huxley & Iterative, collaborative development & Open-source, platform-independent & High biological realism, modular & Computational efficiency challenges \\
\hline
NEURON \cite{hines1997neuron} & Models with detailed anatomical and biophysical properties; supports electrical and chemical signaling & Hodgkin-Huxley, Integrate-and-Fire (IF) & Density mechanisms, point processes & High flexibility in neuron modeling & Detailed modeling, parameter optimization tools & Requires significant computational resources \\
\hline
Brian \cite{Brian} & Clock-driven simulator in Python; emphasizes ease of use and flexibility & LIF, Hodgkin-Huxley & Direct specification, random/all-to-all connectivity patterns & Integration with Python scientific libraries & Ease of use, highly portable & Limited scalability for large networks \\
\hline
Nengo \cite{bekolay2014nengo} & Based on NEF; improved simulation speed with Python & LIF, Izhikevich & Principles of NEF & Python, NumPy, Matplotlib, IPython & Fast simulation, versatile modeling & Limited to NEF principles \\
\hline
SNNToolbox \cite{rueckauer2017conversion} & Automates conversion of ANNs to SNNs; supports multiple frameworks & LIF, IF & Built-in simulation tools; supports spiking hardware platforms & Integrated with Keras, Lasagne, Caffe & Conversion automation, hardware deployment & Conversion accuracy depends on source models \\
\hline
BindsNET \cite{hazan2018bindsnet} & Rapid prototyping and simulation of SNNs for machine learning & IF, LIF & Hebbian learning, STDP, reward-modulated STDP & PyTorch & Flexibility, GPU support & Requires familiarity with PyTorch \\
\hline
snnTorch \cite{eshraghian2023training} & Integrates spiking neurons in PyTorch; uses BPTT with surrogate gradients & LIF, second-order LIF, recurrent LIF, LSTM & Surrogate gradient functions & PyTorch & Seamless PyTorch integration, multiple neuron models & Requires understanding of PyTorch's gradient-based methods \\
\hline
SpikingJelly \cite{fang2023spikingjelly} & Comprehensive SNN framework; supports CUDA acceleration & IF, LIF & Surrogate gradient method, ANN-to-SNN conversion & PyTorch, Intel Loihi, Tianjic & High-speed training, neuromorphic support & Complexity of setup \\
\hline
NxTF \cite{rueckauer2022nxtf} & Interface and compiler for deep SNNs on Loihi & LIF, IF & Surrogate gradient functions & Keras, Intel Loihi & Optimized resource utilization on Loihi & Specific to Intel Loihi \\
\hline
\end{tabular}
\label{tab:snn_simulation_frameworks}
\end{table*}

\noindent {\bf 1. Using simulated 3D renderings} 

Event camera simulators such as ESIM, DAVIS Simulator, InteriorNet, and DVS Gazebo Plugin use 3D rendering engines to construct photo-realistic 3D scenes, which can then be used to compute the pixel luminosity differences between successively rendered frames \cite{Rebecq2018, mueggler2017event, li2018, Kaiser2016}. When the magnitude of the luminosity difference exceeds a pre-defined firing threshold, an event is generated in the simulated camera. Of note in this category are the following open-source simulators: 

\begin{itemize}
    \item {\bf ESIM \cite{Rebecq2018}:} This simulator provides accurate simulation of arbitrary camera motion in 3D scenes while providing events, standard images, and inertial measurements, with full ground truth information including camera pose, velocity, as well as depth and optical flow maps. ESIM improves upon the previously released DAVIS Simulator \cite{mueggler2017event}.
    
    \item {\bf InteriorNet \cite{li2018}:} This simulator provides an end-to-end pipeline to render an RGB-D-inertial benchmark for large scale interior scene understanding and mapping. Ground truth IMU, events, and monocular or stereo camera trajectories are supported. InteriorNet was used to create computer-aided design (CAD) models of furniture items with pixel-wise annotations and accurate dimensions, resulting in a large-scale event dataset of over five million images \cite{li2018}.

    \item {\bf DVS Gazebo \cite{Kaiser2016}:} This is a plugin for event camera simulation within the Gazebo robot simulator \cite{koenig2004design}. Gazebo is a physics simulator for the Robot Operating System (ROS) and simulates representations of vehicles and the environment. DVS Gazebo is used as a drop-in replacement for the normal Gazebo camera plugins.
\end{itemize}

\noindent {\bf 2. Using a video sequence captured from a digital camera}
Event camera simulators such as PIX2NVS, pyDVS, Video-to-Events, and V2E use sequential video frames to generate events from pre-recorded video or live input from a conventional digital camera. These simulators compute whether pixel-wise brightness changes exceed a pre-defined firing threshold $(V_{th})$ to record event timestamps at pixel locations \cite{garcia2016pydvs, bi2017pix2nvs, Gehrig2020, Hu2021}. Video sequences can only provide intensity measurements at fixed and low temporal resolutions on the order of milliseconds. Therefore, linear interpolation approximates the pixel intensity signal between video frames \cite{Gehrig2020}. The subproblem of reconstructing frames at arbitrary temporal resolutions uses bidirectional optical flow to compute intermediate frames between pairs of video frames \cite{jiang2018super}. Notable simulators in this category are:

\begin{itemize}
    \item {\bf PIX2NVS \cite{bi2017pix2nvs}:} This is a simulator for converting pixel frames to brightness spike events as generated by event-based sensors.
    \item {\bf pyDVS \cite{garcia2016pydvs}:} This extensible behavioral emulator of an event camera uses a conventional digital camera as a sensor.
    \item {\bf Video-to-Events \cite{Gehrig2020}:} This simulator converts any existing video dataset recorded with conventional cameras to synthetic event data.
    \item {\bf V2E \cite{Hu2021}:} This simulator synthesizes realistic event camera data from any real (or synthetic) conventional frame-based video.
\end{itemize}

\subsubsection{Discussion}
Synthetic imagery does not require manual ground truth annotations and measurements, allowing for the rapid generation of large event datasets. This approach also allows for the exploration of critical edge cases and failure situations that would be impractical or impossible to test manually, as was noted by authors in \cite{afshar2020event} who studied satellite-collision scenarios. Additionally, renderings for arbitrary 3D scenes may be created at arbitrary temporal resolutions using this approach \cite{joubert2021event}. The main limitation of this approach is noise simulation. Noise effects such as spatially and temporally varying contrast thresholds due to electronic noise or limited bandwidth of the event pixels are not accounted for in rendering applications.

\section{Implementation Mediums}
\label{sec:implementation_medium}
This section categorizes SNN implementations into two primary mediums: software simulation libraries and neuromorphic hardware platforms. It reviews various software libraries used for SNN simulation, their capabilities, and key frameworks, while also detailing hardware platforms designed for neuromorphic computing, focusing on their architecture and real-world applications.
Each medium plays a crucial role in the advancement of SNN research, enabling experimentation, optimization, and real-world deployment.

\subsection{Simulation of SNNs}
Software simulation libraries are essential for developing and testing SNNs, providing an accessible environment for researchers to explore various network architectures, neuron models, and learning rules. Several libraries have been developed to facilitate the simulation of SNNs. Table \ref{tab:snn_simulation_frameworks} summarizes the key contributions, neuron models, learning rules, compatibility, strengths, and limitations of various SNN simulation frameworks.

The following three papers, while foundational in their respective contributions, represent the earlier stages of development in the realm of SNN simulation. These works laid the groundwork for understanding and leveraging the dynamics of spiking neurons, but they faced limitations in computational efficiency and scalability.

\begin{itemize}
    \item In \cite{diesmann2001nest}, the NEST framework is presented for simulating large, structured neuronal systems with a focus on combining biological realism and computational efficiency. The paper discusses the modular, object-oriented architecture of NEST, which includes a simulation kernel, a simulation language interpreter (SLI), and various auxiliary modules. The simulation kernel manages network structures and temporal updates, while SLI facilitates flexible experiment setups and data manipulation. The authors emphasize the importance of iterative and collaborative development, platform independence, and open-source principles in the success of NEST.
    
    \item In \cite{hines1997neuron}, the NEURON simulation environment is presented as a tool for creating and simulating biologically realistic models of neurons and neuronal networks. The environment addresses the complexities of electrical and chemical signaling in neurons, which span wide spatial and temporal scales and often lack analytical solutions. NEURON allows for the construction of models with detailed anatomical and biophysical properties. NEURON includes tailored functions for neurophysiological simulations, an efficient computational engine, and a separation of biological property specifications from numerical considerations. NEURON uses sections and segments to discretize neuron structures, employing various numerical integration methods such as backward Euler and Crank-Nicholson for stability and accuracy \cite{mascagni1990backward, mascagni1989numerical}. The environment supports density mechanisms and point processes for modeling distributed and localized sources of electrical and chemical signals, respectively. NEURON also includes sophisticated tools like the Function Fitter and Electrotonic Workbench for parameter optimization and analysis of electrotonic properties.
    
    \item In \cite{Brian}, the Brian simulator is presented as an environment to simulate spiking neurons. The authors emphasize that although many existing neural network simulators exist, researchers often prefer writing their own code in Matlab or C due to ease of implementation and flexibility. Brian addresses these issues by being a clock-driven simulator written entirely in Python, making it highly portable and easy to integrate with other Python scientific libraries. Brian supports various neuron models, including LIF and Hodgkin-Huxley type models, and allows for complex network connectivity through direct specification or using random and all-to-all connectivity patterns.
\end{itemize}

As the field matured, Python frameworks such as PyTorch and TensorFlow, along with more specialized libraries, emerged as key solutions for implementing and experimenting with SNNs \cite{paszke2017automatic, tensorflow2015-whitepaper}. Frameworks like NEST and Brian emphasize biological realism, making them well-suited for neuroscience research. Others, such as SNNToolbox, Norse, snnTorch, and SpikingJelly, prioritize computational efficiency and scalability, making them more applicable to machine learning tasks \cite{rueckauer2017conversion, Norse, fang2023spikingjelly, eshraghian2023training}. Norse and snnTorch, built on PyTorch, provide flexible, efficient SNN implementations for users familiar with deep learning workflows. This bifurcation enables researchers to select tools tailored to their needs: whether focused on biological plausibility or computational performance, while some frameworks, like Norse and SpikingJelly, offer versatility across both neuroscience and machine learning domains. These simulators are described in the following: 

\begin{itemize}
\item Nengo \cite{bekolay2014nengo} is a software tool for building and simulating neuron models using the Neural Engineering Framework (NEF) \cite{stewart2012technical}, which provides principles for representing, transforming, and modeling neural population dynamics for cognitive functions. Nengo 1.0, originally implemented in Java, was rewritten in Python as Nengo 2.0, enhancing integration with scientific libraries like NumPy and Matplotlib, and improving usability and simulation speed by up to 50 times over Nengo 1.4. Nengo's core components—Ensembles (neuron groups representing vectors), Nodes (non-neural inputs or computations), Connections (synaptic transformations), Probes (data recording tools), Networks (collections of connected objects), and Models (simulation containers)—are based on NEF principles. The transformation principle enables efficient computation of nonlinear functions, while the dynamics principle supports recurrent connections for complex behaviors like attractor dynamics. Nengo decouples model creation from simulation, allowing models to run on multiple simulators, including an OpenCL-optimized version that leverages GPUs and multicore CPUs for faster execution. Its versatility is demonstrated in applications such as communication channels, Lorenz attractor networks, and circular convolution functions, making it a powerful tool for neuroscience research and large-scale neural simulations.

\item In \cite{rueckauer2017conversion}, the SNNToolbox was created as part of a study to automate the conversion of ANNs into SNNs. It supports models from Keras, Lasagne, and Caffe frameworks and includes built-in simulation tools for evaluating spiking models \cite{chollet2015keras, lasagne, jia2014caffe}. This toolbox allows integration and deployment of converted SNNs on spiking hardware platforms such as TrueNorth and SpiNNaker \cite{akopyan2015truenorth, furber2012overview}. These hardware platforms are discussed in the next subsection. 
    
\item In \cite{lava}, Lava is described as an emerging open-source framework developed by Intel's Neuromorphic Computing Lab. It is designed to address key challenges in neuromorphic hardware, software, and application development through asynchronous event-based processing for iterative refinement of abstract computational processes into architecture-specific implementations and supporting cross-platform execution on both neuromorphic and conventional hardware. The foundational building blocks of Lava are Processes, which are stateful objects with internal variables and ports for message-based communication. Lava promotes modularity and composability, enabling the creation of large-scale parallel applications, and supports cross-platform execution through a flexible compiler and runtime environment. The software stack includes a Communicating Sequential Process API, a compiler, and a runtime, forming the Magma layer, with additional libraries for deep learning (lava-dl), optimization (lava-optim), and dynamic neural fields (lava-dnf).

\item In \cite{hazan2018bindsnet}, the authors present BindsNET, a Python-based library designed to enable the rapid prototyping and simulation of SNNs for ML and RL applications. Built on PyTorch, BindsNET supports GPU acceleration, making it suitable for large-scale simulations. BindsNET includes various neuron models, such as  IF and LIF neurons, and implements learning rules like Hebbian learning, STDP, and reward-modulated STDP. These learning rules will be discussed in Section \ref{sec:learning_rules}. The library's modular design includes components for network construction, encoding of input data into spike trains, and visualization tools for monitoring simulation progress. BindsNET emphasizes flexibility, allowing users to define custom neuron models, connections, and learning rules. The authors provide examples of unsupervised learning with MNIST, supervised learning with Fashion-MNIST, reinforcement learning with Atari games, and reservoir computing, demonstrating the library's versatility. Benchmark comparisons show that BindsNET performs competitively with other SNN simulators, especially in terms of rapid prototyping on CPUs and GPUs, without the need for intermediate compilation steps. The authors highlight ongoing developments and future directions, including the addition of more neuron types, learning rules, and optimization techniques.

    \begin{table*}[t]
\centering
\caption{Summary of Neuromorphic Hardware Platforms}
\begin{tabular}{| m{2cm} | m{1.3cm} | m{2cm} | m{2.2cm} | m{1.5cm} |}
\hline
\textbf{Platform} & \textbf{Technology} & \textbf{Implementation Category} & \textbf{Key Features} & \textbf{Applications} \\
\hline
TrueNorth \cite{akopyan2015truenorth} & 28nm CMOS & Digital ASIC & 1 million neurons, 256 million synapses, LIF model & Character recognition, anomaly detection \\
\hline
Loihi \cite{davies2018loihi} & 14nm CMOS & Digital ASIC & 130,000 neurons, 130 million synapses, LIF model & SLAM, neural fields, graph search \\
\hline
BrainScaleS \cite{pehle2022brainscales} & 180nm CMOS & Analog/Mixed-Signal & High-speed AdExp model, wafer-scale integration & Large-scale neural simulations \\
\hline
Tianjic \cite{deng2020tianjic} & 28nm CMOS & Hybrid (ANN/SNN) & 156 cores, configurable for ANN or SNN & Unmanned bicycle navigation \\
\hline
SpiNNaker \cite{furber2012overview} & 130nm CMOS & General-Purpose CPU-Based & 18 cores per chip, up to 1 million cores, ARM-based & Brain modeling, robotic control \\
\hline
\end{tabular}
\label{tab:neuromorphic_platforms}
\end{table*}
    
\item In \cite{Norse}, Norse is a framework built on PyTorch designed to streamline the creation of SNNs for various ML tasks. It encompasses a wide array of neuron models, including the LIF and Izhikevich models. Norse embraces a functional programming style, where neurons are implemented as functions without an internal state. This requires the previous state of the neuron to be provided as an argument with each iteration. The framework accommodates two main learning paradigms: STDP for local unsupervised learning and SuperSpike for surrogate gradient learning (both are described in Section \ref{sec:learning_rules}). This dual approach supports both biologically inspired learning processes and efficient supervised training methods. Norse is fully compatible with PyTorch, enabling users to utilize PyTorch’s extensive features, such as GPU support, automatic differentiation, and a broad ecosystem of existing tools and libraries.
   
\item In \cite{eshraghian2023training}, the snnTorch framework is discussed as a tool to create SNNs using PyTorch's standard convolutional and fully connected layers. It integrates spiking neurons as intermediate layers, with these neurons modeled as classes containing their own internal states. The framework supports various neuron models, including LIF-based models, second-order LIF models, recurrent LIF models, and LSTM memory cells \cite{lotfi2020long, sun2022intelligence}. Learning is performed through Backpropagation Through Time (BPTT), which is further explained in Section \ref{sec:learning_rules}, with the use of surrogate gradients to approximate the gradient of the spiking neurons. The network outputs can be interpreted via both Rate and Temporal Decoding approaches.
    
\item In \cite{fang2023spikingjelly}, the SpikingJelly framework is introduced as a comprehensive open-source framework designed to enable the development and deployment of SNNs. It offers a full-stack toolkit for preprocessing neuromorphic datasets, building deep SNNs, optimizing their parameters, and deploying them on neuromorphic chips. The framework's architecture, built on top of PyTorch, leverages PyTorch's capabilities for automatic differentiation and parallel computation acceleration. The framework supports spiking neurons with simplified models like IF and LIF and advanced acceleration techniques using CUDA kernels for GPU-level efficiency \cite{cuda}. Key features of SpikingJelly are its support for surrogate gradients, STDP, and ANN-to-SNN conversion, which are further discussed in Section \ref{sec:learning_rules}. The framework also supports spiking encoders like Intensity-to-Latency and Poisson encoders, which are essential for converting non-binary input data into spikes. SpikingJelly supports both single-step and multistep simulation modes, accommodating various propagation patterns. It enhances computational efficiency by wrapping stateless layers with a SeqToANNContainer for parallel computation over time steps and using CuPy for semiautomatic CUDA kernel generation \cite{nishino2017cupy}. These acceleration methods significantly reduce the computational cost and development effort, enabling high-speed training and inference of deep SNNs. The framework also includes monitors for recording input/output data and attributes of specific layers, supporting detailed analysis and debugging. SpikingJelly supports neuromorphic devices, including computing chips like Intel's Loihi and Tianjic \cite{davies2018loihi, deng2020tianjic} which are described in the next subsection.
    
\item In \cite{rueckauer2022nxtf}, the NxTF programming interface and compiler is presented. NxTF is tailored for deploying deep SNNs on Intel's Loihi neuromorphic hardware. NxTF, derived from the Keras framework, allows for rapid development and mapping of deep SCNNs onto the multi-core Loihi architecture \cite{chollet2015keras}. The paper outlines the workflow for using NxTF, which includes training or converting models, defining the network in Python using the NxTF interface, and then compiling and partitioning the network to optimize resource allocation on Loihi’s neurocores. The paper evaluates NxTF using both directly trained SNNs and models converted from traditional DNNs, processing sparse event-based data and dense frame-based datasets. Notably, NxTF achieves near-optimal resource utilization, demonstrated by the deployment of a 28-layer, 4-million parameter MobileNet model on 16 Loihi chips with 80\% core utilization, achieving a CIFAR-10 error rate of 8.5\%, the lowest reported for neuromorphic hardware.
\end{itemize}

\subsubsection{Discussion}
Software simulation libraries are important for developing and testing SNNs, offering a versatile environment for experimenting with diverse network architectures, neuron models, and learning rules. Choosing the right framework for the right application is key in SNN implementation. Some frameworks cater specifically to biological realism, making them ideal for neurophysiological simulations. For instance, NEST and NEURON provide detailed models of neural activity with high biological fidelity, which is essential for understanding complex brain functions and conducting neuroscience research. Their ability to simulate intricate neuron models and synaptic interactions allows researchers to investigate the underpinnings of neural behavior and pathology. 

In contrast, frameworks designed for machine learning applications prioritize computational efficiency and scalability. Python has become the language of choice in this domain, not only due to its ease of use and extensive scientific ecosystem but also because of its ability to interface with high-performance libraries. Frameworks like BindsNET, snnTorch, and SpikingJelly are built on Python and leverage PyTorch's powerful GPU acceleration and automatic differentiation capabilities \cite{paszke2017automatic}. 

Automatic differentiation is particularly critical as it enables the efficient computation of gradients with respect to model parameters during the optimization process. This facilitates the training of SNNs through backpropagation and other gradient-based optimization algorithms, even in the presence of complex temporal dynamics and non-differentiable spiking functions. By constructing dynamic computational graphs and automatically applying the chain rule, automatic differentiation obviates the need for manual derivation and implementation of gradient computations. This dramatically simplifies the development of learning algorithms, reducing the complexity associated with coding gradient calculations manually - a task that is especially challenging in the context of SNNs due to their intricate temporal behaviors and event-driven nature. The open-source repository accompanying this paper offers a curated collection of references for the SNN libraries mentioned in this section, providing streamlined access to their documentation, tutorials, and toolkits \cite{Iaboni2024EventSNN}.

\subsection{Neuromorphic Hardware SNNs}
Traditional Von Neumann computer architectures, characterized by their separation of memory and computation, face challenges in efficiently handling the computational demands of SNNs due to high data transfer costs and limited parallelism \cite{indiveri2015memory}. To overcome these limitations, neuromorphic hardware platforms have emerged as a promising solution. These platforms are designed based on bio-inspired principles, achieving high energy efficiency and parallelism, making them particularly well-suited for real-time applications in machine intelligence and computational neuroscience. Neuromorphic hardware can be implemented in various forms, including digital ASICs, analog/mixed-signal systems, FPGA-based systems, and hybrid systems. Table \ref{tab:neuromorphic_platforms} provides a summary of the neuromorphic platforms explored in this section.

\subsubsection{Digital ASICs (Application-Specific Integrated Circuits)}

In \cite{akopyan2015truenorth}, the TrueNorth processor is presented by IBM. TrueNorth is a 65 mW neurosynaptic processor comprising 4096 neurosynaptic cores, totaling 1 million neurons and 256 million synapses. It uses a simplified LIF neuron model with stochastic elements to replicate a wide range of neural behaviors. TrueNorth achieves high energy efficiency with a power density of 20 $mW/cm^{2}$. The chip uses a 2D mesh network for intra-chip communication and a merge-split peripheral block for inter-chip communication. Performance metrics show the chip achieving up to 58 giga-synaptic operations per second (GSOPS) with an energy efficiency of 400 GSOPS/W. Other works have highlighted the use of TrueNorth in various applications, including visual object recognition \cite{moran2018deep}, and optical flow \cite{haessig2018spiking}.

In \cite{davies2018loihi}, Loihi, from Intel, extends the digital ASICs approach with on-chip learning capabilities. Fabricated in 14nm technology, Loihi features 128 neuromorphic cores, supporting 130,000 neurons and 130 million synapses. It uses a current-based LIF model with programmable synaptic learning rules to enable complex temporal and spatial learning. The Loihi architecture supports a range of SNN models, which incorporate time as an explicit dependency in their computations. This chip utilizes the Current-Based (CUBA) LIF model for spiking neurons, with two internal state variables: synaptic response current and membrane potential \cite{dampfhoffer2022investigating}. Loihi approximates these continuous-time dynamics using a fixed-size discrete time-step model. The chip's architecture consists of a manycore mesh comprising 128 neuromorphic cores, three embedded x86 processor cores, and off-chip communication interfaces. An asynchronous network-on-chip (NoC) manages all communication, including spike messages for SNN computation and time synchronization between cores. Each neuromorphic core implements 1,024 spiking neural units. Loihi uses a flexible learning engine that supports local synaptic plasticity rules, enabling on-chip learning through mechanisms like STDP and reinforcement learning.

\subsubsection{Analog/Mixed-Signal Systems}

In \cite{pehle2022brainscales}, the BrainScaleS neuromorphic architecture is presented. As part of the Human Brain Project, BrainScaleS uses an analog/mixed-signal design to emulate neuron and synapse dynamics efficiently. The analog core supports the adaptive exponential integrate-and-fire (AdEx) model, achieving a 1000x acceleration compared to biological time. The system supports various operational modes, including batch mode for parameter sweeps and in-the-loop operation for tasks requiring data dependencies. It uses analog parameter storage, addressing fixed-pattern noise and parameter drift through calibration. The synaptic crossbar processes weighted spikes and supports analog vector-matrix multiplication, making it suitable for both SNN and ANN tasks.

\subsubsection{Hybrid Systems}

In \cite{deng2020tianjic}, Tianjic is a chip architecture proposed to support various neural network models, including SNN, biological dynamic neural networks (BDyNNs), multilayered perceptrons (MLPs), CNNs, and RNNs \cite{sussillo2009generating, wu2008dynamics, li2016hierarchical, lipton2015critical}. Tianjic features a unified model description framework and a processing architecture that integrates five key building blocks: hybrid activation buffer (HAB), local synapse memory (LSM), shared integration engine (SIE), nonlinear transformation unit (NTU), and network connector (NC). This architecture enables both homogeneous and heterogeneous scalability to execute both spiking and non-spiking neural networks. The chip includes several optimization techniques, such as near-memory processing, data sharing, compute/access skipping, and intra-/inter-core pipelining, to enhance performance and efficiency. Fabricated in 28nm technology, Tianjic consists of 156 cores, achieving over 610 GB/s of internal memory bandwidth. The authors demonstrate the potential of Tianjic's hybrid paradigm through two examples: an autonomous bicycle system equipped with multimodal neural networks and a hybrid ANN-SNN neural network. The bicycle system integrates different models to handle tasks like voice command recognition and object tracking, and the hybrid network leverages ANN components to enhance the accuracy of large-scale SNNs.

\begin{table*}[t!]
\centering
\caption{Summary of Learning Rules for SNNs}
\label{tab:learning_rules_summary}
\begin{tabular}{|p{2cm}|p{2.5cm}|p{2.5cm}|p{6cm}|}
\hline
\textbf{Learning Rules} & \textbf{Key Features} & \textbf{Limitations} & \textbf{Applications and Performance} \\ \hline

Classic Additive STDP & Simple implementation, robust to noise, tolerant to synaptic variability & Unstable weight distributions, limited differentiation of similar objects & Caltech Faces (96.5\% accuracy) \cite{masquelier2007unsupervised} \newline Caltech Motorbikes (96.9\% accuracy) \cite{masquelier2007unsupervised} \newline MNIST (95\% accuracy) \cite{diehl2015unsupervised}                                                                                                                                                                              \\ \hline
Classic Multiplicative STDP & Stable weight distributions, more adaptive than additive STDP                             & Skewed weight distribution, poor differentiation of similar objects       & CIFAR-10 (56.93\% accuracy) \cite{falez2019unsupervised} \newline CIFAR-100 (30.45\% accuracy) \cite{falez2019unsupervised}\newline Caltech 101 (99.1\% accuracy) \cite{kheradpisheh2018stdp}\newline Biologically inspired pattern learning \cite{bichler2012extraction}\newline MNIST for digits 0, 3, and 4 (97.81\% accuracy) \cite{nessler2009stdp} \\ \hline

Mirror STDP  & High LTP correlation, supports feedforward and feed-backward connections   & Limited biological plausibility  & Auto-Encoder learning: Approximation of symmetric synaptic strength changes             \\ \hline

Probabilistic STDP & Simple implementation, robust to noise   & Unstable weight distributions   & Caltech Faces (98\%) \cite{tavanaei2016acquisition} \newline Motorbikes (92\% accuracy) \cite{tavanaei2016acquisition} \newline Enhanced performance under noise conditions \cite{tavanaei2016acquisition}  \\ \hline

Reinforcement STDP & High LTP correlation, focus on discriminative features & Unstable weight distributions & Caltech 101 (98.9\% accuracy) \cite{mozafari2018first} \newline ETH-80 (89.5\% accuracy) \cite{mozafari2018first} \newline NORB (88.4\% accuracy) \cite{mozafari2018first} \newline Sequence learning and syntactic tasks \cite{fang2021brain} \newline Visual navigation in grid-world environment \cite{chevtchenko2021combining}                                                              \\ \hline

Reverse STDP                & Stable top-down weight distributions, complements feedforward learning                   & High focus on correlation                                                 & Stability in top-down synapses \cite{burbank2012depression}                           \\ \hline

Triplet STDP                & Stable weight distributions, supports dense or overlapping time windows                   & Complex time-window definition                                            & MNIST \cite{diehl2015unsupervised} \newline Visual cortex modeling \cite{pfister2006triplets} \newline Nonlinear integration of synaptic changes \cite{caporale2008spike} \newline Nearest-neighbor triplet interactions \cite{krunglevicius2016modified}                                                              \\ \hline

SpikeProp                   & Classic error backpropagation, derived for multi-layer networks                           & Each output neuron can emit a single spike, requires network-wide reset   & Classification tasks \cite{xin2001supervised}\newline QuickProp \cite{mckennoch2006fast} \newline Resilient Propagation \cite{mckennoch2006fast} \\ \hline

ReSuMe  & Uses teacher neurons for spike timing, suitable for various subnetworks  & Requires external spike timings  & Supervised learning for spike timing tasks \cite{ponulak2010supervised}                                                                                                                                                                                                                                                                                                                      \\ \hline
BP-STDP   & Combines STDP with backpropagation, approximate gradient descent    & Approximation accuracy & MNIST (97\% accuracy) \cite{tavanaei2019bp}                                                                                                                                                                                                                                                                                                                                                  \\ \hline
S4NN & Rank-order temporal coding, error backpropagation using spike latencies  & Computational complexity  & MNIST (97.4\% accuracy) \cite{kheradpisheh2020temporal} \newline Caltech Faces/Motorbikes (99.2\% accuracy) \cite{kheradpisheh2020temporal}  \\ \hline

Temporal Backpropagation & Extends traditional backpropagation to temporal dynamics  & Non-differentiability of spikes & MNIST (98.77\% accuracy) \cite{lee2016training} \newline N-MNIST (96.66\% accuracy) \cite{lee2016training}  \\ \hline

SLAYER                      & Temporal credit assignment, GPU-accelerated implementation                                & Non-differentiability of spikes                                           & Temporal tasks \cite{shrestha2018slayer} \\ \hline

Surrogate Gradients         & Approximates gradient of spiking non-linearities, enables backpropagation-like algorithms & Approximation accuracy, computational complexity                          & Tempotron method \cite{gutig2006tempotron} \newline FORCE method \cite{nicola2017} \newline SuperSpike \cite{zenke2018} \newline STBP: MNIST (98.89\% accuracy) \cite{wu2018spatio} \newline N-MNIST (98.78\% accuracy) \cite{wu2018spatio} \newline Surrogate Gradient \cite{neftci2019surrogate} \newline VGG, ResNet: CIFAR-10 (90\% accuracy) \cite{lee2020enabling}      \\ \hline
\end{tabular}
\end{table*}

\subsubsection{General-Purpose CPU-based Systems}

In \cite{furber2012overview}, SpiNNaker, developed by the University of Manchester, uses a massively parallel architecture based on general-purpose ARM processors. The system consists of an array of ARM9 cores interconnected via a custom packet-switched network, achieving a high bisection bandwidth of over 5 billion packets per second. Each SpiNNaker node includes 18 ARM9 cores with associated memory, interconnected in a triangular lattice on a PCB, forming a toroidal topology to balance routing efficiency and physical constraints. The SpiNNaker system aims to model 1\% of the human brain, approximately 1 billion neurons, each with thousands of synapses and firing at an average rate of 10 Hz. Fault tolerance is integral to SpiNNaker, with mechanisms for fault detection and recovery embedded at multiple levels due to the expected component failures in such a large system. Each node has a router that handles multicast, point-to-point, nearest neighbor, and fixed route packets to ensure faultless communication.

\subsubsection{Discussion}
Neuromorphic hardware platforms are a promising development in addressing the computational challenges associated with SNNs. Designed to emulate the neural architectures of the brain, these platforms offer advantages in energy efficiency and parallel processing, making them ideal for real-time applications. Despite their promising capabilities, these neuromorphic hardware platforms are not easily accessible and are primarily research-oriented products that have not yet reached the commercial market, limiting broader adoption. However, ongoing developments in this space are key for advancing neuromorphic computing. As these technologies mature and become more accessible, they hold the potential to revolutionize CV with efficient and scalable solutions.

\begin{figure*}[t!]
\centering
\includegraphics[scale=0.55]{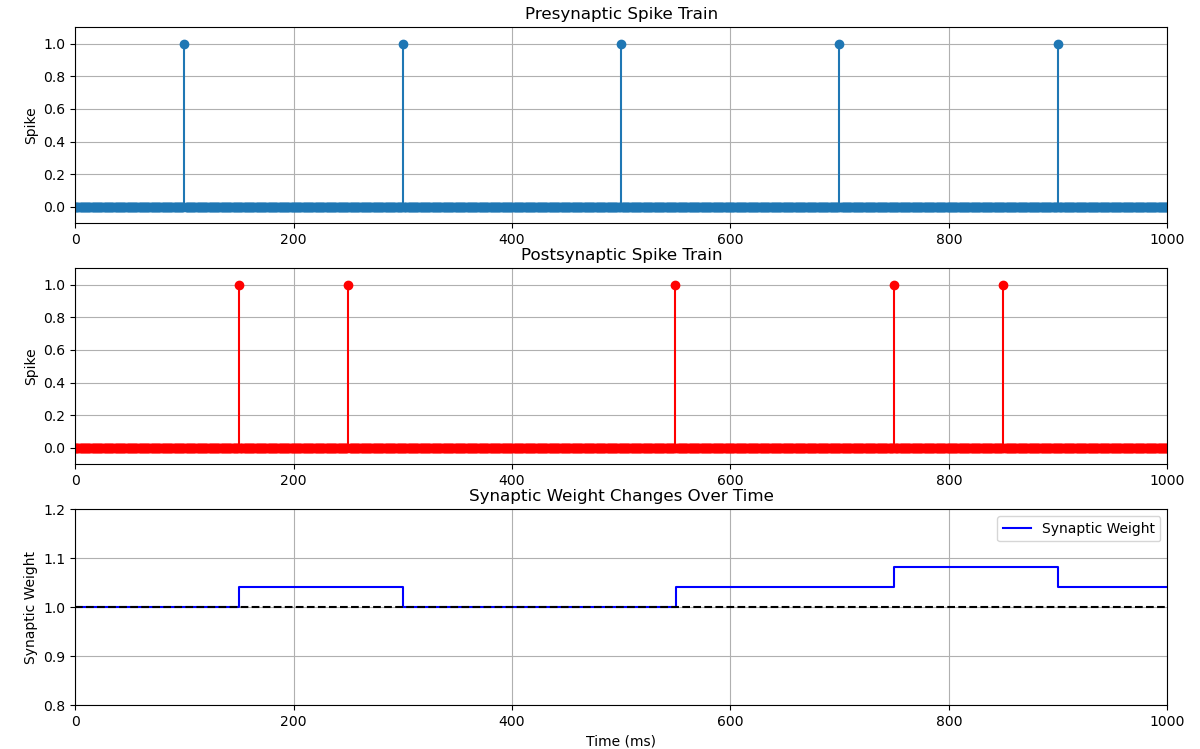}
\caption{This figure depicts the synaptic weight changes over time using standard STDP. The weights are updated based on the relative timing of pre-synaptic (top) and post-synaptic (middle) spikes. Synaptic potentiation and depression are determined by the difference between pre-synaptic and post-synaptic spike times.}
\label{fig:stdp}
\end{figure*}

\section{Learning Rules}
\label{sec:learning_rules}
This section delves into learning methods that enable SNNs to adapt synaptic weights based on neural activity. It covers both unsupervised and supervised learning techniques, highlighting key advancements like Spike-Timing-Dependent Plasticity and backpropagation-inspired methods. Table \ref{tab:learning_rules_summary} summarizes the key features, limitations, applications, and performance of various learning rules for SNNs.

Learning rules and architecture in SNNs are closely intertwined. While the architecture defines the structural layout and flow of information through layers of neurons, learning rules dictate how synaptic weights are updated in response to stimuli. These two components work in harmony, as certain architectures may inherently support or require specific learning rules to achieve optimal performance. For instance, convolutional architectures are designed to capture spatial hierarchies in data, and pairing them with learning algorithms such as supervised learning enables efficient feature extraction for tasks like classification. Although we have separated these sections for clarity in the paper, the relationship remains critical as learning rules often exploit the strengths of a given architecture, or vice versa.

\subsection{Direct Unsupervised Learning}
In the context of SNNs, unsupervised learning involves adapting synaptic weights based on the timing and sequence of incoming spikes. This process enables the network to self-organize and improve its response to recurring patterns in the input data, allowing it to recognize and process stimuli without explicit external supervision or labeled data.

\subsubsection{Spike-Timing-Dependent Plasticity}
A foundational mechanism in unsupervised learning within SNNs is Spike-Timing-Dependent Plasticity (STDP). STDP is often referred to as Hebbian learning, as the concept of synaptic plasticity stems from Hebb's 1949 postulate, which describes how synaptic connections should be modified and is frequently summarized as the statement, "neurons that fire together, wire together" \cite{hebb2005organization}. This mechanism supports the formation of neural circuits that are sensitive to specific temporal patterns of spikes, allowing the network to recognize and react to patterns in sensory input \cite{vigneron2020critical}.

The temporal relationship between spikes of pre- and post-synaptic neurons can be used to define learning rules for updating synaptic weights, as supported by several experiments \cite{masquelier2007unsupervised, kheradpisheh2018stdp, diehl2015unsupervised}. Specifically, let $t_{pre}$ and $t_{post}$ represent the timing of pre- and post-synaptic spikes, respectively. The difference in spike time, $\Delta t = t_{post} - t_{pre}$, is key for this learning mechanism. If $\Delta t$ is positive, indicating that the post-synaptic neuron fires after the pre-synaptic neuron, synaptic potentiation is often observed. Conversely, if $\Delta t$ is negative, indicating that the post-synaptic neuron fires before the pre-synaptic neuron, synaptic depression typically occurs. This temporal difference forms the basis of STDP, a fundamental Hebbian learning rule, where the magnitude of synaptic modification is directly proportional to the precise timing difference between the spikes, enabling the network to capture temporal dependencies in the data.

Moreover, the probabilistic interpretation of STDP aligns with Bayesian inference principles, suggesting that SNNs might perform approximate Bayesian computations to infer the causes behind observed sensory data, using a stochastic version of the expectation-maximization algorithm \cite{guo2017hierarchical}.

\paragraph{Classic Additive STDP}
The classic additive model of STDP adjusts synaptic weights based on the relative timing of spikes between pre- and post-synaptic neurons. The rule is simple: if a pre-synaptic neuron fires just before a post-synaptic neuron, the synaptic connection between them is strengthened (long-term potentiation, LTP); if the order is reversed, the connection is weakened (long-term depression, LTD) \cite{gerstner2002spiking}. Figure \ref{fig:stdp} depicts the classic STDP dynamic.

\begin{equation}\label{eq:plain_stdp}
\Delta w = \left\{\begin{matrix}
A_+ \exp (\frac{t_{pre}-t_{post}}{\tau _{+}}) \text{if $t_{pre} \leq t_{post}$}\\ 
-A_- \exp(-\frac{t_{pre}-t_{post}}{\tau_{-}}) \text{if $t_{pre} > t_{post}$}\\ 
\end{matrix}\right.
\end{equation}

Where $A_+$ and $A_-$ are scaling factors for potentiation and depression of the synapse, $t_{pre}$ and $t_{post}$ are presynaptic and post-synaptic spike times, and 
$\tau _{+}$ and $\tau _{-}$ are time constants of synaptic potentiation and depression.

The strength of the potentiation or depression effect is expressed as a function of the time difference between the pre-synaptic and post-synaptic spikes. Inputs that cause the output spike are made more likely to contribute to firing in the future, while inputs that are unrelated to the output spike are made less likely to cause firing. STDP implementations that modulate synaptic weights entirely based on the above ruleset are known as additive STDP. This ruleset implies that the LTP and LTD can be applied an infinite number of times, which is not biologically plausible and makes learning unstable. 

\begin{itemize}
    \item In \cite{masquelier2007unsupervised}, the authors demonstrated the significant potential of STDP in unsupervised learning of visual features within an asynchronous feedforward SNN. The authors show that when the network is exposed to natural images, it develops a selectivity for intermediate-complexity visual features, identifying recurring patterns that stand out across different images. They evaluate the model's performance on two datasets: one containing faces and another containing motorbikes, with additional background images acting as distractors.
    The model's classification accuracy is measured using two schemes: "Simple Count," which counts the number of spikes to make predictions, and "Potential + RBF," which incorporates the neuron's membrane potential and a Radial Basis Function (RBF) classifier for decision-making. Under the "Simple Count" scheme, the model achieves an equilibrium point accuracy — where the model's performance stabilizes — of 96.5\% for faces and an area under the Receiver Operating Characteristic (ROC) curve of 99.1\%, and 96.9\% accuracy for motorbikes with an area under the ROC curve of 99.7\%. Using the 'Potential + RBF' scheme, the model attains 95.4\% accuracy for faces with a ROC area of 98.4\%, and 97.8\% accuracy for motorbikes with a ROC area of 99.3\%.

    \item In \cite{diehl2015unsupervised}, the standard STDP rule was applied to the MNIST dataset to achieve a classification accuracy of 95\% when configured with 6400 excitatory neurons. This work is significant as it demonstrates that unsupervised STDP learning can lead to high classification performance on a widely used benchmark dataset, showcasing the potential of SNNs in practical applications.
\end{itemize}

\paragraph{Classic Multiplicative (Stable) STDP}
For the sake of biological and artificial plausibility, it would be impractical to keep records of all spike timing occurrences (as is done in Equation \ref{eq:plain_stdp}). To this end, the spike trace $x$ is introduced with multiplicative STDP, which considers the current synaptic weight value when computing how best to modulate synaptic weights.

\begin{equation}\label{multi_stdp_a}
\Delta w =A_{+}x_{pre} \cdot \delta_{post} - A_{-}x_{post} \cdot \delta_{pre}
\end{equation}

\begin{equation}\label{multi_stdp_b}
\left\{\begin{matrix}
 &\frac{dx_{pre}}{dt}=-\frac{x_{pre}(t)}{\tau_{+}}+\delta(t)  \\ 
 &\frac{dx_{post}}{dt}=-\frac{x_{post}(t)}{\tau_{-}}+\delta(t)
\end{matrix}\right.
\end{equation}

Where \(\Delta w\) represents the change in synaptic weight, \(A_{+}\) and \(A_{-}\) represents the learning rate parameters for potentiation and depression, \(x_{pre}\) and \(x_{post}\) represents the pre- and post-synaptic spike traces, \(\delta_{post}\) and \(\delta_{pre}\) represents the occurrences of post- and pre-synaptic spikes, and \(\tau_{+}\) and \(\tau_{-}\) represents the time constants for the decay of the spike traces. 

The two equations together describe how the synaptic weight change (\(\Delta w\)) is modulated by the pre- and post-synaptic spike traces (\(x_{pre}\) and \(x_{post}\)), which evolve over time according to the occurrence of spikes and their respective decay rates. This multiplicative STDP approach, also called stable STDP (S-STDP), is fundamentally more stable by integrating the weight magnitude into the learning rule, preventing uncontrolled growth or decay of synaptic weights \cite{paredes2019unsupervised}. 

\begin{figure*}[t!]
\centering
\includegraphics[scale=0.55]{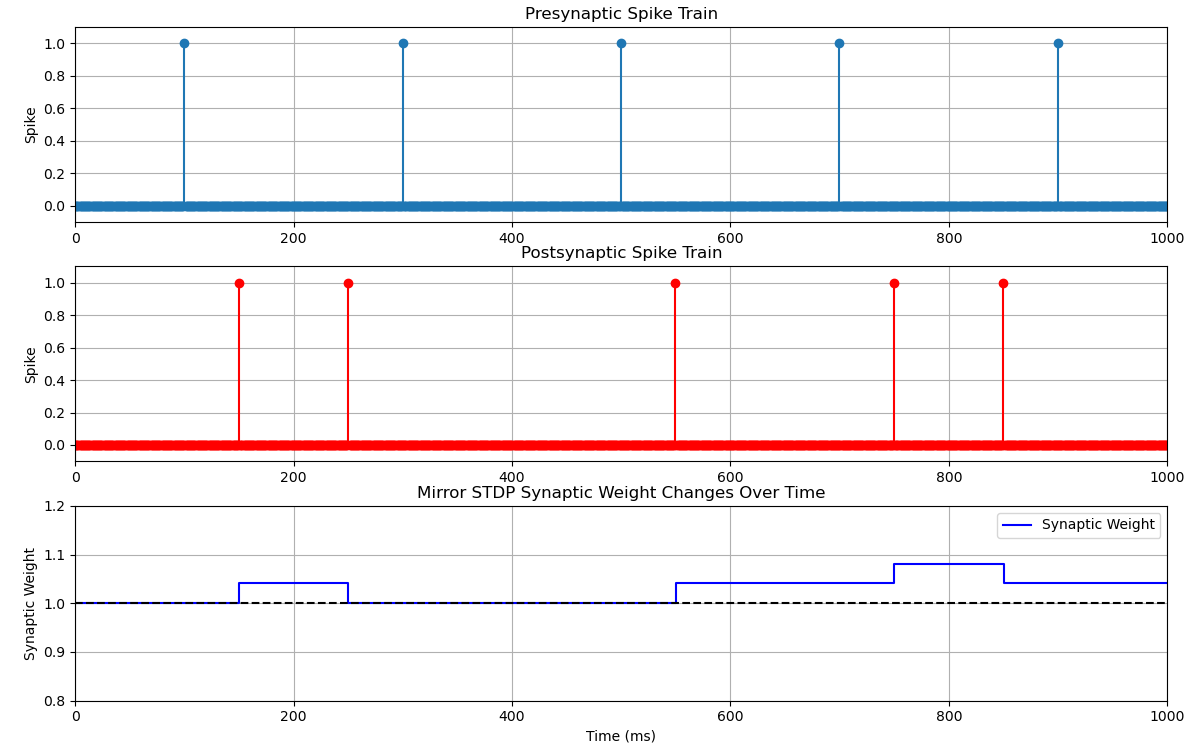}
\caption{This figure depicts the synaptic weight changes over time using mirror spike-timing-dependent plasticity (mSTDP). The weights are updated based on the relative timing of presynaptic (top) and postsynaptic (middle) spikes, centered around the postsynaptic spikes. Synaptic potentiation and depression are determined by the difference between presynaptic and postsynaptic spike times.}
\label{fig:mstdp}
\end{figure*}

\begin{itemize}
    \item In \cite{falez2019unsupervised}, the performance of STDP for unsupervised visual feature learning was compared to sparse Auto-Encoders (AE). The study used the IF neuron model and the mSTDP rule. Image preprocessing is handled through on-center/off-center coding, which employs Difference-of-Gaussian (DoG) filtering to highlight edges and convert pixel values into spike trains using latency coding. Experiments were conducted on the CIFAR-10, CIFAR-100, and STL-10 datasets. For CIFAR-10, SNNs with 1024 features achieved 56.93\% accuracy on color images, while AE attained 66.98\%. With 64 features, SNNs scored 48.27\% on color images, compared to AEs' 57.56\%. On CIFAR-100, SNNs with 1024 features reached 30.45\% on color images, while AEs achieved 36.43\%. With 64 features, SNNs obtained 25.20\% accuracy on color images versus AEs' 37.71\%. Overall, AE architectures consistently outperformed the SNNs across all datasets and feature sizes.

    \item In \cite{kheradpisheh2018stdp}, the use of STDP in SNNs is explored for object detection. The authors designed a deep SNN comprising convolutional and pooling layers, where STDP was used to train the network in an unsupervised manner. The authors emphasize the advantages of the m-STDP rule, which ensures that synaptic weights remain within a stable range between 0 and 1. This rule allows synapses to be reinforced or depressed based on the relative timing of pre- and post-synaptic spikes, with the magnitude of weight changes being proportional to the current weight value, thereby preventing runaway synaptic growth or complete decay. 
    The method uses a winner-take-all mechanism, where the earliest firing neurons inhibit others to enhance feature diversity and learning efficiency. The SNN was evaluated on the Caltech 101, ETH-80, and MNIST datasets, achieving accuracies of 99.1\%, 82.8\%, and 98.4\%, respectively. These results were superior to traditional unsupervised techniques like autoencoders and highlighted the potential of STDP-based learning for efficient and robust object recognition.

    \item In \cite{bichler2012extraction}, a biologically inspired approach to learning temporally correlated patterns from spiking silicon retinas using STDP was presented. The proposed system used a feedforward SNN with an m-STDP rule. The results demonstrated the network's ability to learn entire trajectories with a 98\% detection rate using a second layer, making it suitable for applications like car counting.

    \item In \cite{nessler2009stdp}, the authors investigated how STDP can enable spiking neurons to develop internal models for subclasses of high-dimensional spike patterns originating from hundreds of pre-synaptic neurons. The authors proposed that STDP, combined with a stochastic soft winner-take-all circuit, approximates the Expectation-Maximization (EM) algorithm for discovering hidden causes in input data, effectively performing a dimensionality reduction and creating internal models for input pattern distributions. The authors demonstrated the applicability of their model using the MNIST dataset, focusing on handwritten digits 0, 3, and 4. They encoded pixel values using Poisson spike trains and connected input neurons to a winner-take-all network of output neurons. Accuracy of the learning process was reported with a classification accuracy of 97.81\% for digits 0 and 3 after 2000 training steps and 96.32\% for digits 0, 3, and 4 after 4000 training steps on independent test sets of 10,000 new samples each. 
    
\end{itemize}

\paragraph{Mirror-STDP}
Mirror STDP (mSTDP) is distinguished from other variants by the fact that the time window for analyzing the correlation between afferent and efferent spikes is centered on the efferent spike. This adjustment allows for correlating $\delta < 0$ with LTP instead of LTD. This is achieved by simplifying the complex and unknown mechanisms that cause efferent and afferent neurons to fire together, although this results in biological implausibility. This implementation emphasizes correlated mechanisms without causality, thereby neglecting Hebb's principle that neurons that "take part in firing" strengthen their synaptic connection. Figure \ref{fig:mstdp} depicts the Mirror-STDP dynamic.

In \cite{burbank2015mirrored}, the authors use the mSTDP plasticity rule to implement Auto-Encoder learning in an SNN. The network approximates AE learning, which traditionally requires symmetric changes in synaptic strengths. The network consists of visible and hidden layers with reciprocal connections, and the learning process involves sequential stimulus presentations. Visible neurons receive input spikes proportional to stimulus strength, causing hidden neurons to fire. Feedback from hidden neurons then attempts to reconstruct the input in visible neurons. Synaptic scaling is used to induce sparsity in hidden neuron activity, akin to biological synaptic scaling observed in cortical neurons.

\paragraph{Probabilistic-STDP}
The probabilistic STDP (p-STDP) rule is defined by weight changes following an exponential decay function related to the weight magnitude. The p-STDP rule is mathematically expressed as:

\begin{equation}\label{p-stdp}
\Delta w_{i} = \left\{\begin{matrix}
\alpha_{+} \cdot e^{-w_{i}} & \leftrightarrow \delta \geq 0 & LTP \\ 
\alpha_{-}, &  \leftrightarrow \delta < 0 & LTD 
\end{matrix}\right.
\end{equation}

Where $\alpha_{+}$ is the amplification parameter of LTP and $\alpha_{-}$ is the amplification parameter for LTD. The magnitudes of these variables are restricted to stay within a 4/3 ratio. 

\begin{itemize}
    \item In \cite{tavanaei2016acquisition}, the authors examined modifications to a five-layer SCNN by introducing Izhikevich-like neurons and the p-STDP rule. Specifically, the researchers tested these modifications on the Caltech dataset for face and motorbike recognition tasks. In the original dataset, the network equipped with the probabilistic STDP rule demonstrated an improvement in accuracy, with SNN2 (using probabilistic STDP) reaching around 98\% for face recognition compared to 97\% in the original model (SNN1). For motorbike recognition, SNN2 achieved approximately 92\%, outperforming the original model's 90\%. Additionally, the area under the ROC curve showed slight improvements, with SNN2 achieving about 99.5\% for faces and 96\% for motorbikes. This study is significant as it shows that incorporating probabilistic elements into STDP can enhance the learning performance of SNNs on visual recognition tasks.

    \item In a subsequent experiment, the researchers added Gaussian noise to the test images to mitigate ceiling effects and found that the probabilistic STDP rule continued to enhance performance. For noisy face recognition, SNN2 reached around 92\% accuracy compared to 90\% in the original model. Similarly, motorbike recognition accuracy improved to 87\% with probabilistic STDP. Control experiments using the 3D-Object dataset further validated these findings, showing that the probabilistic STDP consistently outperformed the original rule across multiple object categories.
\end{itemize}

\paragraph{Reinforcement-STDP}
Reinforcement-STDP is a biologically inspired learning approach that integrates principles from reinforcement learning to enhance the training and performance of SNNs \cite{glimcher2011understanding}. It builds upon neuromodulation mechanisms observed in the brain, where neuromodulators such as dopamine influence synaptic behaviors, thereby modulating decision-making processes. This approach leverages the concept of eligibility traces, where the temporal influence of spikes is considered over a broader time window, allowing for more nuanced adjustments of synaptic weights based on reward signals.

\begin{itemize}
 
\item In \cite{mozafari2018first}, the authors proposed a hierarchical feedforward convolutional SNN employing a temporal coding scheme trained using reward-modulated STDP (R-STDP). Their significant contribution lies in combining unsupervised STDP with reinforcement signals to enable the network to learn complex visual tasks without explicit labels. In this network, neurons fire in order of their activation strength, and the network is trained using reward-modulated spike-timing-dependent plasticity (R-STDP). The authors adopt Pavlov's conditioning approach to learning, incorporating elements from Hebbian learning to contribute to the biological learning framework. They use the concept of eligibility trace, where afferent spikes influencing an efferent neuron's firing receive rewards, modeled as the probability of participation in the spike event \cite{tanner2005td}. The network uses R-STDP with parameters $\alpha$ for weight change according to the reward/punishment signal and $\eta$ to prevent overfitting. The effectiveness of the training is quantified using two equations that measure the proportion of missed and correctly labeled samples, respectively:

\begin{equation}\label{r-stdp1}
\eta_{+} = \frac{|\text{missed samples}|}{|\text{training samples}|}
\end{equation}

\begin{equation}\label{r-stdp2}
\eta_{-} = \frac{|\text{correctly labeled samples}|}{|\text{training samples}|}
\end{equation}

Experimental results showed superior performance with accuracies of 98.9\% on Caltech face/motorbike, 89.5\% on ETH-80, and 88.4\% on NORB datasets.

\item In a subsequent study \cite{fang2021brain}, R-STDP is applied to supra-regular grammar tasks using an SNN capable of memorizing and recalling sequences through population coding. The network demonstrates its ability to overcome the "syntax barrier" previously thought to be unique to humans \cite{hauser2002faculty}. The production accuracy showed a U-shaped pattern, indicative of the primacy and recency effects, aligning with biological plausibility and reflecting cognitive processes found in macaques \cite{luchins1957primacy, jiang2018production}. Furthermore, the introduction of the chunking mechanism allows the network to handle longer sequences more effectively by breaking them into shorter segments, thereby reducing cognitive load during the production process.

\item In \cite{chevtchenko2021combining}, a hybrid model is introduced, combining a pre-trained binary CNN for feature extraction with an SNN trained online through reward-modulated R-STDP. The study also demonstrates the significant impact of optimizing synaptic sparsity between input and hidden layers on learning speed and accuracy.  The experimental setup involved a grid-world environment with high-dimensional RGB image observations (256 $\times$ 256 pixels). The agent's task was to navigate to a rewarded state using only visual inputs. Feature extraction was performed using a pre-trained binary CNN (BinaryNet) to reduce the dimensionality of the input space. This binary CNN was trained on a small set of random images distinct from those used during testing. 

\end{itemize}

\paragraph{Reverse-STDP}
Reverse-STDP or top-down STDP, is a mechanically complex variant of STDP in the context of biological implementations within the brain. Reverse-STDP in the brain takes place with top-down synaptic connections, which transmit information in the opposite direction. Reversed synaptic connections are not usually mutually exclusive and can act as a complimentary feed-backward feature to the standard feedforward architecture if the weight adaptation mechanism supports it. Whereas pre-synaptic and post-synaptic seem to describe a temporal order of firing in most SNNs, they, in fact, relay a topological ordering. In the brain, pre-synaptic afferent neurons are upstream of post-synaptic efferent neurons. In terms of spike timing, $N_{aff}$ will fire before $N_{eff}$ for LTP to occur. In feedforward STDPs, this ordering is also the same as the brain. For the Reversed-STDP case, spikes are transmitted in the reverse direction through the network. $N_{eff}$ will fire first, sending information to $N_{aff}$, which is the next in it's backwards path. The Reversed-STDP style of information transmission often occurs alongside forward STDPs, and can result in stable weight distributions when learning is biased towards depression. 

In \cite{burbank2012depression}, the authors propose Reverse-STDP, where synaptic depression predominates over potentiation when post-synaptic activity precedes pre-synaptic activity. The study compared classical STDP and Reverse-STDP rules, both analytically and through integrate-and-fire simulations. The authors found that stability required Reverse-STDP at top-down synapses, and classical STDP lead to unstable weights. 

\paragraph{Triplet-STDP}
Standard STDP rules define the relationship between a pair of pre-synaptic and post-synaptic spikes. Triplet-STDP expands this to the relationship between a triplet of spikes. The triplet can be composed of either two pre- and one post, or one pre- and two post-synaptic spikes. This rule is unique in that a triplet of spikes can account for the dependence on the repetition frequency for a set of spikes. However, with more spikes, it becomes more difficult to attribute the influence of afferent spikes to efferent neurons. 
The triplet rule has been theorized by experiments which showed that pre-synaptic spikes which enabled LTP for a given efferent neuron and should have caused LTD in other efferent neurons did not do so, suggesting that the brain had access to information about the increase or decrease of the potential of $N_{eff}$ \cite{caporale2008spike}. This suggests that two synaptic traces should be modeled rather than one.

\begin{equation}\label{eq:diehlstdp1}
\Delta w = \eta (x_{pre} - x_{tar})(w_{max}-w)^u
\end{equation}

Where $\eta$ is the learning rate, $w_{max}$ is the maximum weight, and $u$ determines the dependence of the update on the previous weight. $x_{tar}$ is the target value of the pre-synaptic trace at the moment of a post-synaptic spike. Higher target values indicate lower synaptic weights. This offset allows pre-synaptic neurons that rarely cause the firing of post-synaptic neurons to become increasingly disconnected, especially if the post-synaptic neuron is rarely active.

Several works have proposed variants of the triplet rule.
\begin{enumerate}

    \item In \cite{pfister2006triplets}, the triplet-based STDP rule is proposed to overcome inherent limitations imposed by pairs of pre- and post-synaptic spikes failing to account for the frequency dependence of spike pairings. The authors incorporated additional biophysically-aligned variables to represent synaptic events, such as variables to denote the amount of glutamate bound to post-synaptic receptors and calcium concentration for post-synaptic spikes. To validate the model, the authors used visual cortex data from \cite{sjostrom2001rate}, which shows weight changes induced by pairs of pre- and post-synaptic spikes at various frequencies.
    
    \item In \cite{caporale2008spike}, the triplet rule was further explored to address the nonlinear integration of synaptic changes induced by complex spike patterns. It extends the traditional pair-based STDP model by considering the interaction between three spikes, which are more representative of natural neuronal activity. The non-linearity observed in the triplet rule suggests that the influence of a spike on synaptic plasticity is affected by the preceding spikes. For example, earlier spikes can suppress the effect of later spikes, leading to a complex interplay that cannot be captured by simple pair-based STDP models. 

    \item In \cite{krunglevicius2016modified}, the triplet STDP rule is used to test nearest-neighbor triplet interactions. Heuristic optimization using a genetic algorithm was used to tune the neuron and STDP parameters for each pattern size. The objective function maximized the difference in synaptic strengths associated with the pattern versus noise. The triplet rule significantly outperformed the all-to-all and nearest-neighbor STDP rules across multiple single-spike and multi-spike recognition benchmarks.

    \item In \cite{diehl2015unsupervised}, the triplet STDP rule was evaluated in a network of 6400 excitatory neurons. This rule uses no weight dependence for learning, so divisive weight normalization \cite{goodhill1994role} was adapted to ensure equal use of the neurons. On the MNIST dataset, it did not achieve higher classification accuracy than the classical STDP rule tested by the authors.

\end{enumerate}

\subsubsection{Discussion}
Unsupervised learning methods, particularly those relying on STDP, are important in enabling SNNs to self-organize and recognize patterns without explicit supervision. The classic additive STDP, despite its simplicity, has demonstrated significant promise in various feature learning, as evidenced by high classification accuracies on various datasets. However, its limitations in biological plausibility and stability necessitate alternative approaches, such as multiplicative STDP variants. In mSTDP, the incorporation of weight magnitude in synaptic modulation ensures stable learning, preventing uncontrolled growth or decay of synaptic weights. Studies using modified STDPs, such as those on CIFAR-10 and CIFAR-100 datasets \cite{falez2020improving, zhou2020imbalanced}, have shown that while SNNs can achieve competitive performance, they still often lag behind traditional deep learning methods like ANNs and autoencoders in terms of accuracy. Other variants, such as mirror-STDP, probabilistic-STDP, and reinforcement-STDP, offer additional mechanisms to enhance learning stability and biological plausibility.

\subsection{Indirect Learning}
Indirect learning, or training by conversion, offers a practical solution to the challenges faced in training SNNs. This method involves initially training an ANN model and then fine-tuning an SNN model with the same architecture, initializing the weights of the SNN with those from the trained ANN \cite{hunsberger2015spiking}. By converting to SNN format, researchers and practitioners can leverage the well-established training methods of ANNs while circumventing the discrete and non-differentiable nature of spikes and exploit energy-efficient computation for inference \cite{rueckauer2017conversion}. 

Several studies have explored various approaches to indirect learning, each addressing specific challenges posed by the conversion process. 

\begin{itemize}

\item In \cite{hunsberger2015spiking}, an ANN was first trained using the soft-LIF activation function, which approximates the firing rate of LIF neurons but remains differentiable. After training, the soft-LIF was replaced by LIF neurons, converting the input and output to spikes. This work is significant because it provides a smooth transition from ANN training to SNN implementation, preserving the performance of the network while enabling it to operate with spiking dynamics.

\item In \cite{cao2015spiking}, a CNN architecture was modified to produce only positive values and biases were replaced with average pooling. An SNN was then initialized with IF neurons, replacing the CNN structure and transferring the trained weights. This approach is significant as it demonstrated that deep CNNs could be successfully converted into SNNs capable of achieving competitive performance on image recognition tasks, such as achieving 99.1\% accuracy on the MNIST dataset using spiking neurons.

\item In \cite{diehl2015fast}, the method from \cite{cao2015spiking} is extended by normalizing the ANN weights before transferring them to the SNN. This step prevents insufficient membrane potentials or excessively large spikes. They propose two normalization techniques: 
\begin{enumerate}
    \item \textbf{Model-based Normalization:} This method requires knowledge of the network weights but not the training set. The maximum positive activation that could occur as input to a layer is noted as normalization factor $\lambda^{l}$. All synaptic weights preceding a neural layer are scaled by $\lambda^{l}$, and the threshold is set to 1. Alternatively, the threshold is set to $\lambda^{l}$ with the synaptic weights unchanged. This configuration ensures that the network will never produce more than one spike at once from a given neuron when maximum positive input can cause a single spike. 
    
    \item \textbf{Data-based Normalization:} This method requires knowledge of the training set and the network weights. Rather than assuming the worst-case scenario of maximum possible activation, the realistic maximum activation within the training set is used. The maximum ReLU activation in layer $l$ is noted as normalization factor $\lambda^{l}$. All synaptic weights preceding a neural layer are scaled by $\lambda^{l}$, and the threshold is set to 1, or the threshold is set equal to $\lambda^{l}$ with the synaptic weights unchanged. This ensures that performance parity with the source ANN is maintained.
\end{enumerate}

\item In \cite{rueckauer2017conversion}, an SNN is trained with architectural constraints and converted to a SNN with IF neurons and bias terms. Their significant contribution lies in addressing several architectural and functional constraints to ensure that the SNN closely approximates the performance of the original ANN. The architectural constraints for converting continuous-valued DNNs to SNNs can be summarized as follows:
\begin{enumerate}
    \item One-to-One Neuron Mapping: Each neuron in the original DNN is mapped to a corresponding spiking neuron in the SNN. This requires ensuring that the firing rates of the spiking neurons approximate the activations of the DNN neurons.

    \item Input Representation: The input to the SNN can be represented as constant currents proportional to pixel intensities for image data rather than converting these to Poisson spike trains to reduce variability.

    \item Weight Normalization: Weights must be normalized to avoid excessive or insufficient firing in the SNN. This normalization can be data-based or can include biases to ensure the network operates within the spiking neuron firing rate limits. Instead of normalizing by the maximal activation, which can be affected by outliers, a percentile-based normalization (e.g., $99^{\text{th}}$ percentile) was used to improve robustness.

    \item Reset Mechanism: The reset-by-subtraction mechanism is used instead of the reset-to-zero mechanism to prevent excess charge after firing. 

    \item Implementation of Common ANN operations: The conversion method supports common ANN operations like max-pooling, softmax, batch-normalization, and biases.
\end{enumerate}

\item In \cite{severa2018whetstone}, an iterative method is proposed to train ANNs for binary communication, enabling their use on spiking neuromorphic hardware. The method, named Whetstone, gradually sharpens neuron activations during the training process to approximate discrete step functions, thus converting the network's communication to a binary, spike-based format. This approach integrates the conversion into the training phase, ensuring that the trained SNN retains performance levels comparable to its non-spiking counterpart. 

\item In \cite{sengupta2019going}, an ANN is trained and initialized as SNN with the same architecture before performing threshold balancing using a Spike-based normalization (Spike-Norm) method. Spike-Norm determines the firing thresholds using spike activity instead of model or data properties of ReLU neurons. Poisson spike trains are generated for the entire training set. The first layer is initialized with ANN-trained weights, and the maximum summation of weighted spike-input received by the first layer is recorded. The activation threshold is set to the maximum spike-input across time-steps over the entire training dataset, ensuring no neuron will emit more than a single output spike at any given time-step. This process is repeated sequentially for each layer. The authors have demonstrated that this conversion approach ensured near-lossless (less than 0.15\% error loss) conversion for VGG-16. The main difference between Spike-Norm and ReLU-based normalization \cite{diehl2015fast} is that Spike-Norm's iterative spike-based approach accounts for spike-train behavior used as SNN input and output during the conversion process, rather than normalizing based on ANN output.

\item In \cite{han2020rmp}, the RMP-SNN method is used to address conversion loss by using IF neurons with soft-reset, which maintain residual potential post-spike generation. This mechanism is detailed further in Section \ref{sec:neural_dynamics}.

\item In \cite{deng2021optimal}, the conversion loss is reduced by using IF neurons with soft-reset and shifted threshold ReLU functions. The authors first train an ANN with a threshold ReLU activation before converting the model to an SNN. SNN biases were set based on the ANN's biases adjusted by half the threshold over time steps.

\item In \cite{ding2021optimal}, the Rate Norm Layer (RNL) is introduced to replace the ReLU activation layer in the source ANN. The RNL employs a clip function with a trainable upper bound to simulate neuron firing rates, allowing adaptation during training. Furthermore, they present an optimal fit curve to assess the correlation between ANN activation values and SNN firing rates. By fine-tuning the upper bound of this curve, they demonstrate a significant reduction in inference time. This conversion technique is applied to advanced neural architectures such as VGG-16 and PreActResNet.

\item In \cite{patel2021spiking}, the conversion of the U-Net ANN architecture to a SNN is described. The original ANN is scaled down to fit within the constraints of the Intel Loihi chip, reducing input image size and number of layers and replacing max-pooling with strided convolutions. The network was trained with a rate-based non-linearity that approximates spiking behavior. During inference, this non-linearity is replaced with the IF neuron model. The authors introduce a percentile-based regularization loss function to control the firing rates within the range of 50-200 Hz. Additionally, the network parameters were quantized to 8 bits to fit the fixed-point arithmetic used by Loihi.

\end{itemize}

\subsubsection{Discussion}
Indirect learning allows researchers to bypass the inherent difficulties in training SNNs directly, primarily due to their non-differentiable spiking activity. By initializing the SNN weights from the pre-trained ANN, the established and efficient training paradigms of ANNs can be exploited while harnessing the energy-efficient nature of SNNs for inference.
However, the conversion process has its limitations. The primary issue lies in the disparity between the operational formats during training and inference. ANNs operate with continuous activations, whereas SNNs rely on discrete spikes and temporal dynamics. This difference means that the converted SNN does not naturally learn or exploit temporal information inherent in its spiking nature during the initial ANN training phase. Several methods have been proposed to mitigate these challenges, such as the introduction of the soft-LIF activation \cite{hunsberger2015spiking} to approximate LIF neurons' firing during ANN training and normalization techniques to ensure stable membrane potentials and approximate spike-rates post conversion \cite{diehl2015fast, rueckauer2017conversion}. More recent approaches further refine the conversion process. The RMP-SNN method \cite{han2020rmp} addresses residual potentials with IF neurons using a soft-reset mechanism, while shifted threshold ReLU functions \cite{deng2021optimal} optimize threshold settings, and the Rate Norm Layer (RNL) \cite{ding2021optimal} provides a trainable upper bound to simulate neuron firing rates before conversion. 

Despite these advancements, the inherent limitation of indirect learning remains: the lack of native temporal learning during the ANN training phase. The converted SNN, although efficient for inference, does not fully exploit the temporal dynamics that are natural to spiking neurons. This fundamental difference underscores the need for ongoing research to develop training methods that can incorporate temporal learning directly within the ANN training phase or to refine the conversion process further to better capture these dynamics.

\subsection{Direct Supervised Learning}
Supervised learning involves training models with labeled datasets where each input is associated with a correct output. For SNNs, this paradigm enables the network to learn precise spike patterns that correspond to specific target outputs. This section focuses on the methodologies and advancements in training SNNs using supervised learning, with a particular emphasis on techniques analogous to backpropagation.

Training deep SNNs using supervised learning often encounters the well-known challenge of exploding and vanishing gradients. During the backwards pass, the gradients used to update the weights can either grow uncontrollably (exploding gradients) or shrink towards zero (vanishing gradients), leading to unstable or inefficient learning \cite{neftci2019surrogate}. When gradients become too small, learning halts because weight updates become negligible, while excessively large gradients lead to destabilizing weight updates. This issue is not unique to SNNs, but rather a broader challenge faced by all deep networks trained via backpropagation, including traditional ANNs \cite{bengio1994learning}. In deep architectures, the accumulation of gradients over many layers or time steps can exacerbate this problem \cite{glorot2010understanding}. Several works discussed in this section describe methods that have been developed to prevent vanishing or exploding gradients.

\subsubsection{Temporal Backpropagation}
\label{sec:bptt}
Temporal Backpropagation, also known as Backpropagation Through Time (BPTT), extends the traditional backpropagation algorithm to handle the temporal dynamics of SNNs. This technique involves calculating the temporal derivatives of spike trains and propagating errors backward through time.

Among the first supervised learning rules based on the firing time of individual spikes was the SpikeProp algorithm \cite{bohte2000spikeprop}. SpikeProp used the idea of classic error signal back-propagating inspired by second-generation ANNs and was derived for multi-layer networks used in classification tasks. The error function uses the time difference between the desired output spike time and the actual output spike time of a given output neuron. It is defined using the least mean squares error function as:

\begin{equation}\label{eq:spikeprop}
E=\frac{1}{2}\sum_{j \in J}(t_{a}^{j}-t_{d}^{j})^2
\end{equation}

The calculated error signal can then be back-propagated through the deep network layers to tune synaptic weights.

Several modifications to the SpikeProp algorithm have been proposed, each aiming to improve the classification accuracy of the original. Among these variants are Backpropagation with momentum \cite{xin2001supervised}, QuickProp \cite{mckennoch2006fast}, and Resilient Propagation \cite{mckennoch2006fast}.

In \cite{gutig2006tempotron}, the tempotron is introduced as a supervised learning rule for LIF neurons to classify spatiotemporal spike patterns based on precise spike timings. The tempotron was named in analogy to the perceptron because it extends the classical perceptron model to handle temporal spike patterns in SNNs. The tempotron learning rule adjusts synaptic weights upon classification errors, proportional to the difference between the maximum membrane potential during stimulus presentation and the firing threshold. This voltage-based synaptic adjustment enables the neuron to attribute errors to individual synapses based on their temporal contributions. 
The authors demonstrate that the tempotron can learn to classify input patterns encoded in precise spike timings, such as spike latencies, even when mean firing rates are identical across classes. The tempotron's classification capacity scales linearly with the number of synapses, achieving a maximal load of approximately $\alpha = 3$ patterns per synapse, exceeding the classical percetron's capacity of $\alpha = 2$ \cite{rosenblatt1958perceptron}.

In \cite{ponulak2010supervised}, the ReSuMe training algorithm is discussed. ReSuMe is a training algorithm for SNNs composed of a front subnetwork and an output layer, where the front subnetwork can be feedforward, recurrent, or hybrid. It uses teacher neurons to provide desired spike timings without being part of the SNN's connections. The algorithm adjusts the output neuron weights to make the generated spike train resemble the teacher neuron's spike train. The weight update rule considers correlations between spike trains and uses a learning window for time delay adjustments.

In \cite{tavanaei2019bp}, a supervised learning method that combines the biological realism of STDP with the computational efficiency of backpropagation is discussed. The key innovation is to approximate backpropagation weight updates using a temporally local STDP rule in a network of IF neurons. This method addresses the challenge of training SNNs in a biologically plausible manner while retaining the performance benefits of traditional gradient descent techniques. The authors first demonstrate that IF neurons can approximate ReLU. Using this approximation, they develop a learning rule where weight changes are driven by the error signal and spike timings. The proposed BP-STDP algorithm was evaluated on the MNIST dataset to achieve 97\% accuracy with a three-layer architecture.

In \cite{kheradpisheh2020temporal}, the authors describe a supervised learning rule, S4NN, for multi-layer SNNs using a rank-order temporal coding scheme. The S4NN rule, which is analogous to traditional error backpropagation but operates based on spike latencies, calculates the error at the output layer based on the difference between the actual firing time and the target firing time, and this error is propagated backward through the network layers. S4NN achieved a test accuracy of 97.4\% on the MNIST dataset and 99.2\% on the Caltech Face/Motorbike dataset with a supervised multi-layer fully-connected SNN.

In \cite{lee2016training}, Backpropagation Through Time (BPTT) is adapted for SNNs to demonstrate efficacy in training for temporal tasks. By treating the membrane potentials of spiking neurons as differentiable signals and considering discontinuities at spike times as noise, the authors enable the use of gradient-based optimization tasks. Weight decay and threshold regularization techniques are introduced to stabilize the training process and prevent vanishing/exploding gradients. The method was tested on MNIST and N-MNIST benchmarks, achieving 98.77\% and 96.66\%, respectively.

In \cite{shrestha2018slayer}, the "SLAYER: Spike Layer Error Reassignment in Time" method is introduced. SLAYER addresses the challenge of the non-differentiability of the spike generation function. SLAYER uses a temporal credit assignment policy to backpropagate errors through preceding layers over time, allowing for simultaneous learning of synaptic weights and axonal delays. The authors have released a GPU-accelerated software implementation of SLAYER to enable the training of both FC and conv SNN architectures.

\begin{figure}[b!]
\centering
\includegraphics[scale=0.32]{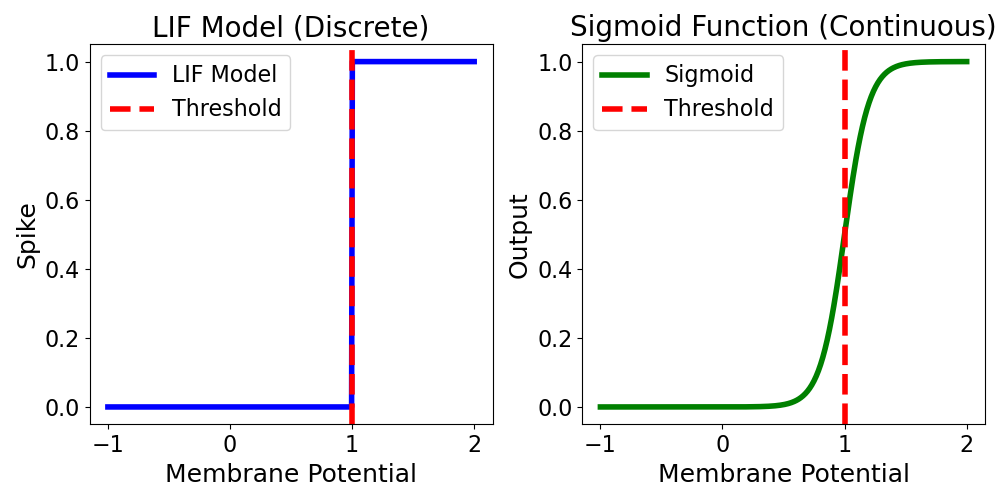}
\caption{Side-by-side comparison of spike generation in the LIF model and its surrogate gradient approximation using the sigmoid function. The LIF model (left) shows discrete, binary spikes generated when the membrane potential crosses a threshold of 1. In contrast, the sigmoid function (right) provides a smooth, continuous approximation of the same threshold behavior, enabling gradient-based optimization during backpropagation in SNNs. The red dashed line indicates the threshold for spike generation in both models.}
\label{fig:lif_vs_sigmoid}
\end{figure}

\subsubsection{Surrogate Gradients}
SNNs process information through discrete spike events, emulating the binary firing behavior of biological neurons. However, this process is inherently non-differentiable because spikes are binary events - neurons either fire (output a 1) or do not fire (output a 0) - creating a discontinuity that makes it impossible to compute gradients directly \cite{lee2020enabling}. In deep learning, differentiability is crucial for computing gradients during backpropagation to update network weights and minimize loss functions \cite{rumelhart1986learning}.

Traditional ANNs use continuous activation functions like sigmoid or ReLU, enabling gradient computation due to their smooth nature \cite{lecun2015deep}. To enable gradient-based training in SNNs, surrogate gradient methods approximate the non-differentiable spiking function with a continuous and differentiable surrogate during the backwards pass \cite{neftci2019surrogate}. A commonly use surrogate is the sigmoid function, defined as:

\begin{equation}
    \sigma(x) = \frac{1}{1 + e^{-x}}
\end{equation}

where $x$ is the difference between the neuron's membrane potential and its firing threshold. The sigmoid function, as shown in Figure \ref{fig:lif_vs_sigmoid}, provides a smooth approximation near the threshold, allowing gradients to be computed and propagated back through the network. By substituting the spiking non-linearities with their surrogate derivatives during backpropagation, standard gradient descent algorithms can be applied to effectively train SNNs.

Several works have applied surrogate gradient methods to train SNNs.

In \cite{nicola2017}, the FORCE method was proposed as a training method for rate-based neural networks. FORCE training combines static chaotic synaptic weights with dynamically learned weights, optimized using Recursive Least Squares (RLS) to minimize the error between the network's output and the supervisory signal. The study demonstrates the use of FORCE training across different neuron models (LIF, Izhikevich) and tasks.

In \cite{zenke2018}, the SuperSpike method is proposed to tackle the challenge of supervised learning in SNNs, particularly focusing on training multi-layer networks of deterministic LIF neurons. The core contribution is the derivation of SuperSpike, a nonlinear voltage-based three-factor learning rule using a surrogate gradient approach. The authors address the issue of non-differentiability in spiking neurons by replacing the spike train with a continuous auxiliary function of the membrane potential, enabling gradient-based optimization. The authors compare SuperSpike learning under different credit assignment strategies, including uniform, symmetric, and random feedback. The results indicated that while simpler tasks could be solved with any feedback type, complex tasks benefit from symmetric feedback. The paper highlights the biological plausibility of the SuperSpike rule, which incorporates Hebbian-like terms, voltage-dependence, and eligibility traces akin to calcium transients in synapses.

In \cite{wu2018spatio}, a spatio-temporal backpropagation (STBP) algorithm is introduced alongside an iterative LIF model suitable for gradient descent training. The STBP algorithm approximates the derivative of spike activities with several smooth functions, including rectangular, polynomial, sigmoid, and Gaussian cumulative distribution functions. The experimental results demonstrated that the STBP algorithm achieved high accuracy on MNIST and N-MNIST datasets, with accuracy scores of 98.89\% and 98.78\%, respectively.

In \cite{neftci2019surrogate}, the authors propose a surrogate gradient supervised learning rule in which the membrane potential of a neuron is tracked over time. The firing rules of a LIF activation function are based on whether the membrane potential of a neuron has crossed the neuron's voltage threshold. This rule is used for the forward pass of the network, where spikes are generated according to the LIF firing rules. For the backward pass, the activation function is substituted for a sigmoid function, which is continuous and differentiable. This method is accessible through the Auto-Differentiation tools provided by popular deep-learning libraries such as PyTorch and TensorFlow.

In \cite{lee2020enabling}, an approximate derivative method is proposed for the LIF neuron model, which incorporates the neuron's leaky behavior into the gradient descent algorithm. This pseudo-derivative enables the use of backpropagation in SNNs, allowing for the direct training of deep SCNNs. The authors discuss the use of well-known deep learning models such as VGG and ResNet. The method includes the use of a straight-through estimator and a leak correction term to approximate the derivative of the LIF neuron's activation function. This allows for the backpropagation of errors through the network layers, allowing the training of very deep SNNs. The experimental results show that deep SNNs trained with this method achieve accuracy comparable to that of their ANN counterparts. On the N-MNIST dataset, the proposed method achieved an accuracy of 99.09\%, slightly below the best-reported accuracy of 99.53\%. For the CIFAR-10 dataset, three network architectures were evaluated: VGG9, Resnet9, and ResNet11, achieving 90.45\%, 90.35\%, and 90.95\%, respectively.

\subsubsection{Discussion}
The exploration of supervised learning methodologies in SNNs presents innovations, particularly those adapting backpropagation techniques to the unique temporal dynamics of spiking neurons. Temporal backpropagation algorithms like SpikeProp introduced error propagation based on spike timings, marking a significant departure from rate-based methods in traditional NNs. These algorithms primarily address the challenge of the non-differentiability of spike events, using approaches such as surrogate gradients and temporal credit assignment mechanisms. 

Building on these foundational techniques, the field has seen significant advancements aimed at enhancing the stability and effectiveness of SNN training. For instance, temporal coding schemes have been developed to use spike latencies instead of firing rates, aligning more closely with the natural functioning of biological neurons. Additionally, the adaptation of traditional deep learning techniques such as BPTT allows researchers to address critical challenges like vanishing and exploding gradients to ensure a more stable and efficient training process. Surrogate gradients have been shown to approximate the gradient of spike activities and enable the application of backpropagation-like algorithms. In this way, SNNs can benefit from gradient-based optimization tools and integrate with popular DL frameworks like PyTorch and TensorFlow.

Innovative methods for approximating derivatives in spiking neurons have demonstrated the feasibility of training very deep SNNs, achieving accuracy levels comparable to traditional NNs. These approaches maintain gradient flow through deep network layers, which is essential for scaling SNNs to handle more complex tasks.

\begin{table*}[t!]
    \centering
    \caption{Summary of SNN Architectures}
    \begin{tabular}{|p{1.5cm}|p{3cm}|p{3cm}|p{4cm}|}
    \hline
    \textbf{Architecture} & \textbf{Description} & \textbf{Key Characteristics} & \textbf{Training Methodologies}
    \\
    \hline
    \textbf{FCSNNs} \cite{tapson2013synthesis, xin2001supervised, dhoble2012online, cohen2016skimming, krithivasan2019dynamic, diehl2015unsupervised, morrison2007spike, iyer2017unsupervised, brader2007learning, bengio2015towards, manna2022simple, putra2020fspinn} & Fully connected SNNs with fan-out connections & Neurons interconnect with every neuron in the subsequent layer & Supervised (e.g., SKIM, backpropagation) and Unsupervised (e.g., STDP) \\ 
    \hline
    \textbf{HSNNs} \cite{zhao2014feedforward, xiao2019event, lu2020event, negri2018scene, liu2020effective, serre2007robust, orchard2015hfirst, diehl2015unsupervised, liu2020unsupervised, zhou2022bio, kheradpisheh2016bio}  & Hierarchical SNNs using predefined filters like Gabor filters & Layered architecture (S1, C1, S2, C2), lateral inhibition & Supervised (e.g., Tempotron learning rule) and Unsupervised (e.g., STDP) \\ 
    \hline
    \textbf{SCNNs} \cite{kheradpisheh2018stdp, panda2016unsupervised, tavanaei2018training, lee2016training, cordone2021learning, jin2018hybrid, meng2022training, xiao2022online, yao2022attention, guo2022real, fang2021deep, samadzadeh2023convolutional, perez2013mapping, hunsberger2016training, rueckauer2016theory, lu2020exploring, singh2021gesture, diehl2015fast, ding2021optimal, esser2015backpropagation, cao2015spiking, massa2020efficient} & Spiking Convolutional Neural Networks designed for spatial data & Convolutional layers with learnable filters, capturing temporal dynamics & Supervised (e.g., backpropagation) and Unsupervised (e.g., STDP) 
    \\
    \hline
    \textbf{SDBNs} \cite{neftci2014event, lee2007sparse, lee2011unsupervised, kaiser2017spiking, o2013real, stromatias2015robustness, stromatias2015scalable, merolla2011digital, neil2014minitaur} & Spiking Deep Belief Networks with multiple layers of spiking neurons & Composed of sRBMs, trained sequentially using Contrastive Divergence & Unsupervised pre-training (e.g., CD), Supervised fine-tuning 
    \\
    \hline
    \textbf{SRNNs} \cite{shrestha2017spike, diehl2016conversion, kim2019simple, lotfi2020long, xing2020new, neil2016phased, yin2020effective} & Spiking Recurrent Neural Networks with feedback connections & Recurrent processing of sequential data, capturing temporal dependencies & Supervised (e.g., BPTT), Hybrid methods (e.g., "train-and-constrain") 
    \\
    \hline
    \textbf{LSMs}\cite{wang2016d, patino2022liquid, kasabov2014neucube, paulun2018retinotopic}  & Liquid State Machines using spiking neurons for temporal pattern recognition & Large, sparsely connected reservoir of neurons, dynamic spatio-temporal patterns & Unsupervised (e.g., STDP), Supervised training of readout layer 
    \\
    \hline
    \end{tabular}
    \label{tab:snn_architectures}
\end{table*}

\section{Architecture}
\label{sec:architecture}
This section summarizes the literature on SNN architectures, including Fully Connected SNNs, Hierarchical SNNs, Spiking Convolutional Neural Networks, Spiking Recurrent Neural Networks, and Spiking Deep Belief Networks. These architectures, designed with distinct structural and functional properties, address different challenges in neural computation, such as feature extraction, spatial-temporal processing, and learning efficiency. This section references various learning rules associated with these SNN architectures, which are discussed in detail in Section \ref{sec:learning_rules}.

The development trajectory of SNN architectures has closely paralleled that of ANNs, leading to sophisticated paradigms for feature extraction and signal propagation. The SNN landscape can be characterized as a diverse collection of archetypes; each uniquely adapted to specific computational tasks and data properties. Fully connected SNNs (\textbf{FCSNNs}) are the foundational archetype, analogous to their ANN counterparts, wherein each neuron in one layer is interconnected with every neuron in the succeeding layer. This architecture supports a broad and generalized approach to data processing, allowing for a wide range of applications but without specific optimizations for any one type of data structure. Hierarchical SNNs (\textbf{HSNNs}) are a class of network variants that use filters with predefined weights, such as Gabor filters, for convolution to allow for specialized processing of spatial features. Spiking convolutional neural networks (\textbf{SCNNs}) represent another archetype specifically designed to learn spatial data from images. These networks exploit local spatial correlations through learnable convolutional filters, thus efficiently processing data with inherent spatial hierarchies while also capturing temporal dynamics through the propagation of spike trains. Spiking recurrent neural networks (\textbf{SRNNs}) are an archetype optimized for sequential data processing. By incorporating feedback loops that allow the network to maintain a form of memory, SRNNs are particularly suited to tasks that require understanding temporal dependencies. Spiking deep belief networks (\textbf{SDBN}) are a generative model approach within the SNN framework. These networks use a layered architecture of stochastic units to model complex distributions and uncover latent structures in data, making them ideal for tasks involving unsupervised learning or feature discovery in high-dimensional spaces. Table \ref{tab:snn_architectures} summarizes the key aspects of different SNN architectures, their defining characteristics, training methodologies, and notable works. For hands-on examples, the accompanying open-source repository contains Jupyter notebooks demonstrating the implementation of various SNN architectures, including both supervised and unsupervised learning examples. These examples showcase feedforward and spiking convolutional models, as well as techniques for applying encoding to static images \cite{Iaboni2024EventSNN}.

\begin{table*}[h]
    \centering
    \caption{Summary of studies on supervised and unsupervised FCSNNs.}
    \begin{tabular}{|p{1.5cm}|p{1.5cm}|p{2cm}|p{9cm}|}
        \hline
        \textbf{Training} & \textbf{Study} & \textbf{Method} & \textbf{Key Findings} \\ \hline
        \multirow{5}{*}{\textbf{Supervised}} & Tapson et al. (2013) \cite{tapson2013synthesis} & Synaptic Kernel Inverse Method (SKIM) & Introduced SKIM for spatio-temporal pattern recognition, using non-linear synaptic response functions to process spike-time data. \\ \cline{2-4}
        & Xin et al. (2001) \cite{xin2001supervised} & Sigmoidal SNN with adapted error backpropagation & Achieved 97.22\% accuracy using a modified backpropagation algorithm to adjust synaptic weights based on spike time differences. \\ \cline{2-4}
        & Dhoble et al. (2012) \cite{dhoble2012online} & Dynamic evolving SNN (deSNN) & Combined rank-order and temporal spike coding with SDSP rule for real-time pattern recognition, evaluated on a simple dataset. \\ \cline{2-4}
        & Cohen et al. (2016) \cite{cohen2016skimming} & SKIM with various training patterns and output methods & Achieved 92.87\% accuracy with Flat training pattern and 10,000 neurons, highlighting the effectiveness of different training patterns and output determination methods. \\ \cline{2-4}
        & Krithivasan et al. (2019) \cite{krithivasan2019dynamic} & Dynamic Spike Bundling (DSB) in B-SNNAP accelerator & Achieved 3.8x reduction in energy consumption with only a 0.1\% accuracy decrease by bundling closely occurring events into single spikes. \\ \hline
        \multirow{5}{*}{\textbf{Unsupervised}} & Diehl \& Cook (2015) \cite{diehl2015unsupervised} & Two-layer network for unsupervised learning of Poisson spike-train inputs & Demonstrated a network structure with 784 input neurons and 100 excitatory/inhibitory neurons, emphasizing lateral inhibition for pattern learning. \\ \cline{2-4}
        & Morrison et al. (2007) \cite{morrison2007spike} & STDP model in balanced random network & Introduced a novel STDP rule with multiplicative synaptic depression and power law potentiation, focusing on maintaining asynchronous irregular activity. \\ \cline{2-4}
        & Iyer et al. (2017) \cite{iyer2017unsupervised} & Two-layer network with modified learning rules for N-MNIST dataset & Achieved 80.63\% accuracy with polarized approach and specific learning rules, highlighting the importance of adjusting membrane time constants and spike frequency adaptation. \\ \cline{2-4}
        & Brader et al. (2007) \cite{brader2007learning} & Spike-driven synaptic plasticity in semi-supervised learning & Achieved 97.1\% accuracy with a model featuring IF neurons and selective teacher signals for output neuron populations. \\ \cline{2-4}
        & Bengio et al. (2015) \cite{bengio2015towards} & STDP as stochastic gradient descent for deep generative networks & Proposed interpreting STDP within a variational EM framework, using approximate inference and target propagation as an alternative to gradient backpropagation. \\ \cline{2-4}
        & Manna et al. (2022) \cite{manna2022simple} & Comparison of IF neuron models (LIF, QIF, EIF) & Evaluated performance of neuron models on classification tasks, with LIF achieving 93\% accuracy on binary classification (0 vs 1) for N-MNIST and EIF outperforming others in rich temporal dynamics scenarios for DVS-Gestures. \\ \cline{2-4}
        & Putra et al. (2020) \cite{putra2020fspinn} & FSpiNN framework for optimizing SNNs for energy and memory efficiency & Achieved 98.5\% and 89.6\% accuracy for MNIST and Fashion MNIST respectively, with significant reductions in memory use and enhanced energy efficiency through simplified neuronal activity and STDP processes. \\ \hline
    \end{tabular}
    \label{tab:fc_snns}
\end{table*}

\subsection{FCSNNs}
In FCSNNs, each neuron in one layer is connected to every neuron in the subsequent layer, a concept known as ``fan-out''. The fan-out refers to the number of synaptic connections a single neuron forms with downstream neurons. In FCSNNs, the fan-out of each neuron is equal to the number of neurons in the subsequent layer, such that each neuron's output (post-synaptic potential) is transmitted to every downstream neuron \cite{davidson2021comparison}. The fan-out characteristic is significant because it impacts the network's computational power and its ability to generalize from the training data. Higher fan-out can lead to more robust dissemination of information and allow more complex patterns to be learned \cite{furber2012overview}. However, it also increases the computational load and the potential for overfitting, which must be managed through techniques such as dropout or regularization \cite{zhao2021spiking}. The typical structure of a FCSNN often consists of an input layer, zero or one hidden layer, and an output layer. By maintaining shallow architectures, the training complexity and hardware requirements are managed. Training FCSNNs can be approached through both supervised and unsupervised learning paradigms. As shown in Table \ref{tab:fc_snns}, various studies have explored supervised and unsupervised FCSNNs.

\subsubsection{Supervised FCSNNs}

\noindent These works focus on the development of NN methodologies or enhancements using simple datasets. The focus here is more on the methodological innovations rather than the complexity of the application domain.

\begin{itemize}
        
    \item In \cite{tapson2013synthesis}, the authors introduce the Synaptic Kernel Inverse Method (SKIM) for spatiotemporal pattern recognition. This paper marks the first application of SKIM to a large-scale, multi-class problem. SKIM is a neural synthesis algorithm designed to create neurons that specifically respond to predefined spatiotemporal patterns of input spikes. This characteristic makes it highly suitable for processing information encoded in spike times. The architecture of SKIM consists of input and output neurons, where traditional hidden layers are replaced by direct synaptic connections. Each sensor pixel from the input is mapped to a distinct neuron in the input layer. The synapses have weights initially set to random values and are assigned to specific dendritic branches. The key feature of these synapses is their implementation of non-linear response functions - such as exponential or decaying-alpha functions - to incoming electrical currents from the dendritic branches. This non-linear processing is crucial for converting the spike-time input data into continuous values that can be further handled by the network.

    \item In \cite{xin2001supervised}, the classical error backpropagation algorithm is adapted for use in a Sigmoidal SNN. This adaptation is significant because traditional backpropagation relies on differentiable activation functions, whereas spike events are inherently non-differentiable due to their discrete nature. By employing a sigmoidal activation function to approximate spiking behavior, the network can utilize gradient-based optimization techniques, facilitating more effective supervised learning in spiking models. The network architecture consists of a feedforward structure with multiple layers of interconnected synapses, each having varying delays. The proposed algorithm operates by adjusting synaptic weights based on the differences between actual and target spike times to minimize a least-squares error function. This method was tested on the Iris dataset and achieved an accuracy of 97.22\%.

    \item In \cite{dhoble2012online}, a dynamic evolving SNN (deSNN) is introduced. This network uniquely integrates both rank-order spike coding and temporal spike coding through the implementation of a Spike Driven Synaptic Plasticity (SDSP) rule. Traditional evolving SNNs (eSNNs) evolve by dynamically generating new spiking neurons and synaptic connections in response to incoming data, enabling them to recognize patterns in an online, real-time manner. The deSNN model enhances this framework by incorporating rank-order spike coding, where inputs are prioritized based on the sequence and timing of spikes—inputs that elicit earlier spikes have a greater influence, adhering to the rank-order learning rule. The integration of the SDSP rule allows synaptic weights to be adjusted based on the precise timing of pre- and post-synaptic spikes, effectively combining unsupervised and supervised learning elements. The deSNN method was rigorously evaluated on a simple two-class detection dataset to demonstrate its efficacy.

\end{itemize}

\noindent While the above works introduced the concept of FCSNN, they apply it to a limited number of datasets. In contrast, the following studies apply supervised FCSNN techniques to more complex real-world datasets involving larger networks and more detailed analysis of the efficiency and effectiveness of handling more challenging data.

\begin{itemize}
    
    \item In \cite{cohen2016skimming}, the Synaptic Kernel Inverse method (SKIM) is applied to the N-MNIST dataset. The study explored various training patterns and output determination methods within the SKIM framework. Using the SKIM method, input neurons are connected directly to output neurons with fan-out synapses, eliminating a clear distinction between input, hidden, and output layers. The authors note that a solitary spike fails to generate a sufficiently strong and lasting error signal necessary for efficient learning. To address this limitation, they proposed using training patterns that span multiple time steps, effectively distributing the learning signal over a temporal window.

    The researchers experimented with three training patterns:
    \begin{enumerate}
        \item {\bf Flat Pattern:} A constant signal over time.

        \item {\bf Gaussian Pattern:} A bell-shaped curve.

        \item {\bf Exponential Pattern:} Starting at maximum and exponentially decaying.
    \end{enumerate}

    Additionally, four output determination methods were explored:
    \begin{enumerate}
        \item {\bf Max Method:} Selecting the class corresponding to the maximum output value.

        \item {\bf Area Method:} Selecting the class with the highest total area under the output curve.

        \item {\bf Weighted Max Method:} Weighing the output in proportion to the used training pattern before identifying the maximum.

        \item {\bf Weighted Area Method:} Selecting the class with the largest weighted area under the curve.
    \end{enumerate}

    Notably, the weighted approaches displayed no additional effect when paired with the Flat training pattern but did have an effect when paired with the Gaussian or Exponential training patterns. The N-MNIST dataset was used for all experiments performed in this work. The highest accuracy of 92.87\% was achieved with a network size of 10,000 neurons. The experimental results showed that the Flat training pattern yielded superior outcomes, except when paired with the Max output determination method. In contrast, the Gaussian training pattern resulted in the lowest performance in all tested cases.

    \item In \cite{krithivasan2019dynamic}, the Dynamic Spike Bundling (DSB) method is introduced to enhance the efficiency of spike processing in SNNs. This method dynamically bundles event triggers that occur closely in time, consolidating them into a single event to represent the entire group. By employing DSB, the frequency of memory accesses and the volume of communication between neurons are significantly reduced, leading to improved computational efficiency. The DSB technique was implemented within a hardware accelerator architecture known as B-SNNAP (Bundled Spiking Neural Network Accelerator Platform), specifically designed to support dynamic spike bundling. The evaluation of the DSB method was conducted using four image recognition datasets: MNIST, CIFAR-10, CIFAR-100, and ImageNet. The results demonstrated that the proposed method achieved a 3.8x reduction in energy consumption with only a 0.1\% decrease in accuracy. This substantial reduction in energy use was attributed to the decreased memory accesses and communication traffic facilitated by the DSB technique.
\end{itemize}

\subsubsection{Unsupervised FCSNNs}
This group of studies explores various architectures for unsupervised learning in SNNs, particularly focusing on how these networks can  learn and process information in a manner analogous to the human brain.

\noindent These studies primarily deal with the interaction between neurons within a network and how different types of neurons and their connectivity influence the overall dynamics and processing capabilities of a network.
\begin{itemize}
\item In \cite{diehl2015unsupervised}, Diehl and Cook propose a two-layer SNN for the unsupervised learning of Poisson-distributed spike train input patterns, applied to the MNIST dataset. The input layer consists of 784 neurons, corresponding to each pixel of a 28 $\times$ 28 image. The second layer, referred to as the processing layer, comprises 100 excitatory neurons and 100 inhibitory neurons. Each excitatory neuron is connected to an inhibitory neuron in a one-to-one manner, forming excitatory-inhibitory pairs. Each inhibitory neuron is then connected to all excitatory neurons except the one it received input from, creating a lateral inhibition mechanism. This structure promotes competition among excitatory neurons, encouraging them to specialize in responding to different input patterns. The firing rates of the input neurons are proportional to the intensity of the corresponding pixels in the input image, converting pixel intensities into spike rates using Poisson encoding. Using this architecture, a classification accuracy of up to 95\% was achieved with 6400 excitatory neurons. The performance increased with the number of neurons. With 100 neurons, the network achieved around 83\% accuracy. With 400 neurons, it achieved around 87\%. With 1600 neurons, it reached around 92\%.

\item In \cite{morrison2007spike}, the authors implement their model of Spike-Timing-Dependent Plasticity (STDP) within a balanced random network. The model comprises IF neurons interconnected randomly with both excitatory and inhibitory synapses. This network ensures a realistic level of sparsity and connectivity, approximating the synaptic architecture found in cortical tissues. The authors introduce a novel STDP rule where synaptic depression is multiplicative—dependent on the existing synaptic weight—and potentiation follows a power-law relationship relative to the synaptic weight. The synaptic updates depend on the precise timing of pre- and post-synaptic spikes, integrating both nearest-neighbor and all-to-all spike interactions to evaluate the effects of different pairing schemes on synaptic plasticity. Instead of using external datasets, the study relies on simulated neural activity generated within the model itself. The inputs, connections, and spike timings are controlled to mimic conditions expected in real cortical tissues. The simulated neural network behavior is evaluated to determine whether the network can maintain asynchronous irregular (AI) activity typical of cortical neurons without developing overly structured synaptic connections.

\end{itemize}

\noindent The studies below focus on how different learning rules and network adjustments affect the performance of FCSNNs on specific tasks.
\begin{itemize}

    \item In \cite{iyer2017unsupervised}, the authors extend the Diehl and Cook \cite{diehl2015unsupervised} model to the neuromorphic N-MNIST dataset, exploring the impact of using and discarding polarity information within the dataset. In the unpolarized case, polarity information is disregarded, treating all events equally, regardless of whether they are "ON" or "OFF." In the polarized case, only "ON" events are considered, while "OFF" events are ignored. The study uses a two-layer network architecture. The first layer, consisting of 34 $\times$ 34 neurons, corresponds to the dimensions of the N-MNIST dataset and serves as the input layer. This layer is connected to an excitatory layer, which in turn is connected one-to-one with an inhibitory layer to provide lateral inhibition among neurons. 
    
    The study experiments with two distinct learning rules for adjusting synaptic weights based on observed spike timings:
    \begin{enumerate}
        \item Learning Rule 1: Based on Diehl and Cook's original rule, this rule increases the synaptic trace to 1 immediately after each presynaptic spike and decays exponentially afterward. It is sensitive to the precise timing between presynaptic and postsynaptic spikes, adhering to the principles of Spike-Timing-Dependent Plasticity (STDP)

        \item Learning Rule 2: Each presynaptic spike increases the synaptic trace by a fixed amount and decays more slowly, allowing the trace to accumulate with higher presynaptic spike rates. This rule is less sensitive to the exact timing between pre- and postsynaptic spikes and instead emphasizes the overall spike rate.
    \end{enumerate}

    The authors modify the membrane time constant to reduce the neuron's excitability compared to the original parameter setting in \cite{diehl2015unsupervised}. Lower excitability ensures that the neuron processes the entire pattern before firing, which is crucial for the N-MNIST dataset, where spike rates are not homogeneously distributed. To adjust spike frequency adaptation, the neuron's firing threshold is adapted based on its activity. After a neuron fires, its threshold is temporarily increased, making it harder for the neuron to fire again immediately. The adaptation term increases with each spike the neuron fires and decays exponentially over time, allowing the neuron to gradually return to its original sensitivity after a period of reduced activity. The N-MNIST dataset was used to evaluate the proposed method. The results indicated that the polarized approach combined with Learning Rule 2 led to the best accuracy of 80.63\% on an 800-neuron network.

    \begin{table*}[ht]
    \centering
    \caption{Summary of HSNN Architectures}
    \begin{tabular}{|p{1.5cm}|p{1cm}|p{4cm}|p{2cm}|p{3cm}|}
        \hline
        \textbf{Study} & \textbf{Layers} & \textbf{Features} & \textbf{Learning Rule} & \textbf{Notes} \\ \hline
        Masquelier and Thorpe \cite{masquelier2007unsupervised} & S1, C1, S2, C2 & Gabor filters, lateral inhibition, template matching & Unsupervised & Improved on original HMAX model \\ \hline
        Zhao et al. \cite{zhao2014feedforward} & S1, C1 & Gabor filters, temporal contrast AER sensor, winner-take-all & Supervised & Focus on human posture categorization \\ \hline
        Xiao et al. \cite{xiao2019event} & S1, C1 & Time-surface encoding, noise-reduction mechanism & Supervised & Emphasizes noise-reduction \\ \hline
        Lu et al. \cite{lu2020event} & S1, C1 & Event-driven convolutions, temporal winner-take-all & Supervised & Highlights prominent events \\ \hline
        Negri et al. \cite{negri2018scene} & S1, C1 & Scene context classification & Supervised & Improved accuracy with larger hidden layer \\ \hline
        Liu et al. \cite{liu2020effective} & S1, C1 & Segmented probability-maximization (SPA) & Supervised & Multi-tasking framework \\ \hline
        Serre et al. \cite{serre2007robust} & S1, C1, S2, C2 & Gabor filters, local max responses, RBF & Supervised & Robust object recognition \\ \hline
        Orchard et al. \cite{orchard2015hfirst} & S1, C1, S2, C2 & TTFS, lateral reset mechanism & Supervised & Time to first spike approach \\ \hline
        Liu et al. \cite{liu2020unsupervised} & S1, C1 & Logarithmic temporal coding, multiscale spatio-temporal representation & Unsupervised & Unsupervised object recognition \\ \hline
        Zhou et al. \cite{zhou2022bio} & S1, C1, S2, C2 & Noise filtering, event stream segmentation & Unsupervised & Multi-stage object recognition \\ \hline
        Kheradpisheh et al. \cite{kheradpisheh2016bio} & S1, C1, S2, C2 & Hierarchical neural network, unsupervised learning & Unsupervised & Invariant recognition capabilities \\ \hline
    \end{tabular}
    \label{tab:hsnn-summary}
\end{table*}

    \item In \cite{brader2007learning}, the authors present a model of spike-driven synaptic plasticity designed for classifying input stimuli in a semi-supervised manner. The network architecture consists of integrate-and-fire neurons with a constant leak value and features a shallow structure with a single feedforward layer. This layer includes $N_{input}$ input neurons fully connected by plastic synapses to $N_{output}$ output neurons. The output neurons lack lateral connections and are divided into populations, each responsive to specific classes of stimuli. Additional signals to the output neurons come from inhibitory and teacher populations. The inhibitory neurons provide signals proportional to the stimulus to compensate for large variations in stimulus coding levels. The teacher population, active during training, enhances the selectivity of output pools by supplying extra excitatory signals. The inclusion of this teacher population enables semi-supervised learning by providing guidance during training. The model was tested with 2000 input neurons, an inhibitory population of 1000 neurons, and a teacher population of 20 neurons. This network achieved a classification accuracy of 97.1\% on the MNIST dataset. The network's shallow architecture and lack of lateral connections among output neurons simplify the model, reducing computational complexity and facilitating faster processing.

    \item In \cite{bengio2015towards}, the authors aim to address the biological implausibility of gradient backpropagation in neural networks. They propose interpreting STDP as a form of stochastic gradient descent, wherein synaptic weight changes are driven by the derivative of the postsynaptic potential. This interpretation situates neuronal dynamics and synaptic updates within the framework of variational Expectation-Maximization (EM). The authors posit that modifications in neural activity serve to optimize an objective function representing a variational lower bound on the log-likelihood, thereby making the neural network function as a generative model. They extend this theoretical framework to propose a training procedure for deep generative networks, which involve multiple layers of latent variables without relying on explicit gradient computation. Instead, they use approximate inference and difference target propagation to estimate gradients for learning. Target propagation is presented as an alternative to backpropagation, where the gradient is estimated across network layers by using differences in target values at each layer to guide updates. The theory and methods were validated through generative tasks performed on the MNIST dataset. 

    \item In \cite{manna2022simple}, the performance of different spiking neuron models was evaluated within an SNN architecture trained using STDP. Specifically, the authors focused on three types of increasingly complex IF neuron models, LIF, Quadratic Integrate-and-Fire (QIF), and Exponential Integrate-and-Fire (EIF) neurons. The performance of these models was tested on classification tasks using the N-MNIST and DVS-Gestures datasets. On the N-MNIST dataset, the more complex models (QIF and EIF) required more parameter tuning but did not consistently outperform the simpler LIF model in terms of accuracy. The best accuracy reported was 93\% for the LIF neuron model on a binary classification task (0 vs 1). On the DVS-Gestures dataset, the complex models (EIF and QIF) outperformed the simpler model in scenarios involving rich temporal dynamics. The best accuracy reported was 77\% for the EIF neuron model on a hand-clapping vs hand-wave classification task.

    \item In \cite{putra2020fspinn}, FSpiNN is introduced as a framework to optimize SNNs for energy and memory efficiency. FSpiNN reduces the computational demand by simplifying the operations involved in neuronal activity and STDP processes. This includes minimizing inhibitory neuron computations through direct lateral inhibition connections and optimizing synaptic weight updates to occur only on postsynaptic spikes, which are more relevant for learning. The framework introduces methods to maintain learning accuracy while reducing computational complexity. FSpiNN implements fixed-point quantization to reduce the bit-width of network parameters. FSpiNN was evaluated using the MNIST and Fashion MNIST datasets, achieving accuracies of 98.5\% and 89.6\%, respectively. The results also indicated significant reductions in memory use (up to 7.5x) and enhanced energy efficiency (up to 3.5x).
\end{itemize}

\subsubsection{Discussion}
The exploration of FCSNNs in this section illustrates their utility as a foundational architecture in the realm of SNNs. Their widespread use in various studies primarily serves as a platform to validate new learning algorithms or to benchmark network performance under controlled conditions. FCSNNs, by virtue of their straightforward architecture consisting of merely input, hidden (optional), and output layers, offer an accessible entry point for implementing and testing novel neuromorphic methods. Their use extends beyond mere proof-of-concept demonstrations to serve as benchmarks for evaluating the robustness and scalability of new SNN algorithms. This benchmarking role is particularly evident in studies that test foundational changes or enhancements in learning paradigms, such as the Synaptic Kernel Inverse Method (SKIM) \cite{cohen2016skimming} and dynamic spike bundling techniques \cite{krithivasan2019dynamic}.

Their simplicity, however, is a double-edged sword. On the one hand, it allows for the emphasis of methodological exploration over architectural complexity, as seen in studies such as \cite{tapson2013synthesis} and \cite{xin2001supervised}. On the other hand, the lack of convolutional layers limits their efficiency in handling complex, high-dimensional data that are commonplace in real-world computer vision tasks. This limitation underscores the intrinsic value of convolutional operations in mimicking the hierarchical and spatially invariant processing capabilities of the biological vision system. To mitigate these limitations, most convolutional architectures aim to leverage the strengths of both configurations - using convolutional layers for feature extraction and fully connected layers for classification tasks.

\subsection{HSNNs}
Interdisciplinary research efforts bridging neuroscience and CV have yielded significant advancements in both fields \cite{rosenblatt1958perceptron, fukushima1988neocognitron, riesenhuber1999, gutig2006tempotron}. These endeavors have led to the development of biologically-inspired visual models that leverage the dynamic capabilities of event-based vision sensors, emphasizing the temporal aspects of neuronal firing \cite{masquelier2007unsupervised, zhao2014feedforward, lu2020event, negri2018scene, liu2020unsupervised, liu2020effective, orchard2015hfirst}. This computational strategy mirrors biological observations where quicker neural responses are indicative of more salient or significant stimuli \cite{hebb2005organization}. The various Hierarchical SNN (HSNN) architectures and their performances are summarized in Table \ref{tab:hsnn-summary}.

A seminal model in this domain is the Hierarchical Max-Pooling (HMAX) model, which was among the first to emulate human-like performance in object recognition tasks, as demonstrated by Riesenhuber and Poggio \cite{riesenhuber1999}. Building upon this foundation, a variant of the HMAX model incorporating spiking neurons and the STDP learning rule was introduced by Masquelier and Thorpe \cite{masquelier2007unsupervised}. This modified model not only adheres more closely to biological plausibility but also surpasses the original HMAX in performance. 

\begin{figure}[t!]
    \centering
    \includegraphics[scale = 0.4]{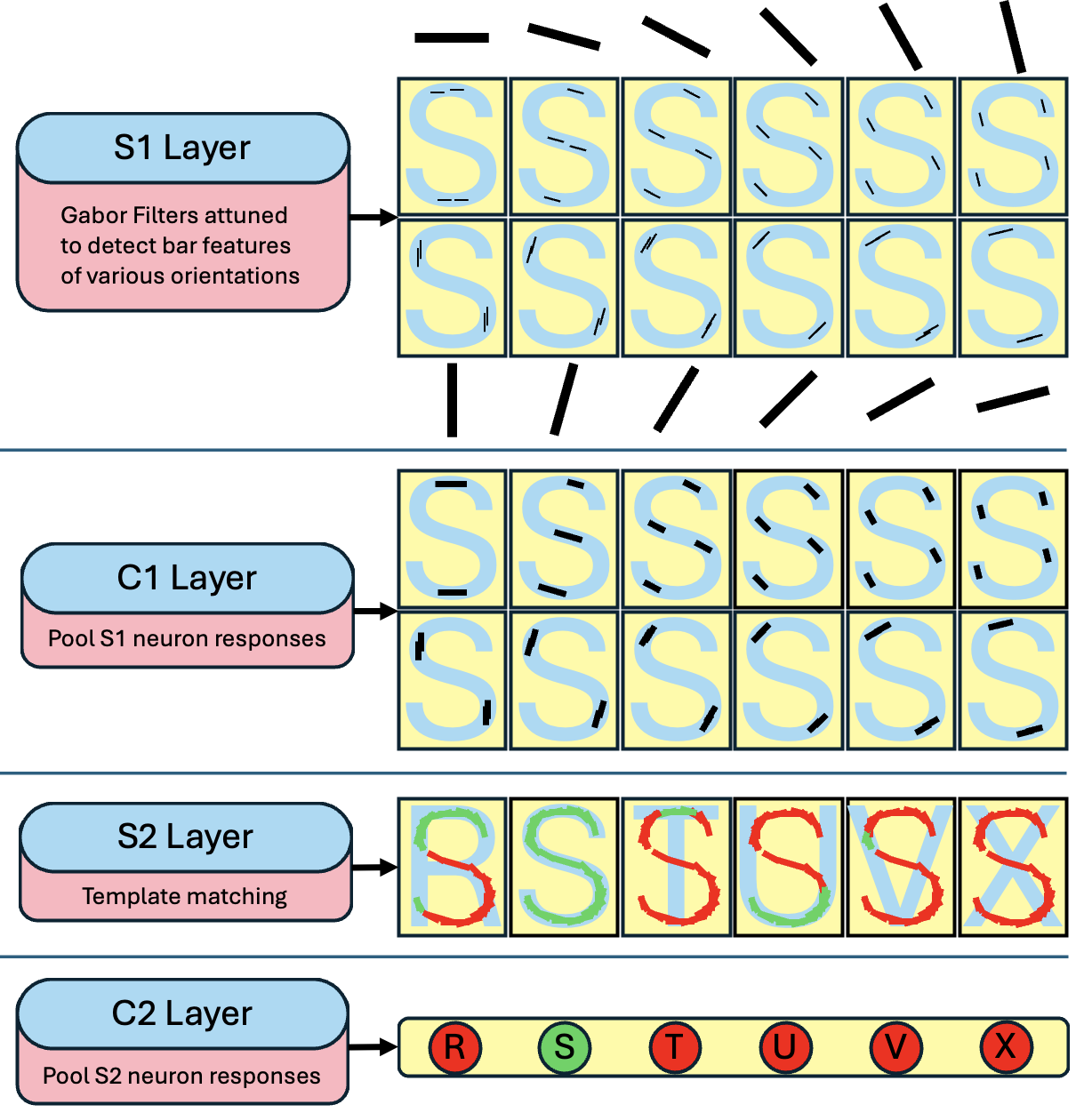}
    \caption{The architecture of a Spiking HMAX-like model is shown, specifically with layers from S1 through C2, adapted from \cite{orchard2015hfirst}. It begins with the S1 layer using Gabor filters to detect bar features at various orientations. The C1 layer then performs pooling and lateral inhibition to enhance feature selectivity. The S2 layer engages in template matching to identify combinations of basic features, and the C2 layer pools these responses. }
    \label{fig:hsnn}
\end{figure}

The HSNN architecture used in Spiking HMAX-like models is structured into layers S1 through C2, as depicted in Figure \ref{fig:hsnn}:
\begin{enumerate}
    \item {\bf S1 Layer:} This initial stage consists of predefined 12 Gabor filters, each specifically tuned to detect bar features at various orientations. This layer is analogous to the simple cells in the primary visual cortex and is organized into S1 units comprising adjacent non-overlapping 4 $\times$ 4 pixel regions. Each S1 unit is designed to feed into a corresponding set of C1 neurons, one for each orientation detected by the S1 filters.

    \item {\bf C1 Layer:} The C1 neurons perform lateral inhibition, ensuring that only the first neuron to spike within each pooling region carries its response forward. This mechanism of lateral inhibition plays a critical role in reducing redundancy and enhancing the selectivity of feature representation. The low voltage threshold used by C1 neurons allows a single spike to generate a significant response unless inhibited, ensuring that the most prominent features are captured.

    \item {\bf S2 Layer:} The S2 layer performs template matching, where neurons are attuned to specific patterns representing combinations of the basic features detected and pooled in earlier stages. These patterns are pre-learned during a training phase and correspond to various components of the objects being classified. The S2 neurons are structured to implement a max operation, resetting neurons that are sensitive to competing classes whenever a dominant object feature is detected, thereby prioritizing the most significant spike.

    \item {\bf C2 Layer:} The C2 layer pools responses from all S2 locations, effectively discarding precise location data to focus on the existence of detected features. This layer's function is key when multiple objects of interest are present within the same visual scene, as it allows for simultaneous detection without conflating the spatial information of each object.
\end{enumerate}
The classifier following the C2 layer operates based on the number of spikes from the neurons associated with each class, determining the classification outcome by which class has the highest spike count.

\subsubsection{Supervised HSNNs}

Some SNN models focus on the initial stages of feature extraction, stopping at the S1 and C1 layers \cite{zhao2014feedforward, lu2020event, negri2018scene, liu2020unsupervised, liu2020effective}. Following the extraction and pooling operations performed by the S1 and C1 layers, respectively, these models transition directly to a dense (or FC) layer before leading to the output. This architecture allows for a relatively straightforward processing pipeline that can efficiently handle recognizing patterns that do not require the abstraction of highly complex features.

\begin{itemize}
    \item In \cite{zhao2014feedforward}, an architecture for human posture categorization is proposed. The authors use the first two layers of the HMAX model (S1, C1) for feature extraction with inputs from a temporal contrast Address-Event Representation (AER) vision sensor. At the S1 layer, convolutions are performed in parallel by a bank of Gabor filters. At the C1 layer, neurons within the same receptive fields compete in a winner-take-all operation, where the neuron with the maximum response propagates a spike to later layers. Simultaneously, an asynchronous motion symbol detector is employed to ensure that convolutions occur only when motion is detected. The spikes generated by this process are fed into an SNN classifier using the tempotron learning rule. This approach achieved 91.29\% accuracy on the MNIST-DVS dataset and 99.48\% accuracy on the AER Posture dataset.

    \item In \cite{xiao2019event}, a model for object detection is presented using time-surface encoding. The architecture employs the S1 and C1 layers, culminating in a tempotron classifier. This model is notable for its noise-reduction mechanism, which calculates the correlation of an event with its surrounding spatial neighborhood to differentiate between noise and meaningful activity. The method was evaluated on several datasets, including N-MNIST, MNIST-DVS, AER-Posture, and DVS-Poker, achieving accuracies of 93.26\%, 96.66\%, 99.95\%, and 96.17\%, respectively.

    \item In \cite{lu2020event}, a method is proposed for feature extraction using time-surfaces. In the S1 layer, event-driven convolutions are used for the initial extraction of local features directly from AER data. This layer captures the spatiotemporal information inherent in the event streams. In the C1 layer, a temporal winner-take-all mechanism implements a max operation to select the most significant features over a temporal window. This helps highlight the prominent events while suppressing the noise. A tempotron is used for the classification stage. Notable results are that the method achieved 100\% accuracy on the AER posture dataset.

    \item The authors of \cite{negri2018scene} propose an S1-C1 architecture for scene context classification. In their work, they also experiment with adding a hidden layer with varying numbers of neurons, up to a depth of 200 neurons, before the output layers. Their experiments revealed that the network's accuracy improved as the size of the hidden layer (denoted as $M$) increased, achieving an accuracy of 85.4\% with $M=200$ neurons.

    \item In \cite{liu2020effective}, a model is described using a segmented probability-maximization (SPA) learning algorithm within a heterogeneous multi-tasking framework. This network uses a two-layer (S1, C1) feature extraction hierarchy, enabling efficient feature sharing among tasks such as object detection, instance segmentation, and keypoint detection. Classification results are reported for the N-MNIST and Poker-DVS datasets, achieving 96.3\% and 100\% classification accuracy, respectively.
\end{itemize}

Some models take the visual processing a step further by incorporating additional layers - namely, the S2 and C2 layers. This extended hierarchy enables the network to build on the foundational features identified in earlier stages. The S2 layer acts to detect complex patterns, which are essentially combinations or arrangements of the basic features identified by the S1 layer. Following the S2 layer, the C2 layer aggregates the complex patterns to achieve a high level of invariance to positional, scale, and rotational changes.
\begin{itemize}
    \item In \cite{serre2007robust}, the authors propose a four-layer model consisting of S1, C1, S2, and C2 layers. They validate this method on several object classification tasks. At the S1 layer, Gabor filters \cite{gabor1946theory} are used. These filters operate at four orientations and have kernel sizes ranging from 7 $\times$ 7 to 37 $\times$ 37 pixels, increasing in steps of 2 pixels. This results in a pyramid of scales, with a total of 64 different S1 receptive fields (16 scales $\times$ 4 orientations). The C1 layer involves computing local maximum responses for each orientation and scale, effectively pooling the S1 responses to achieve some degree of spatial invariance. In the S2 layer, feature maps are generated using a radial basis function (RBF), which processes the Euclidean distance between the input image patches (formed from the afferent C1 responses) and stored prototype patches. The C2 layer then computes a shift and scale-invariant global maximum response (a winner-take-all mechanism) across all scales and positions of the S2 responses. For the final classification stage, simple linear classifiers, specifically support vector machines (SVM) and boosting, are employed to classify the complex features extracted by the model. The performance of this method is evaluated on several datasets, including CalTech5 (97\% accuracy), CalTech101 (65\% accuracy), MIT-CBCL, and StreetScenes.

    \item The HFirst architecture \cite{orchard2015hfirst} implements non-linear pooling operations that are significantly influenced by the first spike received during computation. This "time to first spike" (TTFS) approach is based on the biological observation that strongly activated neurons tend to fire first. The neuron that responds first is considered to have the maximal response to the input stimulus. At the pooling stages (C1 and C2), neuron firing follows a temporal winner-take-all operation, where neurons that fire first inhibit other neurons through lateral reset connections. Instead of pooling multiple feature maps of orientations and scales for maximum response, the first spike response to appear is judged to be the maximal response. A lateral reset mechanism ensures that competing neurons that lost the TTFS operation do not fire. The responses of C2 cells are then classified by counting the number of spikes from neurons corresponding to each class and selecting the class with the highest spike count. The HFirst model achieved an impressive 97.5\% accuracy on the Poker-DVS classification dataset.
\end{itemize}

\subsubsection{Unsupervised HSNNs}

Unsupervised learning in HSNNs leverages mechanisms like STDP to enable networks to autonomously develop representations of input data without explicit labels. STDP adjusts synaptic strengths based on the precise timing of spikes, closely aligning with the Hebbian learning principle that neurons that fire together, wire together \cite{hebb2005organization}. In these architectures, excitatory and inhibitory neurons collaborate to refine synaptic adjustments, ensuring that learning focuses on discriminative patterns while maintaining network stability \cite{maass2000computational}.

Diehl and Cook \cite{diehl2015unsupervised} introduced a foundational learning rule where each presynaptic spike increments a synaptic trace that decays exponentially over time. This mechanism allows synaptic weight adjustments to be proportional to the presynaptic neuron's firing rate, promoting efficient unsupervised learning. The network architecture typically consists of an input layer connected to an excitatory layer, which is paired one-to-one with an inhibitory layer to establish lateral inhibition and a competitive environment among neurons.

Building upon these principles, several studies have developed effective unsupervised HSNNs:

\begin{itemize}
    \item In the paper \cite{liu2020unsupervised}, the authors present a novel approach for unsupervised object recognition using a two-layer NN, consisting of layers S1 and C1, to extract features. An unsupervised STDP learning rule is employed for classification. The features extracted are converted into spikes using a natural logarithmic temporal coding function. These feature spikes, from multiple scales but having the same position and orientation, are fused into a single spike train. This fusion forms a multiscale spatio-temporal feature representation, which is then classified using the STDP rule. The classification performance of this method was evaluated on several datasets: MNIST-DVS, AER Posture, and Poker-DVS, achieving accuracy results of 89.96\%, 99.58\%, and 99\%, respectively.
    
    \item Similarly, \cite{zhou2022bio} proposes a multi-stage object recognition system combining noise filtering and event stream segmentation. Noise filtering is accomplished with LIF neurons, which map densely (1-to-1) onto the sensor pixel resolution and ensure that only recurring pixel activations will stimulate input neurons of the system. The correlation between events at a pixel location and events in surrounding spatial neighborhoods is also calculated here to ensure that sparsely distributed noise events are filtered out. The event stream segmentation method ensures that only meaningful motion events trigger feature extraction and classification so that computational and energy efficiency are respected. Additionally, the event stream segmentation is responsible for chunking the full-length event stream into smaller time slices or numbers of events. This is achieved with a LIF neuron to determine if sufficient event activity is present in the stream to warrant feature extraction. The four-layer (S1, C1, S2, C2) network achieves classification with the R-STDP learning rule \cite{fremaux2016neuromodulated} for feature extraction in the S2 layer and predictive decision assignment in the C2 layer. Classification results are reported for MNIST-DVS, AER Posture, Poker-DVS, and N-CARS datasets, with reported accuracies of 94.99\%, 100\%, 100\%, and 85.16\%, respectively.

    \item In \cite{kheradpisheh2016bio}, a bio-inspired hierarchical neural network architecture is presented using unsupervised learning through STDP. The model implements a competitive learning mechanism and winner-take-all strategy to develop sparse, efficient coding of visual features. The effectiveness of the model is validated on the 3D-Object and ETH-80 datasets, which test invariant recognition capabilities under varying object positions, sizes, and lighting conditions. The model achieved accuracies of 96\% and 81\%, respectively. The authors demonstrate that their model outperforms both traditional HMAX and modern deep-learning-based methods in handling large variations in object appearance.
\end{itemize}

\begin{table*}[t]
    \centering
    \caption{Summary of Spiking Convolutional Neural Networks (SCNNs)}
    \begin{tabular}{|p{1.7cm}|p{3cm}|p{10cm}|}
        \hline
        \textbf{Type} & \textbf{Method} & \textbf{Description} \\
        \hline
        \multirow{3}{*}{Unsupervised} & Kheradpisheh et al. \cite{kheradpisheh2018stdp} & Deep SCNN with STDP, initial DoG filtering, convolutional and pooling layers, classified by SVM \\
        \cline{2-3}
        & Panda et al. \cite{panda2016unsupervised} & Auto-Encoder learning model, unsupervised layer-wise training, partial supervision for output layer \\
        \cline{2-3}
        & Tavanaei et al. \cite{tavanaei2018training} & Hybrid STDP and gradient descent training, STDP for convolutional filters, supervised for classification \\
        \hline
        \multirow{10}{*}{Supervised} & Lee et al. \cite{lee2016training} & Modified backpropagation for SCNNs, dynamic weight adjustment, direct gradient descent on membrane potentials \\
        \cline{2-3}
        & Cordone et al. \cite{cordone2021learning} & Sparse convolutional layers, strided convolutions, surrogate gradient approach, LIF neurons \\
        \cline{2-3}
        & Jin et al. \cite{jin2018hybrid} & Hybrid Macro/Micro Level Backpropagation, macroscopic and microscopic error rate focus \\
        \cline{2-3}
        & Esser et al. \cite{esser2015backpropagation} & Probabilistic spikes and synapses, IBM TrueNorth deployment \\ 
        \cline{2-3}
        & Meng et al. \cite{meng2022training} & Differentiation on Spike Representation (DSR) method, gradient-based optimization without BPTT \\
        \cline{2-3}
        & Xiao et al. \cite{xiao2022online} & Online Training Through Time (OTTT), forward-in-time learning, presynaptic inputs weighted by synaptic strengths \\
        \cline{2-3}
        & Yao et al. \cite{yao2022attention} & Multi-dimensional Attention (temporal, channel, spatial), enhance focus on specific data aspects \\
        \cline{2-3}
        & Guo et al. \cite{guo2022real} & Real Spike framework, real-valued spikes during training, binary spikes during inference \\
        \cline{2-3}
        & Fang et al. \cite{fang2021deep} & Spike-Element-Wise (SEW) ResNet, element-wise identity mapping, mitigate gradient issues \\
        \cline{2-3}
        & Samadzadeh et al. \cite{samadzadeh2023convolutional} & Hybrid training with ANN-to-SNN conversion, stable gradients in initial phase, BPTT in final phase \\
        \hline
        \multirow{8}{*}{Converted} & Perez et al. \cite{perez2013mapping} & Mapping frame-driven CNNs to event-driven architectures, fine-tuning for spiking domain \\
        \cline{2-3}
        & Hunsberger et al. \cite{hunsberger2016training} & Integration of LIF neurons, softening response function, adjustments for spiking implementations \\
        \cline{2-3}
        & Rueckauer et al. \cite{rueckauer2016theory} & SNNs approximating ReLU activations, max-pooling, softmax, batch normalization in SNNs \\
        \cline{2-3}
        & Rueckauer et al. \cite{rueckauer2017conversion} & Max Normalization algorithm, reset-by-subtraction neuron, SNN-conversion toolbox \\
        \cline{2-3}
        & Diehl et al. \cite{diehl2015fast} & Weight and threshold balancing techniques, model-based and data-based normalization \\
        \cline{2-3}
        & Ding et al. \cite{ding2021optimal} & ANN-to-SNN conversion, Rate Norm Layer, optimal fit curve for activation values \\
        \cline{2-3}
        & Cao et al. \cite{cao2015spiking} & Conversion methodology for SNNs, architecture adjustments, power-efficient implementations \\
        \cline{2-3}
        & Massa et al. \cite{massa2020efficient} & Gesture recognition on Intel Loihi, CNN-to-SNN conversion, normalization and partitioning \\
        \hline
    \end{tabular}
    \label{tab:scnns}
\end{table*}

\subsubsection{Discussion}
A primary advantage of Hierarchical Spiking Neural Network (HSNN) architectures, particularly evident in models like the Spiking HMAX, is their ability to process visual information hierarchically—from simple features to complex object representations. This method closely mirrors visual processing in the mammalian cortex, where neurons in early visual areas respond to basic features, and neurons in higher areas respond to more complex patterns. The use of predefined Gabor filters in the S1 layer of HSNNs, analogous to simple cells in the primary visual cortex, allows for efficient feature detection without the need to train this part of the network. These fixed filters rapidly identify fundamental visual elements like edges and bars across various orientations.

However, reliance on fixed weights introduces limitations in adaptability and generalization. In contrast, Spiking Convolutional Neural Networks (SCNNs) and other learning-based SNNs utilize learnable weights that can be adjusted through training to optimize feature extraction for specific tasks and datasets. This learning capability allows SCNNs to develop customized filters that capture more nuanced or task-specific features than Gabor filters. Moreover, learnable weights in SCNNs enable the network to adapt to new, unseen data after the initial training phase, a flexibility that fixed-weight HSNNs lack \cite{xiao2022online, dhoble2012online}.

\subsection{SCNNs}
CNNs are particularly influential in domains requiring the analysis of visual data. Inspired by the neuron responses in the primary visual cortex (V1), the early layers of CNNs are designed to mimic the detection of primary visual features, such as oriented edges, which are akin to the responses evoked by V1 neurons in the brain to visual stimuli \cite{cadena2019deep}. These convolutional layers use trainable kernels or filters to extract simple features from input images, primarily edges and basic textures, while higher layers progress to more abstract features like shapes \cite{kolisnik2021condition}. CNNs typically comprise a sequence of convolution and pooling (subsampling) layers, followed by a feedforward classifier. This architecture, exemplified by early models like the LeNet \cite{lecun1998gradient}, has demonstrated exceptional effectiveness across a range of applications, including image recognition and object detection or segmentation. The first convolutional layer in these networks uses filters that perform similarly to Gabor filters, which are particularly effective at responding to various orientations and spatial frequencies in images. Subsequent layers delve into more complex feature extraction, and pooling layers serve to reduce spatial dimensions through operations like max and average pooling over defined neighborhoods, thus achieving invariance to positional shifts, scales, and rotations. In the context of SCNNs, which integrate the temporal dynamics of spiking neurons, the challenge lies in adapting the well-established CNN architectures to leverage these dynamics effectively. SCNNs have been trained using both supervised and unsupervised learning rules. Table \ref{tab:scnns} provides a summary of various Spiking Convolutional Neural Networks (SCNNs).

\subsubsection{Unsupervised/Hybrid SCNNs}
\begin{itemize}
    \item In \cite{kheradpisheh2018stdp}, a deep SCNN is described, using STDP for unsupervised learning. The architecture begins with an initial layer that uses Difference-of-Gaussian (DoG) filtering to mimic the retinal processing of contrast detection. This is followed by convolutional and pooling layers, which extract visual features subsequently classified by a SVM. The network's performance is evaluated on several benchmark datasets, achieving accuracies of 99.1\% on Caltech 101, 82.8\% on ETH-80, and 98.4\% on MNIST.

    \item In \cite{panda2016unsupervised}, an unsupervised learning framework for training Deep SCNNs is proposed. The method is based on the Auto-Encoder learning model, enabling each layer of the SCNN to be trained unsupervised through an encoder-decoder scheme that uses spike-based timing. The network's output layer is trained with partial supervision, using a subset of labeled data to fine-tune the network for classification tasks. This hybrid training approach, which combines unsupervised feature extraction with supervised fine-tuning, yielded classification accuracies of 95\% on the MNIST dataset and 24.58\% on the CIFAR-10 dataset.

    \item In \cite{tavanaei2018training}, a hybrid method for training SCNNs using a combination of STDP and gradient descent is introduced. The training process begins with STDP to train convolutional filters, followed by a supervised learning rule that applies an STDP-based method approximating gradient descent for classifying spike patterns. The proposed model achieved a classification accuracy of 98.6\% on the MNIST dataset.
\end{itemize}

\subsubsection{Supervised SCNNs}
Non-spiking CNNs are universally trained using variants of the backpropagation algorithm. Recently, backpropagation has also been used to train SCNNs \cite{lee2016training, cordone2021learning, jin2018hybrid, yao2022attention, meng2022training, guo2022attention, fang2021deep, xiao2022online}. Training SCNNs presents unique challenges due to the spatial complexities of convolutional processing and the temporal precision of spike timings. The works discussed here explore innovative training and learning strategies enabling the adaptation of deep learning techniques to the spiking domain.

\begin{itemize}
    \item In \cite{lee2016training}, a modified backpropagation algorithm is introduced for training SCNNs, using dynamic weight adjustment and optimized initialization to manage refractory periods and mitigate gradient instability. The network applies direct gradient descent on membrane potentials, with weight distributions controlled by the inverse square root of synaptic connections to balance neuron activation across layers. The method achieved up to 99.31\% accuracy on the N-MNIST dataset.

    \item In \cite{cordone2021learning}, the authors propose enhancements to the supervised backpropagation-based learning algorithm for SNNs, incorporating strided sparse convolutions. The model uses sparse convolutional layers, which only process non-zero data to maintain data sparsity across the network. To reduce the spatial dimensions of feature maps through the network, strided convolutions with a stride of 2 $\times$ 2 are used on the spatial dimensions. These are used instead of pooling layers due to increased compatibility with neuromorphic hardware while maintaining sparsity. The output feature maps of the convolutional layers are flattened along the feature and spatial dimensions and fed to a fully-connected layer that outputs predictions for each timestep. The authors use a surrogate gradient approach to train their models with LIF neurons. During the forward pass, spikes are generated according to the LIF firing rules, and during the backward pass, the gradient is approximated by the gradient of a sigmoid function. Benchmarking on the event-captured IBM DVS128 Gesture Dataset, the authors compare the test accuracy and training efficiency (epochs and training time) of their sparse SNN model favorably against both dense and sparse CNNs and SNNs.

    \item In \cite{jin2018hybrid}, a Hybrid Macro/Micro Level Backpropagation (HM2-BP) approach is proposed with two distinct levels of computation defined: macroscopic and microscopic levels. FCSNN and SCNN architectures are trained for evaluation of multiple benchmarking datasets. The SCNN consists of two 5$\times$5 convolutional layers, each followed by a 2$\times$2 pooling layer and a fully connected hidden layer. The FCSNN consisted of a single hidden layer (800 neurons for N-MNIST). The number of input neurons is equal to the number of pixels in the input images (784 for N-MNIST). The number of neurons in the output layer is equal to the number of classes for prediction (10 for N-MNIST). The primary (macro) focus is on the overall firing rate of neurons. The error rate at this level is defined based on the difference between the actual and desired firing rate of neurons. In addition, the Spike-Train Level Post-Synaptic Potential (S-PSP) is calculated, which focuses on the precise timing of individual spikes (the authors describe this as a micro-level focus). The S-PSP is a measure of the influence a pre-synaptic neuron's spikes has on a post-synaptic neuron. The derivative of the post-synaptic potential with respect to the synaptic weight is calculated for each pair of pre- and post-synaptic spike trains. The synaptic weight updates are performed during the backpropagation phase, which relies on gradients computed by integrating both micro and macro-level computations. The HM2-BP algorithm was evaluated on both the static MNIST and dynamic N-MNIST datasets, and accuracy scores of 99.49\% and 98.88\% were reached, respectively. The method was additionally tested on the Extended MNIST (EMNIST) dataset to score an accuracy of 85.57\%.

    \item In \cite{meng2022training}, limitations of existing surrogate gradient (SG) learning methods in SNNs are explored. The authors introduce the Differentiation on Spike Representation (DSR) method to address the inherently non-differentiable dynamics of SNNs by encoding spike trains into a continuous-valued domain. This representation transforms the spiking dynamics into a sub-differentiable mapping, enabling the use of gradient-based optimization techniques and circumventing the non-differentiability issue in SNNs. The DSR approach applies backpropagation directly on the spike representation without requiring Backpropagation Through Time (BPTT), reducing both computational complexity and memory requirements. The DSR method was shown to outperform existing ANN-to-SNN conversion and direct training methods for pre-activation ResNet-18 and VGG-11 architectures. This work was benchmarked on CIFAR-10, CIFAR-100, ImageNet, and DVS-CIFAR10, to achieve accuracy rates of 95.24\%, 78.50\%, 67.74\%, and 77.27\%, respectively.

    \item In \cite{xiao2022online}, a training approach for SNNs called Online Training Through Time (OTTT) is presented. OTTT is derived from BPTT but modified to enable forward-in-time learning without requiring backpropagation across time steps. OTTT enables forward-in-time learning by computing gradients instantaneously at each time step. This is distinct from traditional BPTT, which requires storing all temporal computational states, which is expensive in terms of memory usage. A running sum of pre-synaptic inputs weighted by their respective synaptic strengths and decay factors is maintained. These tracked activities are then used to compute gradients for weight updates at each time step. The weight update rule in OTTT resembles the three-factor Hebbian learning rule observed in biological neural systems. This model incorporates pre-synaptic activity, a post-synaptic surrogate derivative (mimicking post-synaptic potential), and a global error term. For practical implementation, the authors discuss removing batch normalization in favor of scaled weight standardization, which standardizes weights based on their mean and variance, calculated independent of batch-wise statistics. Experimental results are demonstrated for online training on CIFAR-10, CIFAR-100, ImageNet, and CIFAR10-DVS, achieving 93.49\%, 71.05\%, 64.16\%, and 76.63\%, respectively.
\end{itemize}

\noindent Further enhancements in the performance and efficiency of SCNNs have led to notable improvements in both accuracy and responsiveness. The advancements highlighted in the following discussions encompass significant innovations, such as the implementation of attention mechanisms and methods for fine-tuning network focus for efficiency and efficacy gains.

\begin{itemize}
    \item In \cite{yao2022attention}, the use of attention mechanisms is presented for SCNNs. Introduced as Multi-dimensional Attention (MA), the core idea is learning to attend to specific aspects of data along three dimensions - temporal, channel, and spatial.
    \begin{enumerate}
        \item Temporal Attention (TA): This dimension focuses on the "when" aspect of the data. It determines when the network should pay more attention, allowing it to emphasize important temporal features while de-emphasizing irrelevant ones. This could mean, for example, focusing more on moments of a video sequence where significant movements occur. This is done by first applying average and max pooling operations across the temporal dimension to condense the information, followed by MLPs to generate a weight for each time step. These weights are used to scale the membrane potentials of relevant neurons.

        \item Channel Attention (CA): This dimension targets the "what" aspect by focusing on specific channels of the network's feature maps. Channels of a feature map in a NN typically correspond to a particular kind of feature extracted from the input data. Through channel attention, the network can amplify channels that are more relevant to the task at hand while suppressing less relevant ones. This is done by applying average and max pooling across the spatial dimensions to aggregate spatial information within each channel. Then, a MLP is used to assign a significant weight to neurons in each channel.

        \item Spatial Attention (SA): This dimension addresses the "where" aspect, enabling the network to focus on specific areas within the input space that are more informative for making a decision. For instance, in image processing, spatial attention might allow the network to concentrate on the part of the image where the object of interest is located rather than attending to the entire image uniformly \cite{guo2022attention}. The implementation uses average and max pooling applied across the channel dimension to create a spatial context description, followed by a convolution operation with a relatively larger filter (e.g., 7 $\times$ 7) applied to this descriptor to create a 2D attention map. The attention map is then element-wise multiplied with the original input feature map to enhance important regions and suppress less relevant ones.
    \end{enumerate}
    The authors note that these three attention mechanisms can be used individually or in combination, depending on specific task requirements. For instance, in a task where temporal dynamics are crucial (like video processing), TA can play a significant role. In contrast, for tasks like image classification, CA and SA are more relevant. On the ImageNet benchmark, the CSA-Res-SNN-104 model achieved accuracies of 75.92\% for a single time step and 77.08\% for four time steps, surpassing the state-of-the-art results in the SNN domain. This model also demonstrated up to 31.8x better compute energy efficiency when compared to its ANN counterpart. For the DVS128 Gesture dataset, MA applied to SNN also led to notable improvements in accuracy. The implementation of the attention mechanism in the three-layer SNN model, following the structure described in \cite{yao2021temporal}, resulted in a 6.3\% accuracy increase over baseline. MA applied to a five-layer SNN model, based on the structure in \cite{fang2021deep} led to an accuracy improvement of 2.73\%.

    \item In \cite{guo2022real}, the Real Spike training and inference framework is introduced. Real Spike is a training and inference decoupling method for SNNs that uses real-valued spikes during training and binary spikes during inference. This approach uses a novel re-parameterization technique to transform the shared convolution kernels used in training to multiple unshared kernels during inference to increase the representation capacity of the network. During training, SNNs learn with real-valued spikes and shared convolutional kernels for each output feature map. During inference, the real-valued spikes are converted to binary format. This method is compatible with both neuromorphic and GPU-based hardware platforms. The authors conducted ablation studies and performance comparisons across the CIFAR-10, CIFAR-100, ImageNet, and CIFAR10-DVS datasets using ResNet and VGG architectures. Results of note are that the ResNet19 architecture achieved 95.71\% accuracy on CIFAR-10, Resnet20 achieved 67.69\% on CIFAR-100 and 78\% on CIFAR10-DVS, and ResNet34 achieved 67.69\% on ImageNet.
\end{itemize}

\noindent Residual learning, proposed in ResNet architectures \cite{he2016deep, he2016identity}, enables the learning of identity functions through added skip connections. Instead of learning outright features or transformations, layers in the ResNet learn residual functions relating to the layer inputs. Residual architectures have been similarly explored in spiking contexts \cite{fang2021deep, samadzadeh2023convolutional}.

\begin{itemize}
    \item In \cite{fang2021deep}, the Spike-Element-Wise (SEW) ResNet is introduced, a novel architecture that aims to address key challenges in training deep SNNs, particularly issues related to the degradation problem and the vanishing/exploding gradient problems. Identity mappings are a challenge for traditional Spiking ResNets due to the non-differentiable nature of spikes. To address this challenge, SEW ResNet introduces a block structure that combines spiking neurons with an element-wise operation. The Element-wise identity mapping takes pairs of spike tensors as input and outputs a new spike tensor. The authors demonstrate that SEW ResNet can mitigate the vanishing and exploding gradient problems by ensuring that the residuals in the blocks do not interfere with the gradient flow. This is achieved through the use of element-wise operations (ADD, AND, and inverse AND), which determine how the input and residual spikes are combined. ResNets with depths of 34, 50, 101, and 152 have been trained using spike-based backpropagation. Experimental results demonstrated SEW ResNet architectures achieved 69.26\% on ImageNet, 97.92\% on DVS-Gesture, and 74.4\% on CIFAR10-DVS.

    \item In \cite{samadzadeh2023convolutional}, a training strategy that combines the advantages of both ANN-to-SNN conversion and BPTT is proposed. This hybrid method starts with a phase of training using non-spiking activations (Leaky-ReLU) for each layer, except the output. This phase allows the network to use conventional backpropagation for weight updates. After a set number of epochs with non-spiking activations, the training transitions to spiking thresholds. Here, the LIF neuron model is used to convert the network to SNN format. During this phase, training is done using BPTT. The initial non-spiking phase is key to providing stable gradients that help address gradient vanishing. The STS-ResNet architecture developed in this work incorporates several design elements from ResNet that have been adapted for spiking dynamics. The network comprises 18 layers, including convolutional FC, thresholding, and dropout layers. Skip connections are used in each block, similar to ResNet. The method achieved 99.6\% on NMIST, 69.2\% on DVS-CIFAR10, and 96.7\% on DVS-Gesture.
\end{itemize}

\subsubsection{Converted SCNNs}
A primary method for using spiking architectures without the extensive training process of SCNNs involves converting a pre-trained CNN to a spike-driven counterpart. This process includes adjusting neuron models (like integrating LIF mechanisms), balancing synaptic weights and thresholds to preserve network behavior post-conversion, and implementing normalization strategies to ensure robust performance across input conditions.

Several works have proposed core methodologies for converting traditional CNNs to SNNs, focusing on techniques that adapt pre-synaptic weights and network architectures to accommodate the discrete and dynamic nature of spike-based processing \cite{perez2013mapping, hunsberger2015spiking, rueckauer2016theory}. These conversion techniques are key for enabling the seamless deployment of deep learning models using existing neural network architectures to reap the benefits of energy-efficient, spike-based processing.

\begin{itemize}
    \item In \cite{perez2013mapping}, a mapping for traditionally trained frame-driven CNNs to event-driven architectures is presented. Their significant contribution lies in demonstrating that pre-trained CNNs can be effectively adapted to process event-based data from DVS cameras, preserving high accuracy while benefiting from the temporal resolution and energy efficiency of event-driven processing. Using aggregated events from a DVS camera to create pseudo-frames, a traditional CNN is trained using standard backpropagation techniques. Then, the CNN is adapted for the spiking domain through weight mapping and architecture adjustments. The mapped SCNN is then fine-tuned, which involves the tuning of thresholds, refractory periods, and synaptic weights. The method was tested on a human silhouette recognition task and the DVS-Poker dataset, with top accuracies of 95.6\% and 92.6\%, respectively.

    \item In \cite{hunsberger2016training}, the integration of LIF neurons into deep SNNs is demonstrated. The authors mitigate the non-differentiability of LIF neurons by softening their response function to enable gradient-based optimization. This work is significant because it bridges the gap between traditional deep learning and SNNs, enabling the training of deep SNNs with backpropagation and achieving competitive performance on complex datasets. Modifications to the standard convolutional architectures include the removal of local response normalization and substitution of max pooling with average pooling to suit spiking implementations. The model achieved competitive accuracy results of 99.22\% on MNIST, 83.54\% on CIFAR-10, 55.13\% on CIFAR-100, and 76.20\% on ImageNet datasets.

    \item In \cite{rueckauer2016theory}, a theoretical explanation of how SNNs approximate the ReLU activations in traditional CNNs is provided. The significance of this work lies in establishing a theoretical foundation for ANN-to-SNN conversion, enabling near-lossless conversion by ensuring that firing rates in SNNs accurately represent activation levels in ANNs. This is critical because it justifies the transformation of analog signals into spiking outputs, where the firing rates of neurons in the SNN correspond to the activation levels in the CNN. Common CNN features like max-pooling, softmax, and batch normalization are implemented within the SNN framework. A modified reset mechanism is introduced that adjusts the membrane potential of neurons after a spike to allow control over neuron firing rates towards better approximations of CNN activations. The authors discuss the use of the 99th percentile of sampled activation level for weight normalization instead of the maximum activation level to address the issue of outlier activations that can disrupt the learning and generalization capabilities of the SNN. The conversion methods were tested on MNIST and CIFAR-10, achieving 99.44\% and 87.62\%, respectively.

    \item In \cite{lu2020exploring}, the authors demonstrate that training SNNs with extreme quantization can achieve near full-precision accuracy on complex datasets such as ImageNet and CIFAR-100. The research proposed the B-SNN (SNN with binary weights) to demonstrate that quantization, especially to binary levels, does not lead to a marked degradation in neural network performance. To support this goal, the authors adapted non-spiking networks through a conversion and pruning technique focusing on a modified VGG-15 structure, which involved reducing the number of linear layers in the network. During conversion, the full-precision weights of the original neural network are quantized into binary values. The analog ReLU neurons are replaced with IF neurons as part of the conversion. The pruning process removes the bias and batch-normalization layers. The results on the CIFAR100 dataset demonstrated a 4.46\% loss in accuracy for a 160.06\% increase in efficiency, measured as the number of operations in the convolutional and linear layers. This study is significant because it shows that SNNs can maintain high accuracy even with binary weights, leading to substantial reductions in computational complexity and energy consumption. 

    \item In \cite{singh2021gesture}, the authors propose the Gesture-SNN model and investigate the accuracy degradation when applying ANN-to-SNN conversion techniques. Their significant contribution is in demonstrating that carefully designed conversion methods can preserve accuracy while benefiting from the energy efficiency of SNNs, particularly for gesture recognition tasks on neuromorphic hardware. Three ANN models with varying topologies are trained and converted using the proposed method. Topologies 1, 2, and 3 contain 5, 6, and 7 stacked convolutional layers, respectively. Each topology also used two average pooling layers and a single fully connected layer. The three CNNs were trained with constraints such as no bias term, average pooling layers for dimensionality reduction, and dropout for regularization. After training, ReLU neurons were replaced by LIF neurons. Neuron firing thresholds were set based on the activation percentiles from the corresponding ANN. Benchmarking was performed using the event-captured IBM DVS-Gesture dataset.
\end{itemize}

\noindent Several authors have reported on techniques for normalization and optimization of synaptic weights and neuron thresholds to manage the network's response to input spikes, ensuring that the SNN mimics the analog behavior of the original CNN \cite{diehl2015fast, ding2021optimal}. These normalization strategies are key to maintaining high accuracy and functionality of SNNs post-conversion, often achieving near-lossless performance compared to the original ANNs.

\begin{itemize}
    \item In \cite{rueckauer2017conversion}, the authors propose the Max Normalization algorithm, a spiking implementation of a max-pooling layer. They propose a reset-by-subtraction neuron to reduce performance loss through the conversion process. For this task, the authors have open-sourced an SNN-conversion toolbox (SNN-TB) capable of transforming models written in Keras, Tensorflow, or Pytorch. This software kit also includes simulation tools for the evaluation of the resultant spiking model and deployment interfaces for SpiNNaker or Loihi. This work is significant as it provides practical tools and methods for converting and simulating SNNs, facilitating wider adoption and experimentation in the community. The results of this work were evaluated on the MNIST and CIFAR-10 datasets and achieved 99.41\% and 88.82\% accuracy, respectively.

    \item In \cite{diehl2015fast}, CNNs and FC-FF-NNs are converted using weight and threshold balancing techniques to preserve accuracy. The significance of this study lies in its demonstration of near-lossless conversion, where the converted SNNs maintain accuracies comparable to their ANN counterparts while benefiting from the efficiency of spike-based computation. The networks are structured with layers of IF neurons without biases, using ReLUs during training. The conversion process uses two novel techniques:
    \begin{itemize}
        \item Model-based Normalization: The maximum positive activation that could occur as input to a layer is noted as normalization factor $\lambda^{l}$. All synaptic weights preceding a neural layer are scaled by a $\lambda^{l}$, and the threshold is set to 1. Alternatively, the threshold is set equal to $\lambda^{l}$ with the synaptic weights unchanged. This configuration is chosen such that the network will never produce more than one spike at once from a given neuron when maximum positive input can cause a single spike. Thus, the resultant network is robust to arbitrarily high input rates and losses due to over-saturated inputs are eliminated. These methods are mathematically equivalent.

        \item Data-based Normalization: Rather than assuming the worst-case scenario of maximum possible activation, the realistic maximum activation within the training set is used to determine the membrane potential to generate a single spike. The maximum ReLU activation in layer $l$ is noted as normalization factor $\lambda^{l}$. All synaptic weights preceding a neural layer are scaled by a $\lambda^{l}$, and the threshold is set to 1. Alternatively, the threshold is set equal to $\lambda^{l}$ with the synaptic weights unchanged. These methods are mathematically equivalent.
    \end{itemize}
    The efficiency of the converted SNNs is demonstrated on the MNIST dataset, where it achieved accuracies of 99.12\% with data-normalization and 99.10\% with model-normalization.

    \item In \cite{ding2021optimal}, the authors analyze the ANN-to-SNN conversion process and derive conditions for optimal conversion. They argue that traditional methods, which focus on direct weight and threshold translation, result in performance losses. To this end, the ReLU activation layer found in the source ANN is replaced with a Rate Norm Layer (RNL). The RNL uses a clip function with a trainable upper bound to simulate the firing rate, which can be adapted during training. Additionally, they propose an optimal fit curve to measure the correlation between activation values of the ANN and firing rates of the SNN. They claim that inference time can be significantly reduced by optimizing the upper bound of this curve. Modern deep neural architectures such as VGG-16 and PreActResNet are converted through this work. The proposed method is validated on MNIST, CIFAR-10, and CIFAR-100 datasets and demonstrates near loss-less conversions, with best accuracies of 99.46\%, 93.54\%, and 75.02\%, respectively.
\end{itemize}

\noindent Several research initiatives illustrate the successful implementation of SCNNs on neuromorphic hardware \cite{esser2015backpropagation, cao2015spiking}. The implementation of SCNNs on neuromorphic hardware enables systems to perform complex visual processing tasks with a fraction of the power required by traditional hardware.
\begin{itemize}
    \item In \cite{esser2015backpropagation}, a probabilistic representation of spikes and synapses is proposed to allow the use of backpropagation for training. The training scheme is such that both neuronal activations and synaptic states are represented as probabilities within the range [0,1], aligning the training network with the operational characteristics of the deployment hardware. The network is deployed on the IBM TrueNorth chip. This work demonstrates favorable performance and energy efficiency on the MNIST dataset, with the trained SNN achieving 99.42\% accuracy at 108 $\mu$J per image using an ensemble of 64 networks and 92.7\% accuracy at 0.268 $\mu$J per image in a single-network setup.

    \item In \cite{cao2015spiking}, a conversion methodology is given with three main steps outlined: 1) adjusting the architecture of a conventional CNN to suit the requirements and limitations of an SNN, 2) training this tailored CNN using standard backpropagation, and 3) mapping the trained CNN onto an SNN. The architecture adjustment includes changing CNN activation functions from hyperbolic tangent (tanh) to rectified linear units (ReLU) and removing negative biases and outputs to adapt to the non-negative nature of spikes. The SNNs are evaluated on two datasets: DARPA's Neovision2 Tower dataset and CIFAR-10 dataset, achieving 99\% and 77.43\%. The authors report a power consumption of 3.7e-7 $\mu$J per spike when implemented on FPGA.

    \item In \cite{massa2020efficient}, a gesture recognition model is implemented for use on the Intel Loihi neuromorphic processor. A traditional CNN is first trained with the following properties: the input layer connects to four convolutional layers, each followed by ReLU activation functions. The convolutional layers vary in the number of filters, kernel sizes, and strides. The convolutional output is flattened and fed to a dense layer. The dense layer uses the flattened vector with SoftMax activations to predict the probabilities of classes recognized by the network. After training of the CNN, the model is converted into an SNN. The authors detail four key steps in the conversion process.
    \begin{itemize}
        \item {\bf Parsing} involves extracting relevant info from the CNN model, discarding layers that are not used during inference (e.g., Dropout, BatchNormalization layers), and converting MaxPooling2D layers into AveragePooling2D, which are supported by Loihi.

        \item {\bf Conversion} wherein CNN-trained weights and biases are normalized to the limited dynamic range of the spiking neurons. The normalization process is guided by maintaining a constant ratio between incoming inputs and their membrane threshold, known as the Desired Threshold to Input Ratio (DThIR).

        \item {\bf Partition} finds a valid partitioning of the converted SNN model onto the physical architecture of the Loihi chip. This step requires consideration of Loihi's constraints, such as a maximum number of neurons and synapses per neurocore, as well as fan-in and fan-out capabilities.

        \item {\bf Deployment} maps the model onto the Loihi chip. The model is now ready for SNN deployment and performing inference.
    \end{itemize}
    This work is implemented on the Intel Loihi neuromorphic processor. Three datasets are tested in this work: MNIST, CIFAR10, and DVS-Gesture. The converted SNN achieves 98.5\% accuracy on MNIST, 77.1\% on CIFAR10, and 89.64\% on DVS-Gesture.
\end{itemize}

\begin{table*}[t]
\centering
\caption{Summary of SDBN Papers}
\begin{tabular}{|p{2.5cm}|p{4cm}|p{4cm}|p{3cm}|}
\hline
\textbf{Paper} & \textbf{Key Contribution} & \textbf{Model Details} & \textbf{Training Method} \\ \hline
Tavanaei et al. \cite{tavanaei2019deep} & Introduced sRBMs for SDBNs & sRBMs with spiking neurons & CD \\ \hline
Neftci et al. \cite{neftci2014event} & Proposed foundational sRBM using IF neurons & sRBMs performing MCMC sampling & Event-driven CD with STDP \\ \hline
Lee et al. \cite{lee2007sparse} & Proposed sparse DBNs for visual cortex modeling & Sparse DBN with hierarchical layers & CD \\ \hline
O'Connor et al. \cite{o2013real} & First SDBN by converting traditional DBN to spiking neurons & Traditional DBN with LIF neurons & Conventional training and Siegert approximation for conversion \\ \hline
Stromatias et al. \cite{stromatias2015robustness} & Explored DBN conversion on neuromorphic hardware & Converted DBN with low precision weights & Dual-copy rounding \\ \hline
Stromatias et al. \cite{stromatias2015scalable} & Mapped SDBN to SpiNNaker platform & SDBN with Poisson spike train input & CD for offline training \\ \hline
Merolla et al. \cite{merolla2011digital} & Implemented S-RBM on neurosynaptic core & S-RBM with binary weights & CD for offline training \\ \hline
Neil et al. \cite{neil2014minitaur} & Discussed Minotaur system for S-RBMs and DBNs on FPGA & S-RBM/DBN with LIF neurons post-training & Weight normalization for spiking dynamics \\ \hline
Lee et al. \cite{lee2011unsupervised} & Extended sparse DBN to convolutional sparse DBN for visual tasks & Convolutional approach to DBN & Unsupervised pre-training and supervised fine-tuning \\ \hline
Kaiser et al. \cite{kaiser2017spiking} & Introduced Spiking Convolutional DBN (SCDBN) & SCRBM with convolutions and lateral inhibitions & CD with STDP \\ \hline
\end{tabular}
\label{table:sdbns}
\end{table*}

\subsubsection{Discussion}
Traditionally, CNNs have dominated tasks involving static image processing by learning hierarchical representations. However, the adaptation of these networks to their spiking counterparts introduces the ability to handle dynamic inputs, leveraging temporal information, which is a significant step forward. This transition not only mirrors the biological processes of the human visual cortex more closely but also enhances the ability to interact with real-world environments in a manner traditional CNNs cannot. For instance, the adaptation of CNNs to SCNNs, especially in applications like video processing or real-time event detection, offers a more granular understanding by processing frame-by-frame changes as continuous data streams. The benchmarking of SCNNs on standard neuromorphic datasets like N-MNIST variants and DVS-Gestures highlights their growing capability to achieve parity with traditional CNNs. However, the reported performances also reflect the sensitivity of SCNNs to network configurations and learning rules. The efficiency of SCNNs in applications such as gesture recognition on neuromorphic hardware demonstrates their potential, albeit with an acknowledgment of the need for more refined training techniques and hardware-specific optimizations.

Despite their advantages, the implementation of SCNNs introduces several challenges, predominantly related to training these networks. The non-differentiable nature of the spike function in SNNs complicates the direct application of backpropagation, a staple in training traditional CNNs. Innovations such as conversion methods, surrogate gradients, and hybrid learning models illustrate significant progress, yet they also underscore the complexity of adapting deep learning techniques to neuromorphic hardware. Each proposed method, whether it be backpropagation through time or unsupervised learning via STDP, presents trade-offs in terms of computational efficiency, learning accuracy, and hardware compatibility.

\subsection{SDBNs}
Spiking Deep Belief Networks (SDBNs) extend the conventional Deep Belief Networks (DBN) by incorporating spiking neurons. In SDBNs, traditional binary or continuous units of DBNs are replaced with spiking neurons, typically modeled using IF or more biologically detailed models. The architecture of an SDBN largely mirrors that of a traditional DBN, with multiple layers of spiking neurons arranged in a greedy layer-wise fashion. Each layer is composed of Restricted Boltzmann Machines (RBMs) or their spiking counterparts (sRBMs), which are trained sequentially.

Similar to classical RBMs, sRBMs in SDBNs learn to reconstruct their input while capturing useful features. However, instead of using visible and hidden units with static activations, sRBMs deal with spike trains and temporal coding \cite{tavanaei2019deep}. Each sRBM layer is trained independently to maximize the likelihood of the spike data reconstructed from the layer below. The training method known as Contrastive Divergence (CD) serves as an approximation to a maximum-likelihood learning algorithm \cite{neftci2014event}. Each sRBM layer's output serves as the input to the next layer. Once all the layers are pre-trained, the entire SDBN can be fine-tuned using supervised learning if labeled data is available. This typically involves a spiking version of backpropagation \cite{bohte2000spikeprop}.

Several studies have reported on the capabilities and improvements of SDBNs on various computational tasks. These include discussions on learning theory \cite{lee2007sparse, neftci2014event, kaiser2017spiking} and classification tasks \cite{o2013real, stromatias2015robustness, stromatias2015scalable, merolla2011digital, neil2014minitaur}. Table \ref{table:sdbns} provides a summary of key contributions, model details, and training methods for various SDBN works.

As a precursor to developing SDBNs, it's key to first establish an effective spiking model of the RBM, which serves as the fundamental unit in traditional DBNs. In \cite{neftci2014event}, a foundational sRBM has been proposed using IF neurons to perform MCMC (Markov Chain Monte Carlo) sampling of the Boltzmann distribution necessary for training RBMs. This includes implementing neurons whose firing rates are modulated by their membrane potential, approximating the sigmoidal activation function required in traditional RBMs. The method incorporates an adaptation of CD that operates in an online and asynchronous manner suitable for neuromorphic hardware. The event-driven CD uses STDP to adjust synaptic weights for continuous learning. The authors test their model on the MNIST dataset, achieving an accuracy of 91.9\%, close to the 93.6\% accuracy achieved with traditional RBMs using standard CD and Gibbs sampling.

\begin{itemize}

\item In \cite{lee2007sparse}, a sparse variant of DBNs is proposed to model the response properties of the visual cortex area V2. The primary motivation behind this work was to extend hierarchical learning models, typically validated against the simpler computations in visual area V1, to deeper levels of cortical hierarchy seen in V2. The proposed model is a sparse variant of Hinton's deep belief network \cite{hinton2006fast}. It includes two layers of nodes where the first layer learns localized, oriented edge filters, and the second layer encodes correlations of these first-layer responses to capture higher-order features like contours, corners, and junctions. The authors further extended this work in \cite{lee2011unsupervised} by implementing a convolutional approach to the sparse DBN. The convolutional sparse DBN architecture exploited the translational invariance inherent in natural images, making the learning process more efficient and robust to variations in object positioning and scale. Notably, these models are not SNNs, but they have contributed to the foundational techniques for using DBNs in visual object detection tasks. Additional research has applied DBNs and their convolutional variants across a range of visual tasks \cite{susskind2008generating, liu2014facial, mleczko2015rough, krizhevsky2010convolutional}.

\item Building on the sRBM and event-based Contrastive Divergence (CD) training method introduced in \cite{neftci2014event}, the authors of \cite{kaiser2017spiking} introduce the Spiking Convolutional DBN (SCDBN), allowing unsupervised learning of features from event data. The SCDBN builds on the sRBM by integrating convolutions. The SCRBM is a foundational component of the SCDBN. This work is significant because it integrates convolutional operations with spiking neurons, enabling efficient processing of spatiotemporal patterns in event-based data. Each SCRBM contains a convolutional layer where input spike trains are convolved with learned kernels. These kernels are shared across different receptive fields, thus reducing the number of parameters and ensuring that the same features can be detected regardless of their position in the input space. To encourage sparsity and reduce redundancy in the feature maps, lateral inhibitions are introduced in two types:
\begin{itemize}
    \item Intra-feature Map Inhibitions: Neurons within the same feature map inhibit each other, enhancing feature sparsity and reducing the likelihood of neighboring neurons firing simultaneously.
    \item Inter-feature Map Inhibition: Neurons located at the same spatial position in different feature maps inhibit each other. This enforces the learning of discriminative features across maps.
\end{itemize}
The SCRBMs are trained using CD, with a STDP local learning rule. Multiple layers of SCRBMs are stacked to form the DBN. Each SCRBM layer's output serves as the input to the next layer. The SCDBN was evaluated on two datasets: a self-recorded DVS dataset of balls, cans, and pens (Ball-Can-Pen Dataset), and the standardized Poker-DVS dataset. On the Poker-DVS, the network achieved classification accuracy of 90\%.
\end{itemize}

The conversion of ANNs to their spiking counterparts remains a common method for SNN training, and this methodology has been similarly applied in the context of SDBNs. 

\begin{itemize}

\item In \cite{o2013real}, the first SDBN was proposed as a traditional DBN converted to work with spiking neurons. Initially, the DBN is trained in a conventional manner using a dataset. The training involves adjusting the network weights in an unsupervised manner to minimize the reconstruction error using CD. After training, the learned weights are transferred to the corresponding spiking network. Each neuron's potential and firing threshold settings are adapted based on the Siegert approximation \cite{siegert1951first} to reflect the trained behavior in the spiking model. In the operational phase, the network processes real-time spiking inputs generated from a DVS. Evaluated on the MNIST dataset, the converted model using LIF neurons achieved 94.09\% accuracy, very close to that of the traditional DBN, which achieved 97.48\%.

\item In \cite{stromatias2015robustness}, the DBN conversion method was extended for efficient implementation on neuromorphic hardware. The study explored the impact of limited-bit precision during training and operation phases, as well as the impacts of noise inherent in the sensor signal and neuromorphic hardware, to quantify how these factors degrade network performance. Their significant contribution lies in demonstrating that spiking DBNs can operate effectively with extremely low-precision weights, making them suitable for deployment on resource-constrained neuromorphic platforms. The authors propose a training approach called "dual-copy rounding," in which both high-precision (WH) and low-precision weights (WL) are maintained during training. After each update, the WH matrix is quantized to create the WL matrix, which is rounded to the nearest representable value in the lower precision format. While the WL matrix is used for network operations during the forward pass to compute neuron activations and during backpropagation to calculate gradients, the gradient updates are applied to the WH matrix. The WH set accumulates subtle adjustments from the training process that might be lost if only WL were updated directly due to quantization thresholds. The rounding step mirrors the actual operation conditions that the network will face on the target hardware, where only lower precision weights can be used due to hardware limitations. The research demonstrated that spiking DBNs can function effectively with very low levels of hardware bit precision, down to approximately two bits. Experimental results on MNIST demonstrate 82\% accuracy under spiking conditions but note that this result, using dual-copy rounding, is twice the accuracy of the post-learning quantization method.

\end{itemize}

The following three studies emphasize the advancements in implementations of spiking DBNs and RBMs and leverage the parallelism and event-driven nature of neuromorphic hardware systems to efficiently process data, a key aspect in the field of neuromorphic computing.
\begin{itemize}
    \item In \cite{stromatias2015scalable}, the process of mapping SDBN parameters from software implementations to a neuromorphic hardware platform is highlighted. The authors implement their converted SDBN on the SpiNNaker platform. The network is trained offline using CD, and then the trained model parameters are deployed on SpiNNaker. Input images are converted into Poisson spike trains to simulate DVS input. The spike-based DBN achieved classification accuracy of 95\% on MNIST, operating in real-time with a mean classification latency of 20ms. The network was also demonstrated to consume only 0.3W during operation, demonstrating substantial energy efficiency compared to traditional CPU and GPU hardware platforms.

    \item In \cite{merolla2011digital}, an S-RBM is implemented on a custom-designed neurosynaptic core. This work is significant as it represents one of the early demonstrations of spiking RBMs on specialized neuromorphic hardware, emphasizing the potential for low-power, high-performance machine learning applications. The core uses binary weights and neuronal models to simulate the RBM's functioning, with the classification task being offloaded to an off-chip linear classifier. The weights are first learned offline using CD and then transformed into a hardware-compatible binary format. The system demonstrated a 94\% accuracy on MNIST while consuming only 4.5e-5 $\mu$J per spike.

    \item In \cite{neil2014minitaur}, the Minotaur system is discussed, an FPGA-based platform that executes spiking versions of RBMs and DBNs. This system replaces traditional sigmoidal or ReLU activation functions with LIF neurons post-training and normalizes the weights to suit the spiking network dynamics. The DBN implemented on the Minotaur system achieved a 92\% accuracy rate on MNIST.
\end{itemize}

\begin{table*}[t]
\centering
\caption{Summary of methodologies and key innovations in Spiking Recurrent Neural Networks (SRNNs) papers.}
\begin{tabular}{|p{3cm}|p{5cm}|p{5cm}|}
\hline
\textbf{Paper} & \textbf{Methodology} & \textbf{Key Innovations} \\ \hline
Shrestha et al., 2017 \cite{shrestha2017spike} & Reconceptualize LSTM components for SNN, approximate sigmoid and tanh using piece-wise linear functions, dual-channel spike encoding & Conversion of continuous weight values into constrained range, rate-coded and burst-coded spike mechanisms \\ \hline
Diehl et al., 2016 \cite{diehl2016conversion} & "Train-and-constrain" process: train RNN with backpropagation, then discretize weights and activities for TrueNorth & Discretization into 16 levels, fan-in constraint of 64 inputs per neuron \\ \hline
Kim et al., 2019 \cite{kim2019simple} & Transfer learned weights from rate RNN to SRNN composed of LIF neurons & Weight scaling to accommodate spiking paradigms \\ \hline
Lotfi et al., 2020 \cite{lotfi2020long} & Approximation of gradient of spike function, surrogate gradients & Gradient approximation using piecewise linear/probability distribution, spike function error propagation \\ \hline
Xing et al., 2020 \cite{xing2020new} & Combine convolutional operations with recurrent dynamics, train using SLAYER algorithm & 3D convolution, recurrent layers for temporal dynamics \\ \hline
Neil et al., 2016 \cite{neil2016phased} & Incorporate time gate controlled by oscillation & Phased updates to memory cells, efficient processing of asynchronous inputs \\ \hline
Yin et al., 2020 \cite{yin2020effective} & Fully interconnected recurrent layers, surrogate gradients for training & Adaptive spiking neuron modulating firing threshold \\ \hline
Wang et al., 2016 \cite{wang2016d} & Deep LSM (D-LSM) architecture combining LSMs with CNNs & Multiple LSM stages with pooling stages, unsupervised training with STDP, high accuracy on MNIST dataset \\ \hline
Patino et al., 2022 \cite{patino2022liquid} & Implementation of LSM on SpiNNaker neuromorphic processor & Mapping LSM on SpiNNaker for spatio-temporal tasks, offline training with BPTT, high accuracy on N-MNIST dataset \\ \hline
Kasabov et al., 2014 \cite{kasabov2014neucube} & NeuCube 3D SNN architecture for brain mapping & 3D SNN reservoir, unsupervised STDP learning, high accuracy on EEG data and visual recognition tasks \\ \hline
Paulun et al., 2018 \cite{paulun2018retinotopic} & Adaptation of NeuCube for Dynamic Vision Sensor (DVS) outputs & Encoding algorithm inspired by retinal ganglion cells, high accuracy on MNIST-DVS dataset \\ \hline
\end{tabular}
\label{tab:srnns}
\end{table*}

\subsubsection{Discussion}
The hierarchical structure of SDBNs, which mirrors that of traditional DBNs, creates a layer-wise greedy learning approach. This is critical because it enables each layer to learn increasingly complex representations, a method that has proven effective in deep learning but is implemented here with a fundamentally different computational paradigm. By using mechanisms such as CD and STDP, SDBNs can perform unsupervised learning to capture useful features without requiring labeled data for every phase of training. In fact, SDBNs can handle both unsupervised learning (during the layer-wise training phase) and supervised learning (during the fine-tuning phase) effectively. This dual capability is key for applications requiring initial feature extraction from unlabeled data followed by task-specific tuning.

Despite their advantages, the implementation of SDBNs also presents several challenges. The primary issue lies in the complexity of training spiking networks. The training process, such as the adaptation of CD for spike-based systems, needs to be refined to improve both efficiency and effectiveness. Moreover, the transfer of learned weights from ANN to SNNs, while useful, often requires additional steps to adjust the dynamics of the model to match those of the spiking neurons, as seen in the Siegert approximation method. While there have been significant advances in the accuracy of these models on standard datasets like MNIST, there is a need to test these networks on more complex, real-world datasets.

\subsection{SRNNs}
\begin{figure}[b!]
    \centering
    \includegraphics[scale = .35]{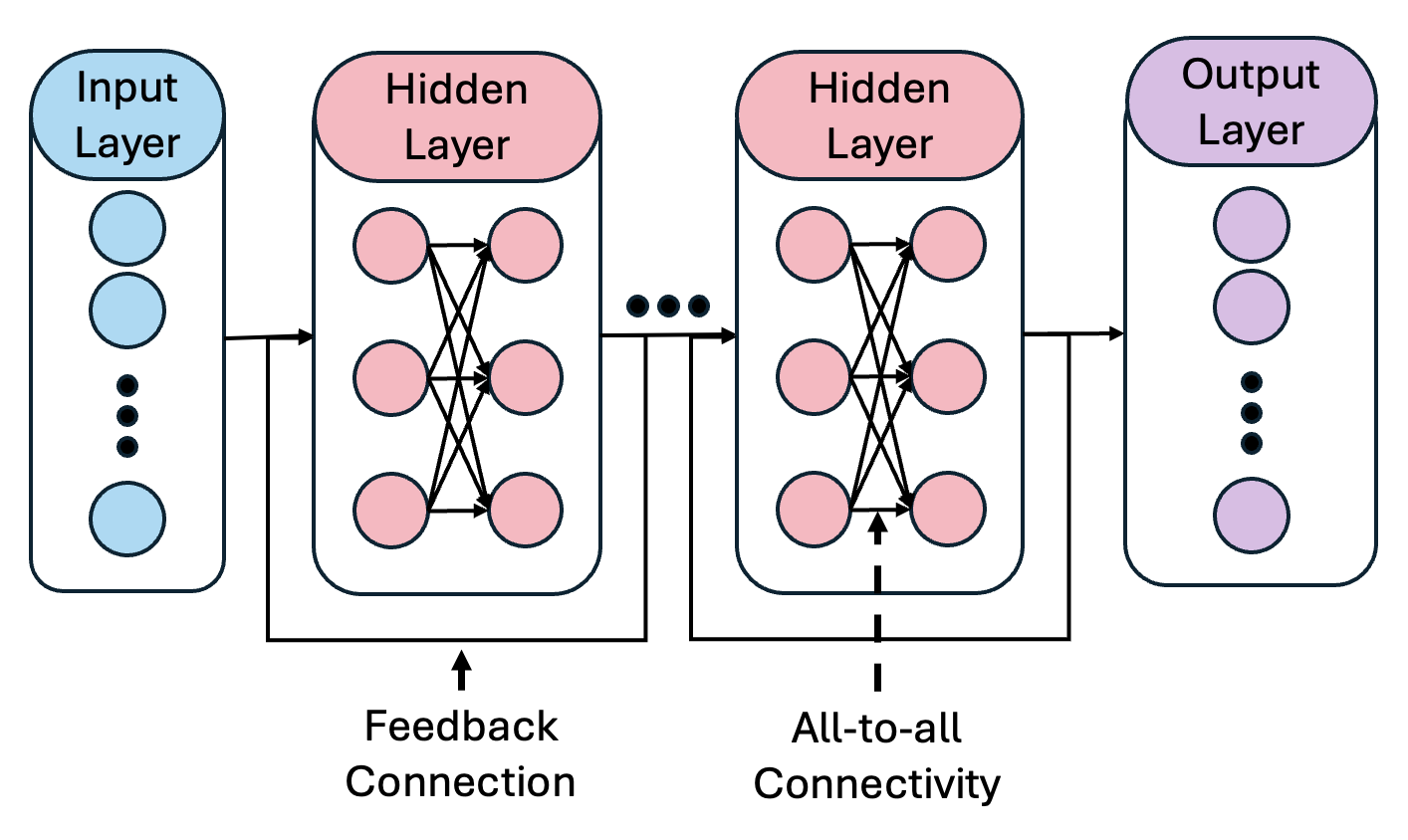}
    \caption{Architecture of a Spiking Recurrent Neural Network (RNN) with two hidden Layers. The network features two hidden layers, each consisting of fully connected spiking neurons. Feedback connections for each hidden layer allow neurons to exhibit recurrent behavior. Each neuron in the hidden layers are connected to every afferent neuron.}
    \label{fig:srnn}
\end{figure}

Spiking Recurrent Neural Networks (SRNNs) are a class of neural networks designed to model and process information in a manner that closely mimics the behavior of biological neurons. Any network trained with backpropagation is implicitly a recurrent neural network (RNN) due to the recurrence of the training algorithm \cite{hochreiter2001gradient}. However, for the purpose of this review, a network is considered to be recurrent if it includes explicit feedback connections where outputs from the network are fed back to itself as inputs for subsequent processing steps. An example of such a network architecture is shown in Figure \ref{fig:srnn}. This definition focuses on architectural features that enable the network to process sequential and temporal data directly through cycles in its graph structure rather than indirect implications of recurrence from a learning algorithm. Recent research has focused on adapting RNN architectures, such as the Long Short-Term Memory (LSTM) networks, into the spiking domain. 

Methodologies have been proposed for adapting traditional RNNs to the constraints and functionalities of neuromorphic hardware \cite{shrestha2017spike, diehl2016conversion, kim2019simple}. Table \ref{tab:srnns} gives a summary of methodologies and key innovations in modern SRNNs.

\begin{itemize}
    \item In \cite{shrestha2017spike}, a spiking LSTM is implemented on IBM's TrueNorth neurosynaptic system. This work aims to overcome challenges associated with mapping RNN architectures, specifically LSTMs, onto spike-based, event-driven hardware. TrueNorth's architecture, designed primarily for SNNs, poses constraints on connectivity and synaptic weight resolution, which complicates the implementation of LSTMs that require high precision for effective learning and performance. This process required a reconceptualization of traditional LSTM components to operate within the constraints of a SNN. The spiking LSTM required the conversion, by scaling and rounding, of continuous weight values of a traditional LSTM into a constrained range. The sigmoid and hyperbolic tangent activation functions were approximated using piece-wise linear functions suitable for implementation on the spiking hardware. The approximations used hard tanh and hard sigmoid functions, which are simpler and require fewer computation steps. To handle both positive and negative values, a dual-channel spike encoding scheme was used. Positive and negative input values were encoded separately into two channels of spikes, which were then processed to determine the net input to the neurons. The internal states used rate-coded and burst-coded spike mechanisms. The main thrust of the paper was to overcome the complications in mapping the LSTM to a neuromorphic chip, and accuracy results on standard benchmarks were not reported. This work was early-stage research where the feasibility and development of techniques were in the spotlight, and benchmarking results were not provided.

    \item In \cite{diehl2016conversion}, a methodology is proposed for the conversion of traditional RNNs into SRNNs suitable for low-power neuromorphic hardware such as IBM's TrueNorth. Their significant contribution lies in the "train-and-constrain" conversion process, which enables trained RNNs to be adapted for deployment on neuromorphic hardware while retaining functional performance. The conversion process involves training the RNN with standard backpropagation through time and then discretizing the weights and neural activities for compatibility with the constraints imposed by the TrueNorth architecture. TrueNorth requires the discretization of synaptic weights and neural activities into 16 levels, and the fan-in per neuron must not exceed 64 inputs. This work primarily addresses the application of this conversion process to a natural language processing task of question classification, achieving a proof-of-concept implementation that retains functional performance while substantially reducing power consumption. The RNNs are initially trained on conventional hardware using gradient descent. Post-training, weights and neuron outputs are discretized to fit the limited precision available on TrueNorth. This involves mapping the continuous values from the trained RNN into a finite set of values that TrueNorth can handle. Performance analysis shows that the converted SNN achieves 74\% accuracy in question classification with an estimated power consumption of about 17 $\mu$W.

    \item In \cite{kim2019simple}, a framework is proposed for constructing SRNNs with comparable efficiency to continuous-variable rate RNNs. Their key contribution is in developing a simple yet effective method for transferring learned weights from rate-based RNNs to spiking RNNs composed of Leaky Integrate-and-Fire (LIF) neurons. The design of the network starts with a rate RNN, trained on tasks using traditional gradient descent. The learned weights, along with synaptic time constraints, are transferred to an SRNN composed of LIF neurons. A key aspect of the transfer involves scaling the weights to accommodate differences between rate and spiking paradigms. This work was not evaluated on any standard benchmarking datasets, but functionality is demonstrated on a Go-NoGo task and a complex context-dependent sensory integration task.
\end{itemize}

\noindent Innovative designs for SRNNs have driven advancements in neuromorphic computing. By leveraging both supervised and unsupervised learning paradigms, these networks are natively trained to exploit the intrinsic temporal dynamics and energy efficiency of spike-based mechanics \cite{lotfi2020long, xing2020new, neil2016phased, yin2020effective}.

\begin{itemize}
    \item In \cite{lotfi2020long}, a framework for training LSTM-based SNNs was introduced. Their significant contribution lies in enabling effective error backpropagation in spiking LSTMs by approximating the gradient of the spike function, thereby bridging the gap between conventional LSTMs and their spiking counterparts. The derivative of a non-differentiable spike function is approximated using a piecewise linear function or probability distribution approach (such as a Gaussian distribution). During backpropagation, the error is propagated through the network by considering the proximity of the membrane potential of neurons to their firing threshold. The gradients are used to update the network parameters using standard optimization techniques (SGD \cite{ruder2016overview} or Adam \cite{kingma2014adam}). Several datasets were tested to evaluate the performance of the LSTM-based SNN. On MNIST, a test accuracy of 98.3\% was reached. On EMNIST, an accuracy of 83.75\% was reached.

    \item In \cite{xing2020new}, a Spiking Convolutional RNN (SCRNN) is described that combines convolutional operations with recurrent neural network dynamics, specifically designed to process data from a neuromorphic camera. This work is significant as it integrates spatiotemporal feature extraction and temporal dynamics in a spiking neural network, enhancing performance on event-based data. The SCRNN uses a 3D convolution operation to process the input data through a series of convolutional filters outputting feature maps of spike trains. After each convolutional layer, the network integrates recurrent layers where the output of each layer at a given time step serves as an input to the same layer at the next time step. This recurrence helps to capture temporal dynamics, and is key for tasks where the order and duration of movements are important. The SLAYER algorithm was used to train the SCRNN. The effectiveness of the SCRNN was measured by its accuracy in classifying different hand gestures from the DVS-Gestures dataset. The SCRNN reached an accuracy of 96.59\% for the 10-class gesture recognition task.

    \item In \cite{neil2016phased}, the Phased LSTM is proposed as an extension of the traditional LSTM model specifically designed to handle event-based sequences that are irregularly sampled. The mechanism incorporates a time gate controlled by a parameterized oscillation. This allows the model to update its memory cell only during specific phases of the oscillation cycle to efficiently process inputs from sensors with varying and asynchronous sampling rates. The oscillation parameters for period, open phase ratio, and phase shift can be learned during training. Although this is not strictly an SNN, the ability of the Phased LSTM to update its memory cells only during specific phases makes them suitable for neuromorphic datasets that exhibit similar asynchronous behavior. This work was tested on the N-MNIST dataset, where the Phased LSTM (97.27\%) outperformed both traditional LSTMs (96.55\%) and CNNs (73.71\%). Notably, the Phased LSTM also required significantly fewer updates per neuron - only 159 compared to the traditional LSTM's 3153, illustrating a substantial reduction in computational load. This efficiency is attributed to its gating mechanism that restricts updates to specific phases within the internal oscillatory cycle.

    \item In \cite{yin2020effective}, an SRNN architecture is proposed featuring layers of LIF and Adaptive spiking neurons fully interconnected both recurrently within the same layer and forward across layers. Their significant contribution is the use of surrogate gradients in combination with adaptive spiking neurons to effectively train SRNNs using backpropagation through time (BPTT). The Adaptive spiking neuron self-modulates the firing threshold based on neuronal activity, such that after each spike, the threshold is increased and then decays back towards a baseline over time. To handle the non-differentiable nature of spikes, the authors use a surrogate gradient. The SRNNs are trained using BPTT to unroll the NN in time and compute the gradient of the loss function with respect to the weights. The implementation uses PyTorch's auto-differentiation capabilities to simplify the integration of surrogate gradients in the training loop. The SRNNs were evaluated on several benchmarking datasets, including MNIST and permuted MNIST, with accuracy scores of 97.82\% and 91.0\% reported, respectively.
\end{itemize}

\subsubsection{Liquid State Machines}
\begin{figure}[t!]
    \centering
    \includegraphics[scale = .4]{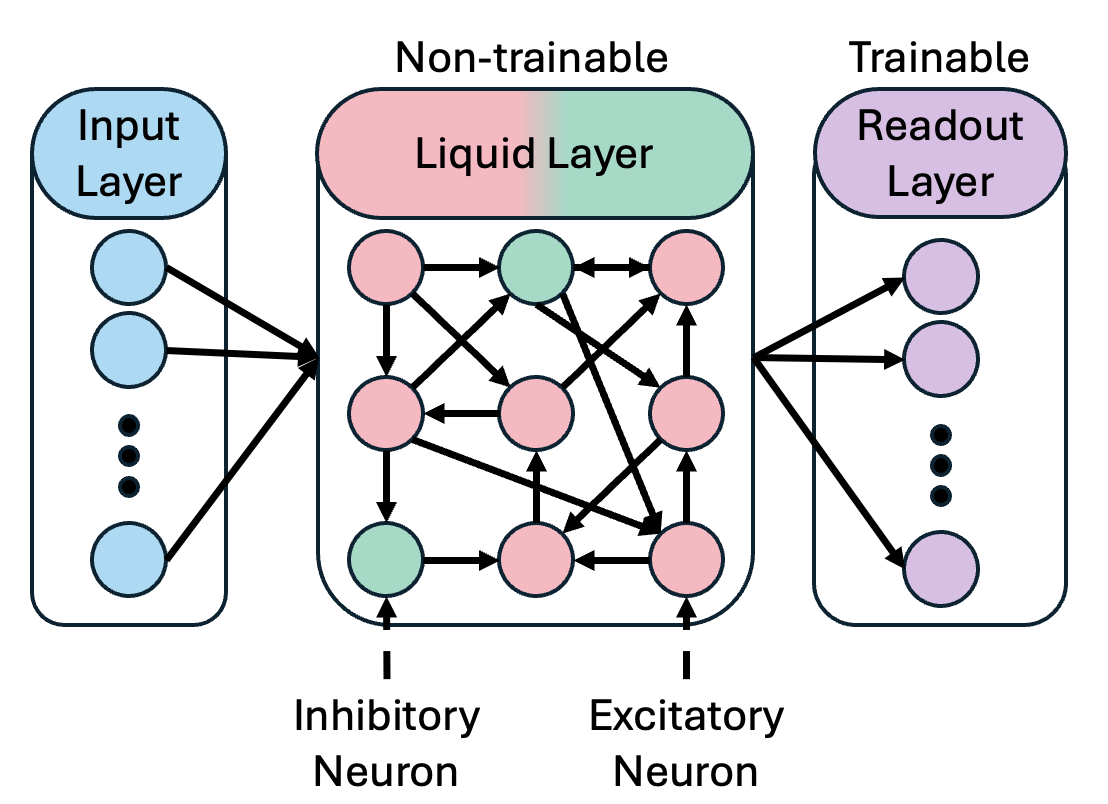}
    \caption{Overview of a Spiking Liquid State Machine (SLSM). The input signal is fed into a non-trainable liquid reservoir, composed of a dynamic and randomly connected network of excitatory and inhibitory neurons. The readout layer, which is trainable, interprets these patterns to produce the final output.}
    \label{fig:slsm}
\end{figure}

Research efforts to mimic neocortical computations involve the use of the Liquid State Machine (LSM), a biologically plausible model of computation for recurrent SNNs that blends into the broader domain of reservoir computing \cite{lukovsevivcius2009reservoir, schrauwen2007overview}. LSMs are significant because they offer a framework for processing temporal data using fixed recurrent networks, requiring training only in the readout layer, which simplifies the learning process and reduces computational requirements. This method uses a neural reservoir consisting of sparsely connected excitatory and inhibitory spiking neurons. The structure is designed to function as a universal analog memory, capable of fading over time, and it effectively transforms varied spike or continuous input streams into dynamic spatiotemporal patterns. These patterns are then decoded by a set of linear units. An example of the Spiking LSM (SLSM) network architecture is shown in Figure \ref{fig:slsm}.

The reservoir computing model can be broken down into two key components: the liquid (reservoir) and the readout.
\begin{enumerate}
    \item The liquid: The core of the model, a large, sparsely connected network of neurons with fixed weights, acts as a dynamic temporal reservoir. The liquid's role is to project the input into a high-dimensional space where the linear separability of different input patterns increases. Input signals are fed into this liquid, causing state changes within the network that reflect both the current and recent past inputs due to the memory properties of the network's dynamics. The liquid layer typically consists of heterogeneous spiking neurons. Connections between liquid neurons are recurrent and possibly random, with a balance between excitation and inhibition to maintain the network's dynamic stability.

    \item The readout layer: Linear or nonlinear readout neurons are used to decode the spatiotemporal patterns represented in the liquid's state to perform specific tasks, such as classification or prediction. Unlike the fixed liquid, the readout layer is trainable using gradient descent \cite{maass2011liquid}.
\end{enumerate}

Building on the foundational concepts of LSMs, there are several research works that explore its applications to vision-based recognition tasks \cite{wang2016d, patino2022liquid, kasabov2014neucube, paulun2018retinotopic}.

\begin{itemize}
    \item In \cite{wang2016d}, the Deep LSM (D-LSM) is introduced as an architecture that combines the recurrent processing capabilities of LSMs with the depth characteristic of CNNs. This work is significant as it enhances the feature extraction capabilities of LSMs by stacking multiple reservoir layers, enabling the processing of complex spatiotemporal patterns. The core architecture of D-LSM comprises multiple LSM stages interleaved with pooling stages, similar to the layers in CNNs. Each LSM stage consists of spiking reservoirs acting as nonlinear spatiotemporal filters that process input spike trains to produce higher-dimensional representations. These representations are then subsampled in the pooling stages to reduce dimensionality and prepare for subsequent layers. Training of the D-LSM is done in two phases: unsupervised training of the recurrent reservoirs using STDP and supervised training of the final readout stage. This training approach aims to optimize the spiking reservoirs for coarse feature extraction and fine-tune the readout layer for classification. The authors evaluate the D-LSM using a subset of MNIST and achieve 95.6\% for a D-LSM setup with two LSM stages and one pooling stage with STDP applied to tune the reservoir; the recognition accuracy was 97.2\%.

    \item In \cite{patino2022liquid}, an implementation of LSM on the SpiNNaker neuromorphic processor is proposed to address spatiotemporal classification tasks. The authors demonstrate mapping and operationalizing the LSM on the SpiNNaker platform for processing neuromorphic sensor inputs. The LSM used LIF neurons trained using BPTT. The LSM was trained offline using PyTorch on a GPU, with only the readout layers being trained, while the recurrent network weights were kept static. The trained model parameters were transferred to the SpiNNaker platform using the spyNNaker tool to map the PyTorch-defined SNN onto the neuromorphic hardware. The system achieved an accuracy of 94.43\% on the N-MNIST dataset.

    \item The NeuCube is a 3D SNN architecture designed to map and mimic the complex interconnectivity of the human brain's structural and functional regions. Initially introduced in \cite{kasabov2014neucube}, the NeuCube's core consists of a three-dimensional SNN reservoir. This reservoir is structured to capture the essence of the brain's network of interconnected regions through the mapping of these regions onto spiking neurons positioned within a virtual 3D space. The NeuCube uses an unsupervised learning mechanism based on STDP to train the SNN for effective input data capture. It also uses supervised learning techniques to develop an output classifier capable of recognizing and interpreting patterns from the SNN's activities. Once trained, NeuCube acts as an associative memory, able to reproduce spatiotemporal trajectories of neuronal activity even when only partial input data is provided. This capability was first demonstrated in a case study where the NeuCube classified EEG data for various cognitive tasks. 
    
    \item Expanding upon its initial applications in \cite{paulun2018retinotopic}, the authors adapted the NeuCube model to process outputs from a Dynamic Vision Sensor (DVS) for visual recognition tasks. The authors discuss an encoding algorithm that compresses and pools the DVS output into fewer channels before feeding it into the NeuCube. This was inspired by the structure of retinal ganglion cells, which aggregate input from multiple photo-receptors. The algorithm adjusts the pooling based on the proximity to the central vision region, where fewer pixels are pooled more densely, reflecting the higher resolution of human central vision. The NeuCube's efficacy in this new domain was validated using the MNIST-DVS dataset, achieving a classification accuracy of 92.90\%.
\end{itemize}

\subsubsection{Discussion}
The transformation of traditional RNN architectures, particularly LSTMs, into their spiking counterparts, poses significant engineering challenges primarily due to the binary nature of spikes and the constraints of neuromorphic hardware architectures. As discussed in the works of \cite{shrestha2017spike} and \cite{diehl2016conversion}, the need to discretize and simplify the neuron models without substantial loss of functional accuracy underscores the complexity of adapting recurrent models to spike-based systems. The methodologies developed, such as the "train-and-constrain" technique, not only allow for the hardware-specific tuning of the networks but also pave the way for future innovations in hardware-software co-design. Innovative designs, such as dual-channel spike encoding and the use of rate-coded and burst-coded spike mechanisms, represent significant advancements in making SRNNs more adaptable and efficient. The integration of convolutional architectures within recurrent frameworks, as seen in \cite{xing2020new} and the new design of the Phased LSTM \cite{neil2016phased}, indicate a growing maturity in SRNN architectures. Moreover, the exploration of LSMs and architectures like NeuCube demonstrates the potential of reservoir computing models in efficiently processing spatiotemporal data on neuromorphic hardware, further expanding the capabilities of SRNNs in real-world applications.

\begin{table*}[t!]
\centering
\caption{Evaluation Metrics for SNNs}
\label{tab:eval}
\begin{tabular}{|p{2cm}|p{5cm}|p{3cm}|p{1.5cm}|}
\hline
\textbf{Metric} & \textbf{Description} & \textbf{Importance} & \textbf{Unit} \\ \hline
Classification Accuracy & Quantifies the percentage of correctly recognized samples in relation to the total number of samples examined. & Pivotal metric for object classification tasks. & Percentage (\%) \\ \hline
Precision & Measures the accuracy of positive predictions made by the model. Calculated as the ratio of true positives to the sum of true positives and false positives. & Indicates the accuracy of relevant object identification. & Ratio \\ \hline
Recall & Assesses the ability of the model to detect all relevant objects in the dataset. Calculated as the ratio of true positives to the sum of true positives and false negatives. & Indicates the effectiveness in identifying most relevant objects. & Ratio \\ \hline
F1 Score & The harmonic mean of precision and recall, providing a single metric that balances both. & Offers a comprehensive measure of model accuracy. & Ratio \\ \hline
Intersection-over-Union (IoU) & Quantifies the overlap between predicted bounding boxes and ground truth bounding boxes. Computed as the ratio of the intersection area to the union area of the bounding boxes. & Measures the accuracy of object localization. & Ratio \\ \hline
Mean Average Precision (mAP) & Evaluates the precision-recall curve across different classes and IoU thresholds. Calculated by averaging the AP values obtained for each class and IoU threshold. & Provides a comprehensive assessment of overall performance in object detection tasks. & Average \\ \hline
Energy Consumption & Quantifies the power requirements of a model during inference. Measured in joules or watts. & Critical for models deployed on resource-constrained devices. & Joules or Watts \\ \hline
Latency & Refers to the time delay between input and the corresponding output during the inference phase. Measured in milliseconds. & Important for real-time applications. & Milliseconds (ms) \\ \hline
Memory Footprint & Measures the amount of memory required to store and execute a model, including synaptic weights, neuron states, and infrastructure for spike communication. & Essential for optimizing storage and execution of models. & Bytes or Megabytes (MB) \\ \hline
\end{tabular}
\end{table*}

\section{Evaluation}
\label{sec:evaluation}
This section elucidates the performance evaluation metrics of SNNs in CV tasks by focusing on their ability to deliver accurate and efficient results during inference. Accuracy reflects how well SNNs perform tasks such as classification and object detection, while efficiency measures the computational resources required, including energy usage, speed, and memory consumption, making these metrics essential for assessing the practicality of SNNs in real-world applications.Table \ref{tab:eval} depicts the evaluation metrics for SNNs in CV tasks, including their descriptions, importance, and units.

\paragraph{Accuracy Metrics}
Various metrics have been proposed to measure the accuracy of SNNs on visual tasks. As classification and object detection exist on a spectrum of complexity, so do the metrics that are used to grade them.

Object classification serves as a fundamental task in computer vision, where the objective is to assign a class label to the input image. Classification Accuracy is a pivotal metric, quantifying the percentage of correctly recognized samples in relation to the total number of samples examined. In the context of SNNs, classification is often determined by summing the firing rates of output neurons during a specific time window in Rate Coding methods \cite{yamazaki2022spiking, pietrzak2023overview}. Alternatively, in Time-to-First-Spike encoding, the first neuron to spike dictates the prediction \cite{bonilla2022analyzing}.

Object detection, a more complex task, involves predicting multiple objects within a single image and necessitates additional accuracy metrics. To evaluate an object detection model effectively, it is essential to assess both the accuracy of the detection and the localization processes. Detection refers to the identification of objects within an image, while localization involves accurately determining the spatial extent or location of each detected object. In this context, precision, recall, Intersection-over-Union (IoU), and Mean Average Precision (mAP) are pivotal evaluation metrics for assessing the performance of object detection models. 

\begin{itemize}
    \item {\bf Precision} measures the accuracy of positive predictions made by the model. It is calculated as the ratio of true positives to the sum of true positives and false positives. In the context of object detection, precision quantifies how many of the predicted objects are actually relevant to the task. A high precision indicates a low rate of false positives, indicating accurate identification of relevant objects.

    \begin{equation}
    \text{Precision} = \frac{\text{True Positives}}{\text{True Positives} + \text{False Positives}}
    \end{equation}

    \item {\bf Recall} assesses the ability of the model to detect all relevant objects in the dataset. It is calculated as the ratio of true positives to the sum of true positives and false negatives. In object detection, recall measures how many of the relevant objects were successfully detected by the model. A high recall indicates that the model can effectively identify a high number of the relevant objects in the dataset.

    \begin{equation}
    \text{Recall} = \frac{\text{True Positives}}{\text{True Positives} + \text{False Negatives}}
    \end{equation}

    \item {\bf F1 Score} is the harmonic mean of precision and recall. It provides a single metric that balances precision and recall, offering a comprehensive measure of a model's accuracy. 

    \begin{equation}
    \text{F1 Score} = 2 \times \frac{\text{Precision} \times \text{Recall}}{\text{Precision} + \text{Recall}}
    \end{equation}

    \item {\bf Intersection-over-Union (IoU)} quantifies the overlap between predicted bounding boxes and ground truth bounding boxes. It is computed as the ratio of the intersection area between the two bounding boxes to the union area of the bounding boxes. IoU ranges from 0 to 1, where 1 indicates perfect overlap between the predicted and ground truth bounding boxes, while 0 indicates no overlap. 

    \begin{equation}
    \text{IoU} = \frac{\text{Area of Intersection}}{\text{Area of Union}}
    \end{equation}

    \item {\bf Mean Average Precision (mAP)} evaluates the precision-recall curve across different classes and IoU thresholds in object detection tasks. It is calculated by averaging the Average Precision (AP) values obtained for each class and IoU threshold. AP measures the area under the precision-recall curve for a specific class and IoU threshold, providing insights into the model's performance in detecting objects of that class. mAP offers a comprehensive assessment of the model's overall performance in object detection across diverse scenarios, considering both precision and recall.

    \begin{equation}
    \text{mAP} = \frac{\text{AP}_1 + \text{AP}_2 + \ldots + \text{AP}_n}{n}
    \end{equation}   

   \item {\bf The Receiver Operating Characteristic (ROC) Curve} is a fundamental tool for evaluating the performance of classification models by plotting the True Positive Rate (TPR) against the False Positive Rate (FPR) across varying decision thresholds \cite{fawcett2006introduction}. TPR, also known as sensitivity or recall, measures the proportion of actual positive instances correctly identified, while FPR represents the proportion of negative instances incorrectly classified as positive. The ROC Curve visualizes the trade-off between sensitivity and specificity (the true negative rate), aiding in the selection of specific thresholds for decision-making \cite{hanley1982meaning, litjens2017survey}.

    An important metric derived from the ROC Curve is the Area Under the Curve (AUC), which quantifies the overall ability of the model to discriminate between classes. As the name suggests, the AUC is the area under the ROC curve, providing a single scalar value that summarizes the model's performance across all classification thresholds. An AUC of 1.0 indicates perfect classification, while an AUC of 0.5 suggests no discriminative ability beyond random chance.

    The ROC Curve originates from signal detection theory, a subset of detection and estimation theory developed in the 20th century for analyzing radar and sonar signals \cite{green1966signal}. Initially used to measure the ability of operators to discern signal from noise, the ROC analysis provided a systematic way to evaluate detection performance under uncertainty. This framework was later adopted in various fields, including machine learning, to assess the performance of classifiers \cite{metz1978basic}.

    In the context of CV applications using SNNs, the ROC Curve and AUC offer insights into the model's discriminative capabilities across multiple classes or output neurons, especially when dealing with imbalanced datasets or varying classification thresholds \cite{tavanaei2019deep}.

\end{itemize}

\paragraph{Efficiency Metrics}
Efficiency metrics such as energy consumption, latency, and memory footprint are crucial for understanding the practical applicability of these models in resource-constrained environments.

\begin{itemize}
    \item {\bf Energy Consumption} The authors in \cite{sorbaro2020optimizing, lemaire2022analytical} quantify power requirements of a model during inference. This is especially important for models deployed on edge devices, where energy resources are limited \cite{davies2018loihi, deng2020tianjic}. Energy consumption can be measured in terms of joules or watts and is influenced by factors such as the complexity of the model architecture, the number of operations required per inference, and the hardware used for deployment. Techniques like model quantization, pruning, and the use of specialized hardware accelerators (e.g., GPUs, Tensor Processing Units (TPUs), and neuromorphic chips) can significantly impact the energy efficiency of a model.

    \item {\bf Latency} The authors in \cite{diehl2015fast, rueckauer2018conversion, hu2021spiking, sengupta2019going} report on the time delay between input and the corresponding output during the inference phase as latency. Latency is typically measured in milliseconds (ms) and is affected by the model's architecture, the efficiency of the inference engine, and the computational power of the hardware. Optimizations such as model compression, efficient layer designs, and the use of parallel processing can help reduce latency \cite{kundu2021towards, rathi2021diet, nageswaran2009efficient}.

    \item {\bf Memory Footprint} \cite{lemaire2022synaptic, putra2020fspinn} measures the amount of memory required to store and execute a model. The memory footprint encompasses the storage of synaptic weights, neuron states, and the infrastructure for spike communication. Optimizations such as weight sharing, synaptic pruning, and efficient encoding schemes can further minimize the memory footprint \cite{song2022spiking, faghihi2022synaptic, wang2022efficient}. Additionally, neuromorphic hardware uses memory hierarchies and local memory storage to efficiently handle the unique requirements of spiking computations \cite{indiveri2015memory, davies2018loihi}.
\end{itemize}

\section{SNN Framework for CV Practitioners In Action}

To enable hands-on understanding and experimentation, this section provides a guide to the Github repository \cite{Iaboni2024EventSNN}. The examples found within serve as a practical starting point for users who wish to explore SNNs and their applications in computer vision. By following these examples, users can easily execute and understand the core functionalities of the framework, including selecting and preparing datasets, configuring network architectures, training SNN models, and evaluating their performance.

\subsection{Repository Contents and Usage Guide}
This repository contains a collection of Jupyter notebooks demonstrating various implementations and concepts related to SNNs. Each notebook focuses on different aspects of SNNs, providing both theoretical background and practical implementations. In addition to the notebooks, the repository includes links to commonly used datasets, software toolkits, and source code for various encoding methods. Below is a summary of the key files and their usage:

\begin{enumerate}
    \item  \texttt{Basic\_Spiking\_Example.ipynb}
    \begin{itemize}
        \item {\bf Purpose:} Presents a basic example of SNN simulation using the Norse library.
        \item {\bf Usage:} Run this notebook to get started with fundamental concepts of spiking neurons and simple network simulations
    \end{itemize}

     \item  \texttt{FFSNN\_Tutorial.ipynb}
    \begin{itemize}
        \item {\bf Purpose:} Offers a tutorial on implementing a feedforward SNN using SNNTorch.
        \item {\bf Usage:} Work through this notebook to understand the basics of feedforward SNNs, including neuron models, network structure, and training methods.
    \end{itemize}

     \item  \texttt{SCNN\_Tutorial.ipynb}
    \begin{itemize}
        \item {\bf Purpose:} Provides a comprehensive tutorial on building SCNNs using PyTorch and the SNNTorch library.
        \item {\bf Usage:} Follow this notebook step-by-step to gain hands-on experience in constructing and training SCNNs.
    \end{itemize}

     \item  \texttt{SCNN\_STDP.ipynb}
    \begin{itemize}
        \item {\bf Purpose:} Implements a SCNN with STDP learning.
        \item {\bf Usage:} Run this notebook to understand the implementation of STDP in a convolutional architecture. It demonstrates how synaptic weights are updated based on the relative timing of pre- and post-synaptic spikes.
    \end{itemize}

     \item  \texttt{SCNN\_STDP\_Poissonencoding.ipynb}
    \begin{itemize}
        \item {\bf Purpose:} Similar to SCNN\_STDP.ipynb, but incorporates Poisson encoding for input spikes.
        \item {\bf Usage:} Execute this notebook to learn how Poisson encoding can be used to convert continuous input signals into spike trains for SNN processing.
    \end{itemize}

    \item Additional Resources
    \begin{itemize}
        \item {\bf Datasets:} The repository provides links to commonly used datasets in SNN research.
        \item {\bf Software Toolkits:} Links to popular SNN simulation and development toolkits are included.
        \item{\bf Encoding Methods:} Source code for various spike encoding methods is provided, allowing users to explore different techniques for converting input data into spike trains.
    \end{itemize}
\end{enumerate}

To use these notebooks, please follow the steps listed in the repository's README section.

These notebooks serve as both educational resources and practical examples for researchers and practitioners working with SNNs. They cover a range of topics from basic neuron models to advanced network architectures, providing a comprehensive starting point for SNN experimentation and development.

\subsection{Creating a Customized Example}
Beyond the prebuilt examples, this section walks users through creating and experimenting with their own custom SNN configurations. By selecting a dataset, defining an architecture, and specifying training parameters, users can adapt the framework to fit their specific needs. While the repository provides multiple examples to showcase different facets of SNNs, this guide will specifically focus on the SCNN example, helping users understand the framework's key functionalities. The example offers a comprehensive walkthrough on replicating and adapting the processes for custom tasks. Below is a step-by-step breakdown:

\subsubsection{Environment Setup:}
Before running the SCNN example, it is crucial to set up the correct environment with all necessary dependencies. Ensure that you have the following installed: 
\begin{enumerate} 
    \item PyTorch: The primary deep learning framework used for model building and training. 
    \item snnTorch (or other SNN library): This enables the spiking neuron functionality, providing the necessary classes for implementing spiking neurons and layers. 
    \item Tonic (if working with event-based data): A library for event-based vision datasets, facilitating easy access and processing of neuromorphic datasets. 
\end{enumerate} 
Ensure necessary dependencies, such as PyTorch, snnTorch, and Tonic, are installed as specified in the GitHub repository's README section.

\subsubsection{Data Type:}
The example begins by choosing the data type, which can either be event-based or RGB data. For this SCNN example, event data is used, specifically event-captured data from a neuromorphic dataset (e.g., N-MNIST). 

\subsubsection{Encoding:}
Once the data type is selected, it is essential to prepare it for SNN processing. For event-based data, encoding is not typically required, as the data is already in a spike-compatible format. If RGB data were used, Poisson encoding is a common approach to convert it into spike trains compatible with SNNs. Tools like snnTorch and Spikingjelly facilitate this transformation:
\begin{itemize}
    \item {\bf snnTorch} uses the \texttt{spikegen.rate} function to encode RGB data, which should be formatted as a tensor (batch size, channels, height, width). By providing the input tensor and specifying the number of time steps, \texttt{spikegen.rate} converts the data into spike trains, resulting in a tensor of shape (time steps, batch size, channels, height, width).
    \item {\bf Spikingjelly} provides a similar function under the name of  \texttt{PoissonEncoder}. Given the RGB tensor and desired time steps, it generates spike trains, matching the shape produced by snnTorch and capturing the temporal dynamics of the input.
\end{itemize}

An example of Spikingjelly's Poisson encoding applied to the MNIST dataset can be found in the repository.

\subsubsection{Architecture:}
The SCNN network architecture is defined in the \texttt{Net} class, which inherits from \texttt{nn.Module}, PyTorch’s base class for neural networks. This class initializes the network layers and defines the forward pass to dictate how data flows through the network.

The SCNN comprises two convolutional layers (\texttt{conv1} and \texttt{conv2}) followed by a fully connected layer (\texttt{fc1}). Each convolutional layer is followed by a max pooling operation to reduce spatial dimensions while retaining prominent features.
\begin{enumerate} 
    \item \texttt{conv1}: Contains 12 filters, each of size 5 $\times$ 5, to extract spatial features from the input data.
    \item \texttt{conv2}: Contains 64 filters of size 5 $\times$ 5, providing deeper feature extraction.
    \item \texttt{fc1}: A fully connected layer mapping the extracted features to the output classes (typically corresponding to the number of target classes).
\end{enumerate}
After each neural layer (convolutional or fully connected), a LIF neuron layer is added. The LIF layers (\texttt{lif1}, \texttt{lif2}, and \texttt{lif3}) transform the continuous outputs of their preceding layers into spikes based on neuron membrane potentials, allowing the network to operate in a spiking paradigm. This combination of conventional layers and LIF neurons forms the backbone of an SNN.

\subsubsection{Learning Method}
The example uses a supervised learning approach, using gradient-based optimization with surrogate gradients to iteratively adjusts the model parameters to minimize the loss function over a series of epochs. Below is a step-by-step breakdown of the training loop for the SCNN:

\begin{enumerate}

    \item {\bf Gradient Reset:} Before performing the forward pass, the gradients from the previous batches must be reset to zero using \texttt{optimizer.zero\_grad()}. This step is vital, as gradients accumulate by default in PyTorch. If not reset, gradients from earlier batches would interfere with subsequent updates, leading to incorrect parameter adjustments.
    
    \item {\bf Forward Pass:} The forward pass processes input data through the convolutional and fully connected layers, as well as max pooling operations and spiking neuron transformations. At the start of the forward pass, the model is set to training mode using \texttt{model.train()}, which ensures that layers like dropout and batch normalization operate in their training configurations, introducing regularization.

    The key components include:

    \begin{itemize}
        \item {\bf First Convolutional Layer:}
        \begin{itemize}
            \item The input data \texttt{x} is passed through the first convolutional layer (\texttt{conv1}). This layer applies 12 filters of size 5 $\times$ 5 to extract spatial features from the input.
            \item The resulting feature maps are then passed through a max pooling operation using \texttt{F.max\_pool2d}, reducing the spatial dimensions and retaining the most prominent features.
            \item The pooled output is then fed into the first LIF neuron layer (\texttt{lif1}). This layer integrates the incoming signal over time and produces spikes (\texttt{spk1}) whenever the membrane potential crosses a certain threshold. Along with the spikes, the membrane potentials are updated (\texttt{mem1}), storing the accumulated potential for use in the next time step.
        \end{itemize}

        \item {\bf Second Convolutional Layer:}
        \begin{itemize}
            \item The spikes generated from the first LIF neuron layer (\texttt{spk1}) are used as input for the second convolutional layer (\texttt{conv2}). This layer uses 64 filters of size 5 $\times$ 5 to extract deeper spatial features from the spiking input.
            \item Similar to the first layer, the output is max-pooled to reduce spatial dimensions while retaining the key features.
            \item The pooled feature maps are then processed through the second LIF neuron layer (\texttt{lif2}). This layer produces a new set of spikes (\texttt{spk2}) based on the accumulated membrane potential and also updates the potential state (\texttt{mem2}).
        \end{itemize}

        \item {\bf Fully Connected Layer:}
        \begin{itemize}
            \item The spikes from the second LIF neuron layer (\texttt{spk2}) are flattened into a vector and passed through the fully connected layer (\texttt{fc1}). This layer maps the spiking features to a set of 10 output neurons, typically corresponding to the number of target classes.

            \item The output from the fully connected layer is then processed through the third LIF neuron layer (\texttt{lif3}). This layer produces final spikes (\texttt{spk3}) and updates the membrane potentials (\texttt{mem3}), simulating the spiking output of the network. 
        \end{itemize}
    \end{itemize}

    \item {\bf Backward Pass \& Optimization:}
    The training loop is structured to iteratively minimize the loss by updating model parameters:
    
    \begin{itemize}

        \item {\bf Loss Function}
        Cross-Entropy Loss is used as the objective function to guide the network's learning process. Cross-Entropy is a standard choice for classification tasks, particularly when dealing with multiple classes, as it measures the dissimilarity between predicted probability distribution and the true label distribution.

        The Cross-Entropy Loss function is defined as:
        \begin{equation}
        \mathcal{L} = -\sum_{i} y_i \log(\hat{y}_i),
        \end{equation}
        Where \(y_i\) represents the true label in a one-hot encoded format \cite{bishop1995neural}, and \(\hat{y}_i\) denotes the  predicted probability for class \(i\). The summation is performed over all classes in the dataset. This formulation leverages the Kullback-Leibler (KL) divergence between the true label distribution and the predicted distribution, effectively penalizing the network when the predicted probabilities diverge from the target distribution \cite{kullback1951information}. 

        The running loss is accumulated to provide an indication of the model's learning progress throughout the epoch. Tracking the average loss over an epoch, which is printed for each epoch, serves as a diagnostic tool. It enables monitoring of the convergence behavior and can signal issues such as vanishing gradients, overfitting, and underfitting based on the loss trajectory.
        
        \item {\bf Backpropagation:} Once the loss is computed, backpropagation (\texttt{loss.backwards()}) is initiated to calculate gradients. Backpropagation is the process of computing the gradient of the loss function with respect to each model parameter by applying the chain rule of calculus through the network layers. This step populates the gradients within each layer, providing the necessary information to adjust the weights and biases.

        \item {\bf Parameter Update:} The optimizer (commonly Adam or SGD) uses the gradients calculated during backpropagation to update the model parameters via \texttt{optimizer.step}. 
        
        Specifically, each parameter $\theta$ is updated as:

        \begin{equation}
            \theta \leftarrow \theta - \eta \cdot \frac{\partial \mathcal{L}}{\partial \theta}
        \end{equation} 

        Where $\eta$ is the learning rate (a hyperparameter controlling the step size of the update), and $\frac{\partial \mathcal{L}}{\partial \theta}$ is the gradient of the loss with respect to the parameter $\theta$.

        The optimizer also incorporates specific strategies for each algorithm:
        \begin{itemize}
            \item {\bf SGD:} Uses a constant learning rate, optionally with momentum to accelerate convergence by using a moving average of past gradients.

            \item {\bf Adam:} Computes adaptive learning rates for each parameter using first (mean) and second (variance) moment estimates of gradients. This helps in faster convergence and better handling of sparse gradients.
        \end{itemize}
    \end{itemize}

    This process (forward pass, loss calculation, backwards pass, parameter update) is repeated across each epoch. An epoch refers to a single pass through the entire training dataset. Multiple epochs are often required for the model to converge to an acceptable local minimum of the loss function. The goal is to iteratively adjust the model's parameters such that the loss decreases, resulting in improved performance on the training data. 
    
\end{enumerate}

\subsubsection{Implementation Medium}
The SCNN framework is implemented and trained using the simulation environment provided by PyTorch and snnTorch. These libraries allow flexible model design, training, and evaluation without the need for specialized hardware. Since the examples in this framework are intended primarily for experimentation, education, and research purposes, they are not specifically designed to run on neuromorphic hardware.

While there are libraries and platforms that support exporting models to neuromorphic hardware (as discussed in Section \ref{sec:implementation_medium}), the focus of this example is on software-based simulation. This enables users to easily prototype, modify, and experiment with SNN architectures without the constraints of hardware-specific implementations.
    
\subsubsection{Evaluation}
The evaluation of the SCNN is a critical phase to assess its performance on a test dataset post-training. This involves transitioning the model into an inference state, making predictions, and calculating metrics such as accuracy to quantify classification performance.

First the SCNN is set to evaluation mode using \texttt{model.eval()}. This step ensures that layers such as dropout and batch normalization, which behave differently during training, are properly adjusted for inference. To improve computational efficiency, gradient calculations are disabled using the \texttt{torch.no\_grad()} context manager. Since gradients are unnecessary during evaluation, this not only reduces memory consumption but also accelerates the process.

The test data is then iterated over in batches using a data loader, allowing the model to efficiencly process large datasets. A forward pass through the network is performed on each batch, generating predictions based on the spikes from the LIF neurons. These spikes are interpreted as a probability distribution over the output classes. The predicted class is typically determined by taking the argmax of the model's output, which represents the highest possibility among the possible classes.

Performance metrics are computed by comparing the model's predictions against the true labels. Accuracy is commonly used as an evaluation metric, calculated as the proportion of correct predictions out of the total number of samples. This provides a direct measure of the model's ability to generalize unseen data.

\section{Future Directions}
\label{sec:future_directions}
The field of SNNs is evolving rapidly, and numerous exciting avenues remain to be explored. This section highlights key future research directions for advancing SNNs, focusing on improved learning algorithms, neuromorphic hardware integration, and scalability. The following areas hold promise for advancing the state-of-the-art in SNNs.

\begin{enumerate}

 \item {\bf Advanced Learning Algorithms:}
While significant progress has been made in developing learning algorithms for SNNs, there is still a need for more sophisticated methods that can better exploit the temporal dynamics of spiking neurons \cite{pietrzak2023overview, zenke2018, javanshir2022advancements}. Future research should explore hybrid learning algorithms that combine the strengths of existing learning methods, such as STDP, backpropagation, and reinforcement learning \cite{rathi2020enabling}. Additionally, the development of learning rules that can adapt in real time and handle continuous streams of data will be crucial for applications in dynamic environments \cite{zenke2017continual}.

\item {\bf Neuromorphic Hardware Integration:}
The integration of SNNs with neuromorphic hardware provides substantial benefits in energy efficiency and computational speed \cite{davies2018loihi}. Future work will focus on optimizing SNN architectures for specific neuromorphic platforms like Intel’s Loihi or IBM’s TrueNorth, which includes developing hardware-friendly algorithms that leverage the unique capabilities of these platforms and exploring new hardware designs tailored for SNNs \cite{schuman2022opportunities, javanshir2022advancements}. Additionally, while efforts toward standardizing benchmarks and evaluation protocols for neuromorphic systems are underway, further progress is necessary to enable clearer comparisons and improvements across different approaches \cite{ostrau2020benchmarking, kulkarni2021benchmarking, muller2022braille}.

 \item {\bf Scalability and Efficiency Improvements:}
Scalability remains a significant challenge for SNNs, especially when dealing with large-scale datasets and complex tasks \cite{roy2019towards}. Research should aim to develop more efficient SNN architectures that can scale to millions of neurons and synapses without sacrificing performance \cite{pfeiffer2018deep}. Techniques such as model compression, pruning, and weight sharing can be explored to reduce the memory and computational requirements of SNNs \cite{deng2021comprehensive}. Additionally, the development of more efficient spike encoding schemes and communication protocols will be important for improving the overall efficiency of SNNs \cite{park2016hierarchical}.

 \item {\bf Temporal and Spatio-Temporal Learning:}
One of the unique advantages of SNNs is their ability to process temporal information \cite{maass1997networks}. Further research is needed on leveraging this capability by developing algorithms that can learn complex temporal and spatio-temporal patterns \cite{she2021sequence}. This includes exploring novel temporal coding schemes and learning rules that can capture the intricate dynamics of time-dependent data \cite{bellec2018long}. Applications such as object tracking, action recognition, and autonomous navigation, where temporal processing is critical, stand to benefit greatly from these advancements \cite{chen2020event}.

 \item {\bf Interpretability and Explainability:}
Understanding the decision-making process of SNNs is crucial for their adoption in critical applications \cite{pfeiffer2018deep}. Future work should aim to improve the interpretability and explainability of SNN models \cite{nguyen2021temporal, kim2021visual}. This involves developing techniques to visualize and analyze the activity of spiking neurons and the synaptic weight changes during learning \cite{tavanaei2019deep}. Additionally, research is needed on creating methods to extract meaningful features and patterns from the spiking activity, providing insights into how SNNs process information and make decisions \cite{gardner2015learning}.

 \item {\bf Benchmarking and Standardization:}
To facilitate the comparison and evaluation of different SNN approaches, standardized benchmarking datasets and evaluation protocols are needed \cite{pfeiffer2018deep}. Future research is needed to establish comprehensive benchmarking suites that cover a wide range of tasks and scenarios, including both synthetic and real-world datasets \cite{liu2016benchmarking, lin2024benchmarking}. Standardized benchmarks will enable researchers to systematically evaluate the performance, efficiency, and robustness of SNN models, driving further improvements in the field \cite{gardner2015learning}.

 \item {\bf Applications in Real-World Scenarios:}
Finally, the application of SNNs in real-world scenarios remains an exciting and largely untapped area \cite{pfeiffer2018deep, roy2019towards}. Future research should explore the deployment of SNNs in various domains, such as autonomous vehicles, healthcare, and IoT \cite{yamazaki2022spiking}. These applications will not only provide valuable insights into the practical challenges and limitations of SNNs but also demonstrate their potential to revolutionize various industries with their unique capabilities \cite{o2013real}.

\end{enumerate}

\section{Conclusion}
Spiking Neural Networks (SNNs) represent a significant advancement toward biologically plausible and energy-efficient computational models by emulating the temporal dynamics and spike-based communication of biological neurons, offering advantages for real-time processing and low-power applications. This paper has explored foundational aspects of SNNs, including neuronal dynamics, architectural innovations, learning rules, and evaluation methodologies. Understanding biological neurons and neural coding is essential for designing effective SNN models that capture temporal characteristics.

Prominent datasets have facilitated benchmarking and advancement in SNN research, enabling training and evaluation on complex visual tasks. The interplay between software simulations and neuromorphic hardware is crucial; simulations offer flexibility for experimentation, while neuromorphic hardware enhances speed and energy efficiency for practical applications. Architectural innovations such as hierarchical, convolutional, and recurrent networks demonstrate the versatility and scalability of SNNs, allowing them to address tasks from pattern recognition to complex sequential data processing. Advances in learning rules—from unsupervised methods like Spike-Timing-Dependent Plasticity (STDP) to supervised strategies—have improved training efficiency and performance by leveraging the temporal nature of spikes.Evaluating SNNs requires considering metrics like accuracy, energy consumption, latency, and memory footprint to ensure practical viability. 

Looking forward, the future of SNN research is bright, with numerous avenues for innovation. Enhancements in neuromorphic hardware, coupled with more robust training algorithms and better integration with existing deep learning frameworks, will drive the next wave of advancements. As the SNN footprint continues to expand, this paper will serve as a comprehensive guide for researchers, students, and practitioners, providing insights and foundational knowledge to support future advancements in SNN development and application.

\bibliographystyle{IEEEtran}
\bibliography{ref}

\begin{IEEEbiography}
[{\includegraphics[width=1in,height=1.25in,clip,keepaspectratio]{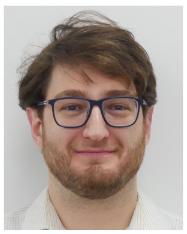}}]{\textbf{Craig Iaboni}} received his B.S. (2020) in Computer Science from the New Jersey Institute of Technology (NJIT). He is currently pursuing a Ph.D. in Computer Science from NJIT. His research interests are at the intersection of computer vision, deep learning, and neuromorphic computing with event-based vision systems.
\end{IEEEbiography}

\begin{IEEEbiography}
[{\includegraphics[width=1in,height=1.25in,clip,keepaspectratio]{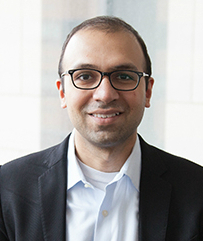}}]{\textbf{Pramod Abichandani, Ph.D.}} is an assistant professor in the ECET/ECE/CS department at New Jersey Institute of Technology (NJIT). He received his Bachelors of Engineering (B.E.) degree in 2005 from Nirma Institute of Technology, Gujarat University, India, and his M.S and Ph.D. degrees in Electrical and Computer Engineering from Drexel University in 2007 and 2011, respectively. His research interests are centered around optimal, multi-dimensional, data-driven decision-making, through the use of techniques from mathematical programming, linear and nonlinear systems theory, statistics, and machine learning. On the education front, he works on bringing innovation to the classroom by introducing novel course content, pedagogical methodologies, and assessment techniques.
\end{IEEEbiography}

\end{document}